\documentclass[11pt]{article}

\PassOptionsToPackage{table}{xcolor}
\usepackage[final]{acl}

\usepackage{times}
\usepackage{latexsym}
\usepackage[T1]{fontenc}
\usepackage[utf8]{inputenc}
\usepackage{microtype}
\usepackage{inconsolata}
\usepackage{graphicx}

\usepackage{amsmath,amssymb,mathtools}
\usepackage{booktabs}
\usepackage{multirow}
\usepackage{array}
\newcolumntype{P}[1]{>{\raggedright\arraybackslash}p{#1}}
\usepackage[normalem]{ulem}
\usepackage[most]{tcolorbox}
\usepackage{tikz}
\usepackage{fontawesome5}
\usepackage{xspace}
\usepackage{enumitem}
\usepackage{placeins}
\usepackage{algorithm}
\usepackage{algpseudocode}
\extrafloats{150}
\algrenewcommand\algorithmicrequire{\textbf{Input:}}
\algrenewcommand\algorithmicensure{\textbf{Output:}}
\usetikzlibrary{arrows.meta,backgrounds,calc,fit,positioning,shapes.geometric,decorations.pathreplacing,matrix}

\newcommand{\figfont}{\normalfont\footnotesize}
\tikzset{
  air figure/.style={every node/.append style={font=\figfont}},
  air title/.style={font=\figfont\bfseries}
}

\definecolor{refreshblue}{HTML}{1F77B4}
\definecolor{reusegray}{HTML}{8C8C8C}
\definecolor{agentgreen}{HTML}{2CA02C}
\definecolor{lossred}{HTML}{D62728}
\definecolor{warmorange}{HTML}{FF7F0E}
\definecolor{violet}{HTML}{9467BD}
\definecolor{softblue}{HTML}{DEEBF7}
\definecolor{softgreen}{HTML}{E3F4DF}
\definecolor{softred}{HTML}{FBE0DF}
\definecolor{softorange}{HTML}{FFE8D1}
\definecolor{softviolet}{HTML}{ECE4F4}
\definecolor{softgray}{HTML}{F3F3F3}

\definecolor{AIRNavy}{HTML}{1F3B5A}
\definecolor{AIRTeal}{HTML}{138D99}
\definecolor{AIRAccent}{HTML}{D62728}
\definecolor{AIRGroup}{HTML}{F2F2F2}
\definecolor{AIRHeader}{HTML}{EAF1F8}
\definecolor{AIRRow}{HTML}{FCE9E9}
\definecolor{AIRBest}{HTML}{DEEBF7}
\definecolor{AIRGain}{HTML}{2CA02C}
\definecolor{AIRLoss}{HTML}{D62728}
\definecolor{AIRTie}{HTML}{ECECEC}
\definecolor{AIRWinText}{HTML}{1B6E1B}
\definecolor{AIRLossText}{HTML}{B2182B}
\definecolor{AIRFallback}{HTML}{FFE8D1}
\definecolor{AIRLearned}{HTML}{E3F4DF}

\newcommand{\AIRswatch}[1]{\colorbox{#1}{\rule{0pt}{4pt}\hspace{4pt}}}

\definecolor{airnavy}{HTML}{1F3B5A}
\definecolor{taskpurple}{HTML}{8063AE}
\definecolor{agentblue}{HTML}{1F77B4}
\definecolor{toolslate}{HTML}{6E7781}
\definecolor{controlteal}{HTML}{138D99}
\definecolor{authoramber}{HTML}{C9720E}
\tcbset{aircard/.style={enhanced,boxrule=.4pt,arc=1.2mm,left=1.4mm,right=1.4mm,top=1.0mm,bottom=1.0mm,before skip=2.2pt,after skip=2.2pt,colback=white,colframe=toolslate!65,fonttitle=\figfont\bfseries,coltitle=white,attach boxed title to top left={yshift=-.5mm,xshift=1.0mm},boxed title style={boxrule=0pt,arc=.7mm,left=1mm,right=1mm,top=.6mm,bottom=.6mm},fontupper=\figfont\raggedright}}
\newtcolorbox{rolebox}[2][]{aircard,colback=#2!5,colframe=#2!60,colbacktitle=#2,#1}
\newcommand{\caseheading}[2]{\par\noindent{\color{airnavy}\figfont\bfseries #1}\par\vspace{1pt}{\figfont #2}\par\vspace{3pt}}
\newcommand{\smalllabel}[1]{{\color{airnavy}\bfseries #1}}
\newcommand{\caseboxes}{\setlength{\parindent}{0pt}\setlength{\parskip}{3pt}\emergencystretch=2em\figfont}

\newcommand{\E}{\mathbb{E}}
\newcommand{\Bset}{\mathcal{B}}
\newcommand{\D}{\mathcal{D}}
\newcommand{\TopK}[1]{\operatorname{Top\text{-}\mathnormal{#1}}}
\let\origleft\left
\let\origright\right
\renewcommand{\left}{\mathopen{}\mathclose\bgroup\origleft}
\renewcommand{\right}{\aftergroup\egroup\origright}
\newcommand{\Full}{\texttt{F}}
\newcommand{\Shared}{\texttt{S}}
\newcommand{\Tcache}{\mathcal T_{\mathrm{cache}}}
\newcommand{\Lidx}{\mathcal N_{\idx}}
\newcommand{\Nidx}{N_{\idx}}
\newcommand{\NLL}{\mathrm{NLL}}
\newcommand{\method}{\textsc{AIR}\xspace}
\newcommand{\idx}{\mathrm{idx}}
\newcommand{\ctrl}{\mathrm{ctrl}}

\title{When Should an Agent Refresh? Adaptive Index Reuse for Efficient Long-Horizon Reasoning}

\author{%
  \textbf{Guangyuan Wang}\textsuperscript{1,\,2}\quad
  \textbf{Yifan Zhang}\textsuperscript{3}\quad
  \textbf{Shilong Liu}\textsuperscript{3,\,4}
  \\[0.6em]
  {\normalfont
    \textsuperscript{1}Mila -- Qu\'ebec Artificial Intelligence Institute\quad
    \textsuperscript{2}McGill University}
  \\
  {\textsuperscript{3}Princeton University\quad
    \textsuperscript{4}Columbia University}
}

\begin{document}
\maketitle
\begin{abstract}
Language agents read code, tool output, and dialogue in contexts that grow at every step. Dynamic sparse attention lets a small indexer choose the earlier tokens each query attends to, but the indexer still scores the whole prefix at every layer. Cross-layer reuse methods such as IndexCache rerun the indexer at only some layers, with one fixed pattern for every input. In an agent, the relevant evidence can change abruptly when a traceback or test result arrives, so fresh selections matter more at some steps than at others. We propose Adaptive Index Reuse (\method), which keeps IndexCache's execution and learns, for each segment of an agent episode, how many layers refresh their selection. A small controller reads signals from completed computation and the latest tool event, chooses a pattern from a nested library, and learns from terminal task reward under an expected compute budget. On Qwen3-4B-Instruct-2507 with a 128K context, \method improves success over the strongest static schedule at matched cost on SWE-bench Verified (27.0\% to 32.4\%), Terminal-Bench 2.0, and three $\tau^2$-bench domains, and it lowers time per successful task on all five. Across 50 backbone, context, and benchmark settings, it matches or exceeds that schedule in 44.
\end{abstract}

\section{Introduction}
\label{sec:intro}

Tool-using language agents work over contexts that keep growing. A coding agent reads an issue, searches files, reads a traceback, edits code, and runs tests; a customer-service agent alternates between dialogue, policy lookup, and API calls. Every generated token attends over a key-value cache that holds this whole history, so long trajectories make each decision expensive.

\begin{figure*}[t]
\centering
\begin{tikzpicture}[
  air figure, x=0.98cm, y=1cm,
  every node/.append style={font=\figfont\scriptsize},
  card/.style={rounded corners=4pt, line width=.6pt},
  tab/.style={rounded corners=2.5pt, line width=.5pt, inner sep=1.8pt, font=\figfont\scriptsize\bfseries, fill=white},
  txt/.style={anchor=north west, align=left, inner sep=0pt, text width=#1},
  ev/.style={rounded corners=3pt, line width=.5pt, inner sep=2pt, align=center, minimum height=.62cm},
  cell/.style={draw=reusegray!80, fill=white, minimum width=1.9mm, minimum height=2.5mm, inner sep=0pt},
  fcell/.style={cell, draw=refreshblue, fill=softblue},
  lay/.style={draw=reusegray!80, fill=white, rounded corners=1.2pt, minimum width=4.6mm, minimum height=4.2mm, inner sep=0pt, font=\ttfamily\scriptsize},
  flay/.style={lay, draw=refreshblue, fill=softblue},
  kv/.style={draw=agentgreen!70, fill=softgreen, minimum width=1.5mm, minimum height=1.5mm, inner sep=0pt},
  ik/.style={draw=violet!70, fill=softviolet, minimum width=1.5mm, minimum height=1.5mm, inner sep=0pt},
  arr/.style={-{Latex[length=1.5mm,width=1.1mm]}, line width=.55pt, draw=black!60},
  thin/.style={-{Latex[length=1.1mm,width=.9mm]}, line width=.4pt}
]
\node[anchor=west, font=\figfont\footnotesize\bfseries] at (0,5.55) {\faRobot\ \ Refresh effort follows the agent's state};
\node[ev, draw=agentgreen!70, fill=softgreen!60, text width=2.55cm] (e1) at (1.45,4.75) {\faCode\ routine edit\\[-1pt]{\color{refreshblue}segment $u{-}1$: $R = 3$}};
\node[ev, draw=lossred!70, fill=softred, text width=2.6cm] (e2) at (5.0,4.75) {\faTerminal\ tool returns a\\[-1pt]{\color{lossred}new test failure}};
\node[ev, draw=refreshblue!80, fill=softblue!70, text width=2.55cm] (e3) at (8.55,4.75) {\faSearch\ read the failing branch\\[-1pt]{\color{refreshblue}segment $u$: $R = 18$}};
\node[ev, draw=agentgreen!70, fill=softgreen!60, text width=2.4cm] (e4) at (12.0,4.75) {\faCheckCircle\ next result\\[-1pt]{\color{black!70}feeds decision $u{+}1$}};
\draw[arr] (e1) -- (e2); \draw[arr] (e2) -- (e3); \draw[arr] (e3) -- (e4);
\draw[black!35, line width=.5pt, -{Latex[length=1.3mm]}] (0,4.15) -- (15.9,4.15) node[anchor=south east, inner sep=1pt, text=black!55] {time};
\node[anchor=west, text=black!60, text width=2.4cm, align=left] at (13.4,4.75) {budget for $u$ is fixed\\[-1pt]before $u$ computes};
\draw[card, draw=agentgreen!70, fill=softgreen!45] (0,0.35) rectangle (3.55,3.7);
\node[tab, draw=agentgreen!70, text=agentgreen!60!black, anchor=west] at (0.15,3.7) {\faEye\ State};
\node[txt=3.25cm] at (0.15,3.35) {\textbf{latest tool event}\\type, length, test or\\traceback flag};
\node[txt=3.25cm] at (0.15,2.35) {\textbf{trajectory}\\context length, steps};
\node[txt=3.25cm] at (0.15,1.55) {\textbf{finished segment}\\entropy, selection overlap,\\last budget, cost};
\draw[card, draw=violet!70, fill=softviolet!55] (3.95,0.35) rectangle (6.6,3.7);
\node[tab, draw=violet!70, text=violet!75!black, anchor=west] at (4.1,3.7) {\faSlidersH\ Controller};
\node[txt=2.35cm] at (4.1,3.35) {small MLP $\pi_\theta$};
\foreach \r/\p/\i in {3/0.06/0, 5/0.10/1, 9/0.22/2, 18/0.46/3, 36/0.16/4} {
  \node[anchor=east] at (5,2.6-0.34*\i) {$R = \r$};
  \fill[violet!55] (5.05,2.51-0.34*\i) rectangle ++(2.6*\p,0.18);
}
\draw[violet!85!black, line width=.7pt] (5.03,1.43) rectangle (6.32,1.74);
\node[txt=2.35cm] at (4.1,0.95) {sample $b_u$\\runs as sampled};
\draw[card, draw=warmorange!80, fill=softorange!60] (7.0,0.35) rectangle (9.85,3.7);
\node[tab, draw=warmorange!80, text=warmorange!70!black, anchor=west] at (7.15,3.7) {\faLayerGroup\ Mask library};
\node[txt=2.55cm] at (7.15,3.35) {nested, built offline};
\foreach \r/\i/\fl in {3/0/{0,5}, 5/1/{0,3,5}, 9/2/{0,2,3,5,7}, 18/3/{0,1,2,3,5,6,7}, 36/4/{0,1,2,3,4,5,6,7}} {
  \node[anchor=east] at (7.75,2.6-0.34*\i) {\r};
  \foreach \x in {0,...,7} {\node[cell] at (8.12+0.2*\x,2.6-0.34*\i) {};}
  \foreach \x in \fl {\node[fcell] at (8.12+0.2*\x,2.6-0.34*\i) {};}
}
\draw[warmorange!85!black, line width=.7pt, rounded corners=1pt] (7.99,1.42) rectangle (9.66,1.75);
\node[txt=2.55cm] at (7.15,0.8) {chosen mask $\mathbf c^{(b_u)}$};
\draw[card, draw=refreshblue!75, fill=softblue!55] (10.25,0.35) rectangle (15.9,3.7);
\node[tab, draw=refreshblue!75, text=refreshblue!75!black, anchor=west] at (10.4,3.7) {\faCogs\ Execute segment};
\node[txt=5.3cm] at (10.4,3.35) {prefill new tokens, then decode up to $Q=512$};
\foreach \lab/\i/\st in {F/0/flay, S/1/lay, S/2/lay, F/3/flay, S/4/lay} {
  \node[\st] (L\i) at (10.95+0.98*\i,2.05) {\lab};
  \foreach \k in {0,1,2} {\node[kv] at (10.78+0.98*\i+0.17*\k,1.6) {}; \node[ik] at (10.78+0.98*\i+0.17*\k,1.4) {};}
}
\foreach \i/\j in {0/1,1/2,2/3,3/4} {\draw[thin, draw=black!50] (L\i) -- (L\j);}
\node[draw=refreshblue, fill=white, rounded corners=1pt, inner sep=1.2pt] (B1) at (11.45,2.75) {IDs};
\draw[thin, draw=refreshblue] (L0.north) |- (B1.west); \draw[thin, draw=agentgreen!80!black] (B1.east) -| (L1.north); \draw[thin, draw=agentgreen!80!black] (B1.east) -| (L2.north);
\node[draw=refreshblue, fill=white, rounded corners=1pt, inner sep=1.2pt] (B2) at (14.39,2.75) {IDs};
\draw[thin, draw=refreshblue] (L3.north) |- (B2.west); \draw[thin, draw=agentgreen!80!black] (B2.east) -| (L4.north);
\node[kv] at (10.5,0.98) {}; \node[txt=2.0cm, anchor=west] at (10.62,0.98) {own KV per layer};
\node[ik] at (12.75,0.98) {}; \node[txt=2.9cm, anchor=west] at (12.87,0.98) {index keys, eligible layers};
\node[txt=5.3cm] at (10.4,0.72) {F writes position IDs; S reuses them};
\draw[arr] (3.55,2.05) -- (3.95,2.05); \draw[arr] (6.6,2.05) -- (7.0,2.05); \draw[arr] (9.85,2.05) -- (10.25,2.05);
\draw[arr] (e2.south) -- ++(0,-0.25) -| (1.78,3.7);
\draw[arr, draw=agentgreen!75!black] (13.1,0.35) -- (13.1,0.05) -- (1.78,0.05) -- (1.78,0.35);
\node[anchor=north, fill=white, inner sep=1pt, text=agentgreen!55!black] at (7.45,0.05) {next tool output and finished-segment statistics form $s_{u+1}$};
\draw[arr, dashed, draw=violet!70] (15.4,0.35) -- (15.4,-0.36) -- (5.28,-0.36) -- (5.28,0.35);
\node[anchor=north, fill=white, inner sep=1pt, text=violet!75!black] at (10.35,-0.36) {\faTrophy\ training only: terminal reward $V(\tau)$ and charged cost $C_u$ update $\pi_\theta$};
\end{tikzpicture}
\caption{\method's decision loop. Before segment $u$ computes, the controller maps state from finished work to a budget; the segment runs under that budget's nested mask. Full layers select position IDs; Shared layers reuse them. Every layer keeps its own KV cache, and the next observation feeds decision $u{+}1$.}
\label{fig:air-overview}
\end{figure*}
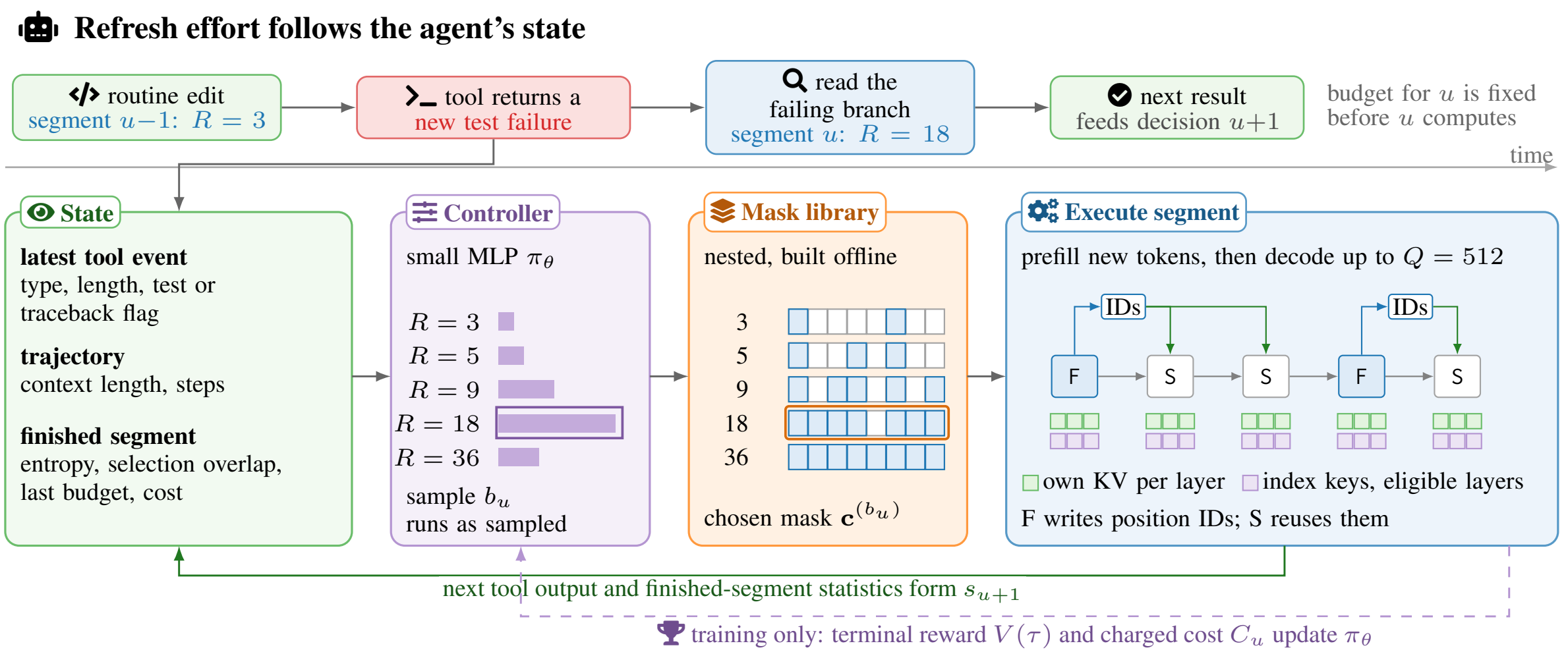

Dynamic sparse attention lowers this cost by letting each query attend to a small, query-dependent set of positions \citep{yuan2025nsa,deepseek2025v32}. In DeepSeek Sparse Attention (DSA), a lightweight indexer at every layer scores all earlier positions and passes the top $k$ to core attention \citep{deepseek2025v32}. Core attention then reads $k$ positions, while the indexer still scores the whole prefix at every layer, and this selection work becomes the cost that grows with context. IndexCache observes that adjacent DSA layers select largely overlapping positions \citep{bai2026indexcache}. It runs the indexer only at some \emph{Full} layers and lets each \emph{Shared} layer reuse the positions chosen by the nearest preceding Full layer; on a 30B DSA model its authors report removing 75\% of indexer computation, with up to $1.82\times$ prefill and $1.48\times$ decode speedups at 200K tokens. A greedy search on calibration loss fixes the pattern once, and the same pattern serves every input and every point of a generation.

We call recomputing the selected positions at a layer a \emph{refresh}. An agent's need for refreshes changes during an episode. During routine generation the relevant evidence moves slowly and tolerates aggressive reuse; a new traceback, file, test result, or user request can change it abruptly and make stale selections costly. Static reuse decides where in depth a model refreshes. This paper asks when, along a trajectory, a larger refresh budget is worth its cost, and how to learn that decision from task outcomes.

\begin{table}[t]
\centering
\small
\setlength{\tabcolsep}{3.5pt}
\begin{tabular}{lrrrrr}
\toprule
& \multicolumn{3}{c}{Success (\%) $\uparrow$} & \multicolumn{2}{c}{Min/success $\downarrow$} \\
\cmidrule(lr){2-4}\cmidrule(lr){5-6}
Benchmark & Static & \method & $\Delta$ & Static & \method \\
\midrule
SWE-bench Verified & 27.0 & \textbf{32.4} & $+5.4$ & 64.6 & \textbf{52.7} \\
Terminal-Bench 2.0 & 22.5 & \textbf{24.7} & $+2.2$ & 61.9 & \textbf{56.4} \\
$\tau^2$ retail & 53.3 & \textbf{57.2} & $+3.9$ & 14.1 & \textbf{13.1} \\
$\tau^2$ airline & 42.5 & \textbf{46.5} & $+4.0$ & 14.1 & \textbf{12.9} \\
$\tau^2$ telecom & 42.5 & \textbf{46.5} & $+3.9$ & 19.5 & \textbf{17.9} \\
\bottomrule
\multicolumn{6}{l}{\footnotesize Differences are calculated before rounding.}
\end{tabular}

\caption{\method raises success and lowers time per successful task on every benchmark. Qwen3-4B-Instruct-2507, 128K cap; Static is the strongest feasible static schedule ($R=18$) at matched expected cost.}
\label{tab:opening}
\end{table}

We introduce Adaptive Index Reuse (\method), which learns to allocate refresh computation from observable agent state (Figure~\ref{fig:air-overview}). \method keeps IndexCache's Full/Shared execution and its greedy pattern search, and it stores one pattern per refresh budget from the search path as a nested runtime library. Before each segment of the agent's work, a small controller reads signals that already exist, such as the type of the latest tool output and statistics of finished computation, and chooses a library entry for the next segment. Every eligible layer keeps storing index keys while it is Shared, so any entry can run at any segment, and the cost model charges this upkeep. The controller is trained from terminal task reward under an expected budget on indexing and controller compute, and validation data decide whether a trained controller or a static fallback is deployed.

On SWE-bench Verified, Terminal-Bench 2.0, and three $\tau^2$-bench domains, \method raises success over the strongest static schedule at matched cost by 2.2 to 5.4 points and cuts time per successful task by 6.8 to 18.4\% (Table~\ref{tab:opening}). Controls show that the gain needs both state information and credit for delayed effects. The gain is uneven: it shrinks, ties, or reverses in 13 of 50 settings, all-refresh execution still scores higher on two benchmarks, and tools take enough of each episode that whole-agent speedups over all-refresh are 1.05 to 1.11. Our contributions are:
\begin{itemize}[leftmargin=*,itemsep=1pt,topsep=2pt,parsep=0pt]
\item a formulation of refresh allocation in agents as constrained control from causal observations, with the nested library as a hierarchical action and a cost model that charges index-key maintenance, selection, and controller overhead;
\item a training and deployment procedure that combines a replay-based warm start with standard Lagrangian policy gradient and selects the deployed policy on held-out validation data;
\item a controlled evaluation on five backbones, two context caps, and five agent benchmarks, with controls that separate layer placement, timing, state information, and delayed credit.
\end{itemize}

\section{Related Work}
\label{sec:related}

\begin{figure}[t]
\centering
\begin{tikzpicture}[
  air figure, x=1cm, y=1cm,
  st/.style={draw=#1!75, fill=#1!12, rounded corners=3pt, line width=.5pt, text width=4.4cm, minimum height=.62cm, align=left, inner sep=2.2pt, font=\figfont},
  data/.style={font=\figfont, text=black!60, align=right, anchor=east},
  arr/.style={-{Latex[length=1.4mm,width=1mm]}, line width=.5pt, draw=black!60}
]
\node[st=refreshblue] (s1) at (4.75,3.6) {\textbf{1 Indexer:} distill each layer's dense attention};
\node[st=warmorange] (s2) at (4.75,2.7) {\textbf{2 Library:} greedy elimination gives nested masks};
\node[st=agentgreen] (s3) at (4.75,1.8) {\textbf{3 Warm start:} replay masks on logged text};
\node[st=violet] (s4) at (4.75,0.9) {\textbf{4 Policy gradient:} reward minus priced cost};
\node[st=lossred] (s5) at (4.75,0.0) {\textbf{5 Selection:} learned checkpoint or static fallback};
\foreach \a/\b in {s1/s2,s2/s3,s3/s4,s4/s5} {\draw[arr] (\a) -- (\b);}
\node[data] at (2.3,3.6) {\tt \tiny indexer training};
\node[data] at (2.3,2.7) {\tt \tiny  schedule calibration};
\node[data] at (2.3,1.8) {\tt \tiny  controller training};
\node[data] at (2.3,0.9) {\tt \tiny  controller training};
\node[data] at (2.3,0.0) {\tt \tiny  controller validation};
\end{tikzpicture}
\caption{Offline preparation and training. Each stage uses its own owner-disjoint fitting partition and is frozen before the next; validation selects a learned controller or a static fallback before any final run.}
\label{fig:pipeline}
\end{figure}
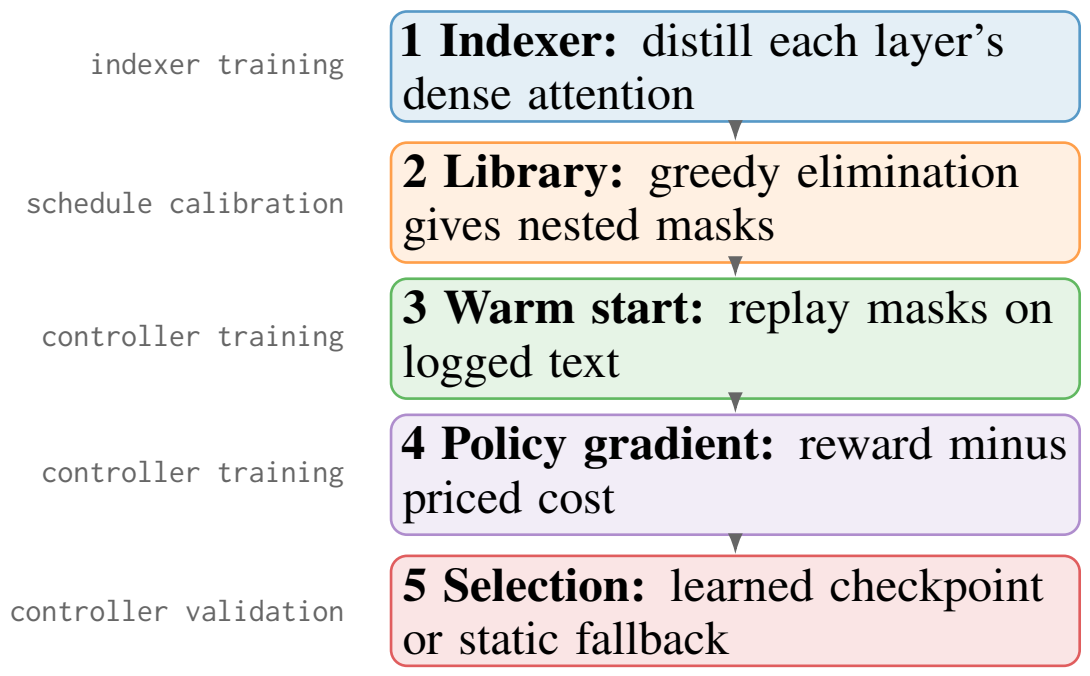

\paragraph{Cheaper selection.}
Many methods make one sparse selection cheaper: token, block, or page selection \citep{jiang2024minference,tang2024quest,yuan2025nsa}, a lightweight learned indexer \citep{deepseek2025v32}, routing each query to a few indexer heads \citep{zhou2026misa}, block-then-token search \citep{xu2026hisa}, or a compressed index-key representation shared across layers with layer-specific selection \citep{wang2026latentindex}. \method leaves the cost of each indexer call unchanged and decides which layers make a call, so these methods are both competitors and candidates for composition.

\paragraph{Cross-layer reuse.}
TidalDecode, Kascade, IndexCache, and LongCat Sparse Attention reuse selections across layers with a structure fixed per model \citep{yang2025tidaldecode,deshmukh2025kascade,bai2026indexcache,zan2026longcat}; some architectures also share key-value tensors across layers or are pretrained with sparse attention \citep{brandon2024cla,sun2024yoco,sun2026yoio,gao2026hysparse,wei2026hysparse2,lai2026msa}. \method inherits IndexCache's Full/Shared execution, its nearest-Full reuse rule, and its greedy search. Its additions are the per-segment choice among nested patterns, the observations that drive it, index-key maintenance at every eligible layer, and training from task reward under a compute budget.

\paragraph{Online refresh decisions.}
HShare fixes a sharing configuration per input batch after prefill \citep{wu2025hshare}, Slow-Fast Inference refreshes a reused token set near sentence boundaries \citep{xie2026slowfast}, ReTopK reuses the selections of similar earlier queries with an exact fallback \citep{yao2026retopk}, and a model can be prompted to declare its own attention scope \citep{ho2026declarative}. Mixture-of-Depths and CALM allocate computation across depth adaptively \citep{raposo2024mod,schuster2022calm}. In \method the decision variable is a complete depth pattern of indexer calls for each agent segment, learned from terminal reward. The learning procedure uses standard Lagrangian actor-critic tools \citep{williams1992reinforce,borkar2005actor,tessler2019rcpo}.

\paragraph{Serving systems.}
Inference engines such as vLLM and SGLang \citep{kwon2023vllm,zheng2024sglang} manage cache memory and batching, and several compared methods are released as patches to them; our MiniMax-M2 runs attach the indexer through an attention adapter in a serving engine. Speculative decoding \citep{leviathan2023speculative} and KV-cache eviction \citep{zhang2023h2o} reduce other costs; we do not combine them with \method, and speculative decoding is disabled in our MiniMax-M2 runs. Systems that tier model state across memories \citep{niko1221strata} compete with \method's index keys for accelerator memory (Appendix~\ref{app:limitations}).

\section{Adaptive Index Reuse}
\label{sec:method}

\subsection{Sparse selection and cross-layer reuse}
\label{sec:background}

At layer $\ell$ and query position $t$, a DSA-style indexer scores every candidate position $p$ with $H_{\idx}$ small heads,
\begin{equation}
I^{(\ell)}_{t,p}=\sum_{r=1}^{H_{\idx}}g^{(\ell,r)}_t\operatorname{ReLU} \left(\frac{\langle\widetilde{\mathbf q}^{(\ell,r)}_t,\widetilde{\mathbf k}^{(\ell,r)}_p\rangle}{\sqrt{d_{\idx}}}\right),
\label{eq:index-score}
\end{equation}
where $\widetilde{\mathbf q}$ and $\widetilde{\mathbf k}$ are index-query and index-key projections of the layer input and $g$ is a softmax gate over heads. The layer keeps the $k$ highest-scoring candidates, $\mathcal T^{(\ell)}_t=\TopK{k}(\mathbf I^{(\ell)}_t)$, and core attention reads them together with a few sink positions and a recent window from the layer's own key-value cache. Candidates exclude the sinks and the window, so for a fixed number of attended positions the core-attention work is the same whichever layer chose them (Appendix~\ref{app:ext-background}).

IndexCache assigns each eligible layer a role, Full or Shared. A Full layer runs the indexer; a Shared layer reuses the positions of the nearest preceding eligible Full layer,
\begin{equation}
\begin{aligned}\mathcal T^{(\ell)}_t&=\mathcal T^{(f(\ell))}_t,\\ f(\ell)&=\max\{\ell'\in\Lidx:\ell'<\ell,\ c_{\ell'}=\Full\},\end{aligned}
\label{eq:indexcache}
\end{equation}
and the first eligible layer is always Full. Reuse shares only position IDs, which travel through a buffer $\Tcache$ that lives for one query position. With the pattern F, S, S, F, S, the first layer writes its selection, the next two attend over it, the fourth overwrites it, and the fifth attends over the fourth layer's selection; each layer gathers its own keys and values at the shared positions. In \method, every eligible layer also projects and stores the index key of each new token while it is Shared. Its index-key cache then covers the whole prefix, so a Shared layer can become Full at a later segment without replaying past states.

\subsection{Decisions, budgets, and the library}
\label{sec:execution}

An agent episode alternates between environment observations and model responses. The controller decides at segment boundaries: a segment starts at every model invocation and after every $Q=512$ decoded tokens, and it covers the prefill of newly appended tokens and the decoding that follows. The budget for segment $u$ is chosen before any of its tokens is computed, because it fixes which layers run the indexer for all of them.

Each budget names one complete depth pattern from a library built offline. Following training-free IndexCache, we score a pattern by the next-token loss of agent-generated calibration sequences run under it. Greedy elimination starts from the all-Full pattern and repeatedly converts to Shared the layer whose conversion raises this loss least. Each budget stores the pattern reached on this single path when its number of Full layers $R_b$ is reached, so the library is nested: a larger budget adds Full layers without moving existing ones. The primary backbone, with 36 eligible layers, uses $R_b\in\{3,5,9,18,36\}$; other backbones scale these counts to their number of eligible layers (Appendix~\ref{app:library}).

\paragraph{Observations.}
Before segment $u$ the controller reads $s_u=\phi(\mathcal H_u)$, a fixed-length summary of the history $\mathcal H_u$ computed only from finished computation. It combines the position in the trajectory (context length, segment index, tool steps), the latest tool event (its type, length, and traceback or test-summary flags), the previous budget and cumulative cost, and model and index statistics of the completed segment, such as predictive entropy and the overlap between recent selections. A low budget therefore never receives information that only a refresh would have produced (Appendix~\ref{app:features}).

\subsection{Cost model}
\label{sec:cost}

The controller's cost signal counts, analytically from tensor shapes, every operation whose amount the controller changes or adds. Index-key maintenance is paid at every eligible layer for every token under every budget; selection, comprising the index query, scoring, and top-$k$, is paid only at Full layers; and runtime and controller overhead covers buffer traffic, pattern switches, feature extraction, and the controller's inference. Each operation costs its floating-point operations plus its bytes moved times a fixed rate $\kappa$ (the accelerator's peak BF16 throughput divided by its memory bandwidth). The indexing cost of segment $u$ is
\begin{equation}
C_{\idx,u}=\sum_{t\in\mathcal X_u}\sum_{\ell\in\Lidx}\left(F^{t\ell}_{\mathrm{key}}+\mathbf 1[c_{u,\ell}=\Full]\,F^{t\ell}_{\mathrm{sel}}\right),
\label{eq:index-cost}
\end{equation}
over the segment's query positions $\mathcal X_u$, and the segment cost $C_u$ adds the overhead terms. Core attention is excluded, because its work per query is the same under every budget. Episode costs are normalized by $C_0$, the mean all-refresh episode cost on controller-training data, and the budget $\overline B$ bounds the expected normalized cost. Measured latency, full-model compute, and memory are reported separately (Appendix~\ref{app:counting}).

\subsection{Learning and deployment}
\label{sec:rl}

The controller $\pi_\theta(b\mid s)$ is a two-layer MLP network that outputs a categorical distribution over budgets, while the backbone, indexer, and library stay frozen. Each episode $\tau$ ends with a verifier score $V(\tau)\in[0,1]$. Learning solves
\begin{equation}
\max_\theta J_V(\theta)\ \ \text{s.t.}\ \ J_{\overline C}(\theta)\leq\overline B,
\label{eq:cmdp}
\end{equation}
where $J_V$ and $J_{\overline C}$ are the expected score and expected normalized cost, through the Lagrangian with a compute price $\lambda\geq0$ (Figure~\ref{fig:pipeline}).

\paragraph{Warm start.}
Terminal reward is sparse, so training starts from a supervised estimate. We replay logged all-refresh continuations under every library pattern with teacher forcing, which yields each budget's increase in token loss and its normalized cost on the same text. Cross-fitted regressors predict these labels from the observation, and the controller is fitted to a soft target that trades predicted loss against predicted cost at the smallest price meeting $\overline B$ (Appendix~\ref{app:warm}).

\paragraph{Constrained policy gradient.}
Replay labels cannot show how a budget changes later text, tool calls, or success. The second stage therefore trains on-policy, crediting each decision with the terminal score minus the priced cost still to come,
\begin{equation}
G_u=V(\tau)-\lambda\textstyle\sum_{u'\geq u}\overline C_{u'},
\label{eq:policy-gradient}
\end{equation}
with a learned value baseline \citep{williams1992reinforce}. The price rises when a batch overspends the budget and falls otherwise, on a slower timescale than the policy \citep{borkar2005actor,tessler2019rcpo}. During training, every sampled budget runs as sampled, so the gradient credits actions that actually executed (Appendix~\ref{app:rl}).

Algorithm~\ref{alg:overview} summarizes training. \textsc{ExecuteSegment}$(u,\mathbf c)$ runs segment $u$ under mask $\mathbf c$: the queued prefill and up to $Q$ decoded tokens. Each query position passes through every layer and appends its key and value, and at eligible layers its index key, exactly once; Full layers write selected position IDs to $\Tcache$, and Shared layers read them and attend with their own key-value tensors. The helper returns the segment cost $C_u$, including controller overhead (Algorithm~\ref{alg:main} gives the query- and layer-level procedure).

\begin{algorithm}[t]
\caption{\method training (overview).}
\label{alg:overview}
\small
\begin{algorithmic}[1]
\Require frozen backbone, indexer, nested library $\{\mathbf c^{(b)}\}$; warm start $\theta_0$; budget $\overline B$
\State $\theta\gets\theta_0$; initialize critic $\psi$ and price $\lambda\geq0$
\Repeat
  \For{each episode $\tau$ in a batch $\mathcal K$}
    \For{each segment $u$ until $\tau$ ends}
      \State $s_u\gets\phi(\mathcal H_u)$ \Comment{completed computation only}
      \State sample $b_u\sim\pi_\theta(\cdot\mid s_u)$ \Comment{runs as sampled}
      \State $C_u\gets$ \Call{ExecuteSegment}{$u,\mathbf c^{(b_u)}$}
      \State if the response is complete and calls a tool, run it and queue its output
    \EndFor
    \State record the verifier reward $V(\tau)$
  \EndFor
  \State compute the returns $G_u$ (Eq.~\ref{eq:policy-gradient}) at the current $\lambda$
  \State one actor step along $\widehat g_\theta$
  \State regress the critic $v_\psi(s_u)$ onto $G_u$
  \State update $\lambda$ from the batch's overspend every $P_\lambda$ batches
  \State save the checkpoint if its position is predeclared
\Until{the rollout budget is exhausted}
\end{algorithmic}
\end{algorithm}

\paragraph{Deployment.}
The constraint holds only in expectation, so feasibility is decided on held-out validation tasks from repository owners that appear in no other partition. Every saved checkpoint and every static library entry receives simultaneous bounds on expected cost and success. The deployed controller is the feasible learned checkpoint with the largest success bound; if none is feasible, the best feasible static entry is deployed as a static fallback. This choice is frozen before final evaluation. A controller that always returns one budget recovers a static IndexCache schedule.

\section{Experimental Setup}
\label{sec:experiments}

\paragraph{Backbones.}
Every backbone is dense in attention as released and receives an injected indexer distilled from its own attention ($H_{\idx}=8$, $k=2048$), its own library, and its own controller. Qwen3-4B-Instruct-2507 is the primary model \citep{qwen2025instruct2507}; Qwen3-1.7B, Qwen3-8B, the original Qwen3-4B, Gemma-3-4B-it, and the mixture-of-experts MiniMax-M2 extend the study \citep{yang2025qwen3,gemma3report,minimax2026m2series}. All backbones run at a 32K cap, and those that support it also at 128K (Appendix~\ref{sec:models}).

\paragraph{Benchmarks and metrics.}
SWE-bench Verified (500 tasks) carries the primary comparison \citep{jimenez2024swebench,openai2024sweverified}; Terminal-Bench 2.0 (89) \citep{merrill2026terminalbench} and the retail (114), airline (50), and telecom (114) domains of $\tau^2$-bench \citep{barres2025tau2} test transfer. Success is the percentage of attempts solved; for $\tau^2$, each task runs four times and success is pass@1, the mean over trials, while pass$^4$ is the four-trial reliability, the share of tasks solved in all four trials. Time is mean end-to-end minutes per attempt, counting model, tool, environment, and user-simulator time for failed attempts too; minutes per successful task divides total time by the number of successes, and speedup divides the all-refresh time by a method's time.

\paragraph{Fitting data.}
Controllers are fitted on 37,968 repository-repair tasks from SWE-Gym and SWE-smith \citep{pan2025swegym,yang2025swesmith}, after removing every task whose repository owner also owns a SWE-bench Verified repository; owners are split into disjoint partitions, one per fitting stage (Appendix~\ref{app:splits}). Each backbone's repair controller is evaluated unchanged on all five benchmarks; separate in-domain studies train on subsets of the target benchmarks and report held-out subsets.

\paragraph{Comparators and cost matching.}
Native dense attention and all-refresh sparse execution bound what the injected sparse attention keeps. Static IndexCache runs one library pattern for the whole episode in a specialized runtime that omits index keys at permanently Shared layers. Three static selections have distinct roles: a development reference fixed before controller training; the deployed policy, chosen on validation as above; and the \emph{analysis comparator}, the static library entry with the highest success on the final sample among those within the cost tolerance. Because the analysis comparator sees the final sample, it favors the static side. The budget is $\overline B=0.50$ with tolerance $0.53$. Controls replace \method's ingredients one at a time (placement, state-free timing, rules, local learning, restricted state), and external selectors change how a selection is computed or stored (Appendix~\ref{app:comparisons}).

\section{Results}
\label{sec:results}

\begin{table*}[t]
\centering
\small
\setlength{\tabcolsep}{2.75pt}
\begin{tabular}{lrrrrrrrrrr}
\toprule
& \multicolumn{2}{c}{SWE-bench V.} & \multicolumn{2}{c}{Terminal-Bench} & \multicolumn{2}{c}{$\tau^2$ retail} & \multicolumn{2}{c}{$\tau^2$ airline} & \multicolumn{2}{c}{$\tau^2$ telecom} \\
\cmidrule(lr){2-3}\cmidrule(lr){4-5}\cmidrule(lr){6-7}\cmidrule(lr){8-9}\cmidrule(lr){10-11}
Method & Succ.$\uparrow$ & Min/s.$\downarrow$ & Succ.$\uparrow$ & Min/s.$\downarrow$ & Succ.$\uparrow$ & Min/s.$\downarrow$ & Succ.$\uparrow$ & Min/s.$\downarrow$ & Succ.$\uparrow$ & Min/s.$\downarrow$ \\
\midrule
\multicolumn{11}{l}{\textit{References}} \\
\quad \textcolor{black!55}{Native attention} & \textcolor{black!55}{32.0} & \textcolor{black!55}{71.5} & \textcolor{black!55}{28.1} & \textcolor{black!55}{67.3} & \textcolor{black!55}{57.7} & \textcolor{black!55}{20.8} & \textcolor{black!55}{50.0} & \textcolor{black!55}{19.1} & \textcolor{black!55}{46.5} & \textcolor{black!55}{27.8} \\
\quad All-refresh & 31.6 & 56.8 & \textbf{27.0} & 54.1 & 57.0 & 14.6 & 48.5 & 13.8 & 45.4 & 20.2 \\
\addlinespace[2pt]
\multicolumn{11}{l}{\textit{Static allocation}} \\
\quad Static ($R=18$) & 27.0 & 64.6 & 22.5 & 61.9 & 53.3 & 14.1 & 42.5 & 14.1 & 42.5 & 19.5 \\
\addlinespace[2pt]
\multicolumn{11}{l}{\textit{Alternative selectors}} \\
\quad Training-aware IndexCache & 30.4 & 56.3 & 23.6 & 58.8 & 56.4 & 13.3 & 49.0 & 12.2 & 45.0 & 18.4 \\
\quad LatentIndex & 31.6 & 55.1 & \textbf{27.0} & 52.3 & 56.4 & 14.1 & \textbf{49.5} & 12.8 & 45.2 & 19.4 \\
\quad LongCat CLI & 31.4 & 53.8 & \textbf{27.0} & \textbf{50.6} & \textbf{58.6} & \textbf{12.7} & 49.0 & 12.2 & 45.0 & 18.5 \\
\quad ReTopK (per head) & 29.8 & 60.0 & \textbf{27.0} & 52.6 & 56.6 & 14.0 & 47.5 & 13.4 & 46.1 & 19.0 \\
\addlinespace[2pt]
\multicolumn{11}{l}{\textit{{\upshape\method}-based methods}} \\
\rowcolor{AIRRow}\quad HISA + \method & 30.6 & 55.5 & 24.7 & 55.2 & 56.1 & 12.9 & 48.5 & \textbf{11.9} & \textbf{46.9} & \textbf{17.0} \\
\rowcolor{AIRRow}\quad \method & \textbf{32.4} & \textbf{52.7} & 24.7 & 56.4 & 57.2 & 13.1 & 46.5 & 12.9 & 46.5 & 17.9 \\
\bottomrule
\end{tabular}

\caption{\method-based methods have the highest success on two benchmarks and the lowest time per successful task on three, and \method improves over the matched static schedule on all five. Qwen3-4B-Instruct-2507, 128K; bold: best frozen-backbone value (ties included); tint: \method-based; gray: native attention, unranked.}
\label{tab:main}
\end{table*}

\begin{figure*}[t]
\centering
\includegraphics{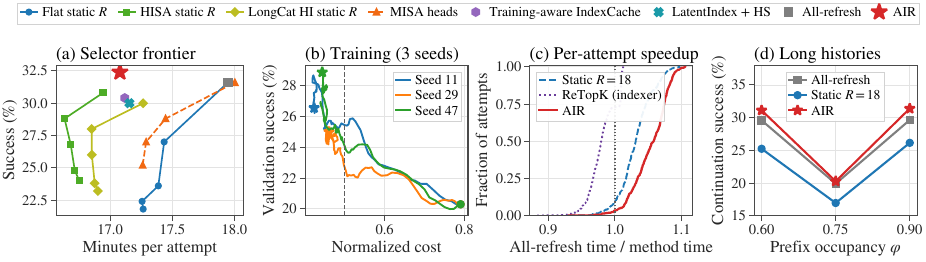}
\caption{\method on SWE-bench Verified and in training (Qwen3-4B-Instruct-2507, 128K). (a) Success against minutes per attempt for 24 selector settings. (b) Validation success against normalized cost for three seeds; dashed: the 0.50 budget. (c) All-refresh time divided by each method's time, per attempt. (d) Success on long-history continuations at prefix occupancy $\varphi$.}
\label{fig:evidence}
\end{figure*}

\subsection{Main comparison}
\label{sec:results-primary}

\paragraph{Higher success at matched cost.}
On the primary model at 128K, validation deploys a learned controller (Appendix~\ref{app:results-primary}). On SWE-bench Verified it solves 32.4\% of tasks, against 27.0\% for the analysis comparator, the static schedule $R=18$, at a normalized indexing and control cost of 0.524 against 0.502; both lie within the 0.53 tolerance. Across the five benchmarks, success rises by 2.2 to 5.4 points, and minutes per successful task fall on all five, from 64.6 to 52.7 on SWE-bench Verified (Table~\ref{tab:opening}). Time per attempt changes little: it is 2.0\% lower on SWE-bench Verified (1,025 against 1,046 seconds), so the gain per successful task comes mainly from solving more tasks.

\paragraph{Strong comparators remain visible.}
Table~\ref{tab:main} adds external selectors. \method has the highest success (32.4\%) and the lowest time per success (52.7 minutes) of the frozen-backbone methods on SWE-bench Verified, and HISA with \method leads telecom on both measures (46.9\%, 17.0 minutes) and airline on time per success (11.9 minutes). On Terminal-Bench 2.0, four settings, including all-refresh and LongCat cross-layer indexing, reach 27.0\% against \method's 24.7\%; LongCat cross-layer indexing leads retail, LatentIndex airline, and HISA with \method telecom. All-refresh beats \method on airline as well (48.5 against 46.5\%) while taking 11.2\% longer per attempt there, and 4.6\% longer on Terminal-Bench 2.0. All-refresh stays within 1.5 points of native dense attention, which takes 1.27 to 1.43 times as long per attempt.

\paragraph{Against the selector frontier.}
Figure~\ref{fig:evidence}a places \method among 24 settings on SWE-bench Verified that vary the refresh count of the flat, HISA, and LongCat hierarchical selectors or the number of active MISA heads. Static settings trade success for time along their curves; \method reaches 32.4\%, above every one of them, at 17.1 minutes per attempt. The Terminal-Bench 2.0 and airline exceptions above hold for this frontier as well (Appendix~\ref{app:results-selectors}).

\paragraph{Learning, per-attempt gains, and long histories.}
The deployed policy comes from a consistent training process: every seed of the primary controller moves from high cost to below the budget while its validation success rises, and it ends with a mixed allocation over the menu (Figure~\ref{fig:evidence}b). The time saving holds attempt by attempt: the median paired speedup over all-refresh is 1.06 on SWE-bench Verified, while the indexer adaptation of ReTopK is slower than all-refresh on most attempts (Figure~\ref{fig:evidence}c). In continuations that resume from shared snapshots once the prefix fills 60, 75, or 90\% of usable context, \method solves more of the remaining SWE-bench Verified work than the static comparator and all-refresh in every band (Figure~\ref{fig:evidence}d); across all five benchmarks, all-refresh scores higher in 5 of the 15 bands, including all three airline bands (Appendix~\ref{app:results-transfer}).

\subsection{What the gain depends on}
\label{sec:results-ablations}

\begin{table}[t]
\centering
\small
\setlength{\tabcolsep}{2.0pt}
\begin{tabular}{lrrrrr}
\toprule
Variant & SWE & Term. & Retail & Airl. & Tel. \\
\midrule
\multicolumn{6}{l}{\textit{Static and state-free}} \\
\quad Greedy mask, $R=18$ & 27.0 & 22.5 & 53.3 & 42.5 & 42.5 \\
\quad Staggered mask, $R=18$ & 24.0 & 21.3 & 50.4 & 40.5 & 40.1 \\
\quad State-free mixture & 27.0 & 22.5 & 52.9 & 42.5 & 42.5 \\
\multicolumn{6}{l}{\textit{Rules and local learning}} \\
\quad Event trigger & 29.4 & 23.6 & 53.3 & 48.0 & 41.7 \\
\quad Lagged staleness & 30.0 & 22.5 & 55.0 & 47.5 & 44.7 \\
\quad Supervised router & 30.4 & 23.6 & 56.6 & \textbf{49.5} & 46.1 \\
\quad Contextual bandit & 30.6 & 23.6 & 56.8 & 46.0 & 45.0 \\
\multicolumn{6}{l}{\textit{Restricted state}} \\
\quad Event metadata only & 28.2 & 23.6 & 56.6 & 48.0 & 45.0 \\
\quad Model + index stats & 31.4 & \textbf{24.7} & 56.6 & 47.5 & 44.5 \\
\midrule\rowcolor{AIRRow}\method (terminal reward) & \textbf{32.4} & \textbf{24.7} & \textbf{57.2} & 46.5 & \textbf{46.5} \\
\bottomrule
\end{tabular}

\caption{Among the variants shown, \method leads on SWE-bench Verified, retail, and telecom, ties the best Terminal-Bench 2.0 score, and the supervised router leads on airline. Success (\%), Qwen3-4B-Instruct-2507 at 128K, $\beta=0.50$; bold: best per column, ties included.}
\label{tab:ablation-main}
\end{table}

Table~\ref{tab:ablation-main} removes one ingredient at a time. \emph{Placement} matters at a fixed refresh count: seeded staggered masks lose 1.8 to 3.0 points on SWE-bench Verified to the greedy mask with the same $R$. \emph{Varying the budget without reading the state} adds nothing: a state-independent mixture scores 27.0\% and a fixed temporal schedule 27.2\%. Rules that read the state recover part of the gap (29.4\% for an event trigger, 30.0\% for lagged staleness), a supervised router reaches 30.4\%, and a contextual bandit trained on immediate feedback 30.6\%, so \emph{credit for delayed effects} adds 1.8 to 2.0 points over local learning. Restricting the controller's state costs 1.0 to 4.2 points, with event metadata alone losing most. Airline is the exception: five of six restricted or locally trained variants score 1.0 to 3.0 points above \method there. Across 96 factorial configurations, the budget has the largest effect, and shorter decision intervals and the warm start both raise success (Figure~\ref{fig:ablation-panels}); Figure~\ref{fig:factorial-main} in Appendix~\ref{app:results-ablations} gives every factor with its time per attempt.

\begin{figure}[t]
\centering
\includegraphics{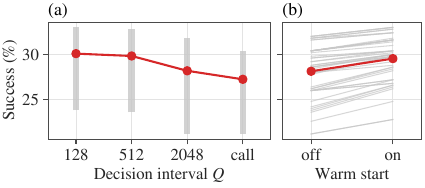}
\caption{Decision-interval and warm-start ablations on SWE-bench Verified (Qwen3-4B-Instruct-2507, 128K). Points are mean success over the 96 factorial configurations; gray marks show min--max ranges in (a) and the 48 matched pairs in (b).}
\label{fig:ablation-panels}
\end{figure}

\paragraph{An allocation example.}
\label{sec:results-case}
In one SWE-bench Verified repair, a visible test returns a new reversed-range diagnostic. The controller sees the test-result flag on the latest tool event and raises the budget from $R=3$ to $R=18$ at the next model call, while fixed IndexCache keeps $R=9$; both then read the same branch to revise. The example shows where each method spends index work; the success rates come from the aggregate results (Appendix~\ref{app:cases} gives this event and full traces, including one in which \method ends worse).

\subsection{Across backbones and context caps}
\label{sec:results-backbones}

\begin{figure}[t]
\centering
\includegraphics{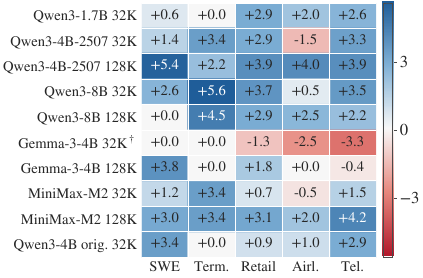}
\caption{\method's success minus the analysis comparator's (percentage points) for every backbone, cap, and benchmark; blue favors \method. Gemma-3-4B-it at 32K deploys the static fallback $R=2$.}
\label{fig:gains-main}
\end{figure}

Across the 50 backbone, cap, and benchmark settings, \method's success exceeds the comparator's in 37, ties in 7, and falls below it in 6 (Figure~\ref{fig:gains-main}). Four losses are on Gemma-3-4B-it, where only five of 34 layers are eligible: three are its 32K $\tau^2$ settings, where no learned controller met the validation cost bound and the static fallback is deployed, and one is 128K telecom ($-0.4$). The other two are 32K airline on the primary model ($-1.5$) and MiniMax-M2 ($-0.5$). The largest gain is on Qwen3-8B at 32K on Terminal-Bench 2.0 (+5.6). Time per attempt stays between 3.9\% longer and 7.0\% shorter than the comparator's. Training behaves alike on every backbone: normalized cost falls from 0.79 below the 0.50 budget within about 4,000 rollouts while validation success rises by 6.2 to 7.6 points at 128K, except Gemma-3-4B-it at 32K, whose cost stays at 0.59 (Appendix~\ref{app:results-training}).

\paragraph{Transfer and reliability.}
The repair controller transfers to terminal and customer-service tasks with the gains above, and training in the target domain adds 1.3 to 5.0 points on held-out subsets at equal time per attempt. On $\tau^2$, \method's four-trial reliability pass$^4$ exceeds the comparator's in five of six domain and cap settings and is lower on 32K airline (Appendix~\ref{app:results-transfer}).

\subsection{Where the time goes}
\label{sec:results-efficiency}

\begin{figure}[t]
\centering
\includegraphics{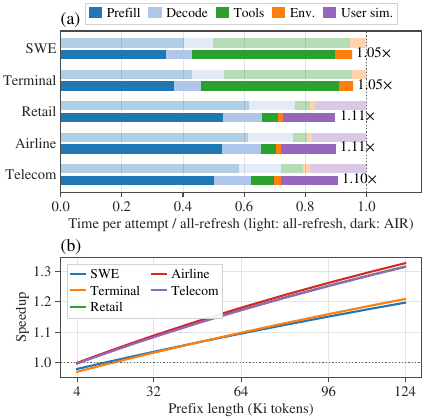}
\caption{Tools and the environment bound the whole-agent speedup, which grows with the prefix (Qwen3-4B-Instruct-2507, 128K). (a) Time per attempt relative to all-refresh; (b) \method's speedup against prefix length.}
\label{fig:efficiency-main}
\end{figure}

\method halves the indexing work of all-refresh execution, and the agent loop dilutes the saving (Figure~\ref{fig:efficiency-main}). Tools, the environment, and the user simulator take 52 to 55\% of \method's time on SWE-bench Verified and Terminal-Bench 2.0 and 27 to 31\% on $\tau^2$, so whole-agent speedups over all-refresh are 1.05 and 1.10 to 1.11. Model-side gains are larger: across backbones and caps on SWE-bench Verified, \method runs prefill 1.14 to 1.30 times and decoding 1.05 to 1.17 times as fast as all-refresh. The speedup grows from 0.97--1.00 at 4,096 tokens to 1.20--1.33 at 126,976 tokens, because selection work grows with the prefix while key maintenance and controller work stay constant per query. The design's own costs stay small: keeping every eligible layer ready for promotion costs up to 2.3\% more time and up to 12 times the index-cache memory at small refresh counts, and controller cost barely moves end-to-end time (Appendix~\ref{app:results-efficiency}).

\paragraph{Composition.}
With the flat indexer, learned allocation beats the selector's static schedule by 2.2 to 5.4 points; with HISA, four-head MISA, or LongCat hierarchical indexing, the change ranges from $-2.5$ to $+3.4$ points, so the benefit of learned allocation depends on the selector (Appendix~\ref{app:results-selectors}).

\section{Conclusion}
\label{sec:conclusion}

\method starts from an execution scheme that already works, IndexCache's Full/Shared reuse, and adds a decision about when to spend fresh selections. A small controller that reads completed computation and tool events, chooses among nested patterns, and learns from terminal reward under an expected compute budget improves success over the strongest static schedule at matched cost on every benchmark of the primary model and lowers time per successful task. The settings where a static schedule ties or wins, a backbone with few eligible layers and benchmarks where all-refresh still scores higher, mark where state-dependent allocation has the least to offer.

\section*{Limitations}

The gains depend on the backbone and the benchmark: on Gemma-3-4B-it, where five of 34 layers are eligible, validation deploys a static fallback at 32K, and all-refresh sparse attention still scores higher than \method on Terminal-Bench 2.0 and $\tau^2$ airline. The indexer is injected into dense checkpoints and may lose quality that a natively sparse model would keep. Index keys at every eligible layer cost memory that static IndexCache avoids. The finite library cannot reach patterns outside it, and the expected-cost constraint lets individual episodes exceed the budget. Tool and environment time bound the whole-agent speedup. Small benchmarks (50 airline tasks, 20 to 40 tasks in the in-domain subsets) limit precision, SWE-bench Verified concentrates nearly half of its instances in one repository, and each setting reports one deployed policy. Results use fixed benchmark releases and one user simulator (Appendix~\ref{app:limitations}).

\section*{Ethical Considerations}

\method reduces indexing compute for agents and so slightly lowers the energy needed per successful task; it does not change what an agent is allowed to do. The benchmarks use public repositories, terminal tasks, and simulated customers; no human participants or personal data were involved. The $\tau^2$ user simulator is a commercial model used under its terms. Faster agents can be deployed more widely, raising the usual concerns about automated code changes and customer-facing actions; the method inherits those risks from the underlying agent.

\bibliography{adaptive_indexing_references}

\clearpage

\appendix
\newcommand{\casefloat}[1]{\begin{figure*}[tbp]\begin{minipage}{\textwidth}\caseboxes\input{#1}\end{minipage}\end{figure*}}
\makeatletter
\renewenvironment{table}[1][tbp]{\@dblfloat{table}[#1]}{\end@dblfloat}
\renewenvironment{figure}[1][tbp]{\@dblfloat{figure}[#1]}{\end@dblfloat}
\makeatother
\renewcommand{\arraystretch}{0.94}
\makeatletter\setlength{\@dblfptop}{0pt}\setlength{\@dblfpbot}{0pt plus 1fil}\setlength{\@fptop}{0pt}\setlength{\@fpbot}{0pt plus 1fil}\makeatother
\setlength{\textfloatsep}{10pt plus 2pt minus 3pt}\setlength{\dbltextfloatsep}{10pt plus 2pt minus 3pt}
\section{Extended Background: Sparse Selection and Cross-Layer Reuse}
\label{app:ext-background}

\method builds on DSA-style dynamic sparse attention and on IndexCache's cross-layer reuse. For the objects the two methods share we adopt IndexCache's notation \citep{bai2026indexcache}. $N$ is the number of transformer layers, $L$ the sequence length, $\ell$ the layer index, and $t$ the position of the current query token. For that query, layer $\ell$'s indexer produces a score vector $\mathbf I^{(\ell)}_t$ with one entry per candidate position, and $\mathcal T^{(\ell)}_t$ holds the IDs of the $k$ highest-scoring candidates. The count $k$ refers to these selected candidates only; the sink and recent-window positions introduced below are attended in addition.

\paragraph{Query positions and candidate positions.}
Positions are numbered from 1, and the query at position $t$ is the token whose layer outputs are being computed: at each layer, its hidden state supplies the attention query vector, and the earlier positions supply keys and values from the layer's cache. During prefill, a block of newly appended tokens is processed together, and under the causal mask each of them is a query that sees only the positions up to itself. During incremental decoding, one token is generated at a time, and the query is the most recently appended token. Each query's own key and value are appended before attention, so the query at position $t$ sees exactly the positions $\{1,\ldots,t\}$, itself included, and $t$ equals the context length at that query.

Two small groups of positions are always attended without being scored. The sink set $\mathcal S$ is a fixed set of positions, chosen per backbone, that every query attends regardless of the ranking; it keeps a stable set of anchor positions in the attended support. A query can attend only positions that already exist, so at position $t$ the sinks in use are $\mathcal S_t=\mathcal S\cap\{1,\ldots,t\}$. The recent window $\mathcal W_t$ holds the last $w$ positions up to $t$, or all of $\{1,\ldots,t\}$ when $t\leq w$, and supplies the immediate local context. The indexer ranks the remaining positions, the candidate positions $\mathcal E_t=\{1,\ldots,t\}\setminus(\mathcal S_t\cup\mathcal W_t)$. Candidate positions are tokens of the prefix; Section~\ref{sec:indexcache-semantics} defines which transformer layers take part in index reuse.

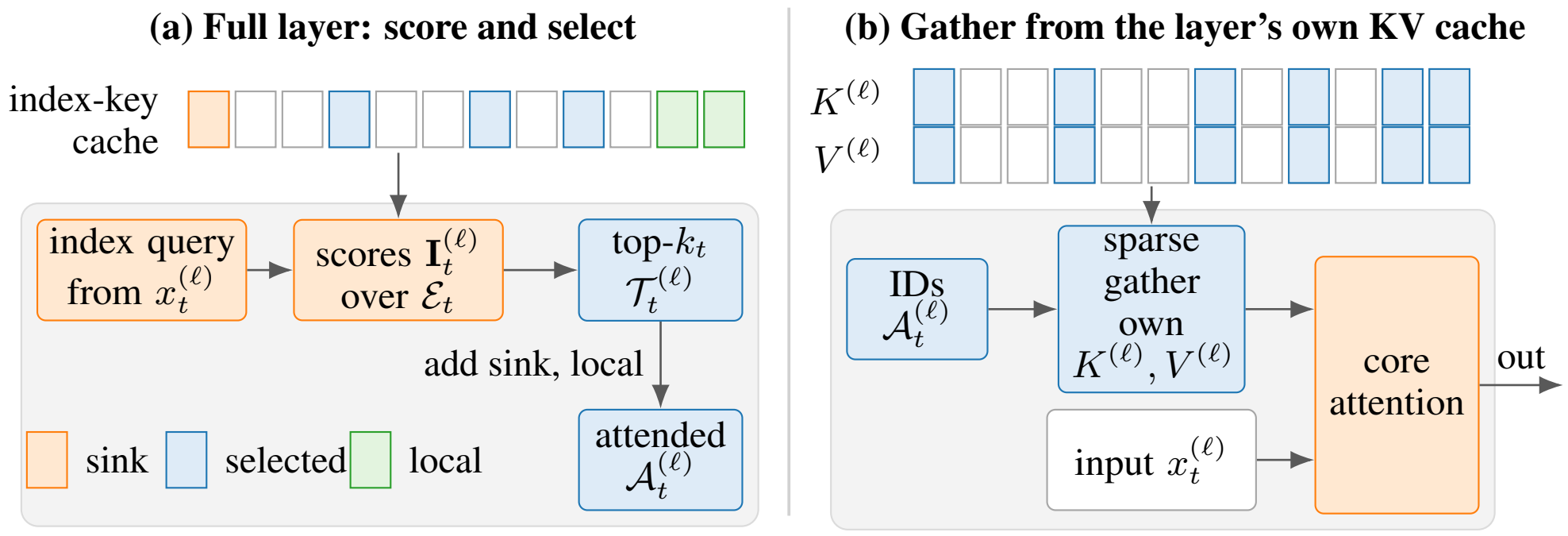
\begin{figure*}[tbp]
\centering
\begin{tikzpicture}[
  air figure, x=0.93cm, y=0.93cm,
  token/.style={draw=black!35, fill=white, minimum width=3.35mm, minimum height=4.65mm, inner sep=0pt},
  chosen/.style={token, draw=refreshblue, fill=softblue},
  sink/.style={token, draw=warmorange, fill=softorange},
  local/.style={token, draw=agentgreen, fill=softgreen},
  box/.style={draw=black!35, fill=white, rounded corners=2pt, align=center, minimum height=8.37mm, inner sep=1.8pt},
  bluebox/.style={box, draw=refreshblue, fill=softblue},
  orangebox/.style={box, draw=warmorange, fill=softorange},
  arrow/.style={-{Latex[length=2mm]}, draw=black!65, line width=0.5pt}
]
\node[air title] at (3.30,3.90) {(a) Full layer: score and select};
\node[anchor=east, align=right] at (1.35,3.10) {index-key\\cache};
\foreach \x in {0,...,11} {\node[token] at (1.65+0.42*\x,3.10) {};}
\node[sink] at (1.65,3.10) {};
\foreach \x in {3,6,8} {\node[chosen] at (1.65+0.42*\x,3.10) {};}
\foreach \x in {10,11} {\node[local] at (1.65+0.42*\x,3.10) {};}
\node[orangebox, text width=16.15mm] (query) at (1.05,1.75) {index query\\from $x^{(\ell)}_t$};
\node[orangebox, text width=16.15mm] (score) at (3.35,1.75) {scores $\mathbf I^{(\ell)}_t$\\over $\mathcal E_t$};
\node[bluebox, text width=12.35mm] (topk) at (5.70,1.75) {top-$k_t$\\$\mathcal T^{(\ell)}_t$};
\node[bluebox, text width=12.35mm] (ids) at (5.70,0.05) {attended\\$\mathcal A^{(\ell)}_t$};
\draw[arrow] (query) -- (score);
\draw[arrow] (score) -- (topk);
\draw[arrow] (topk) -- node[left] {add sink, local} (ids);
\draw[arrow] (3.35,2.80) -- (score.north);
\node[sink] at (0.20,0.05) {};
\node[anchor=west] at (0.40,0.05) {sink};
\node[chosen] at (1.45,0.05) {};
\node[anchor=west] at (1.65,0.05) {selected};
\node[local] at (3.10,0.05) {};
\node[anchor=west] at (3.30,0.05) {local};
\begin{scope}[on background layer]
\node[draw=black!14, fill=softgray, rounded corners=4pt, fit=(query)(score)(topk)(ids), inner sep=3.6pt] {};
\end{scope}
\draw[black!18, line width=0.7pt] (6.85,-0.45) -- (6.85,4.10);
\node[air title] at (10.40,3.90) {(b) Gather from the layer's own KV cache};
\node[anchor=east] at (7.85,3.30) {$K^{(\ell)}$};
\node[anchor=east] at (7.85,2.78) {$V^{(\ell)}$};
\foreach \x in {0,...,11} {
  \node[token] at (8.15+0.42*\x,3.30) {};
  \node[token] at (8.15+0.42*\x,2.78) {};
}
\foreach \x in {0,3,6,8,10,11} {
  \node[chosen] at (8.15+0.42*\x,3.30) {};
  \node[chosen] at (8.15+0.42*\x,2.78) {};
}
\node[bluebox, text width=10.45mm] (ids2) at (8.00,1.40) {IDs\\$\mathcal A^{(\ell)}_t$};
\node[bluebox, text width=14.25mm] (gather) at (10.10,1.40) {sparse gather\\own $K^{(\ell)},V^{(\ell)}$};
\node[box, text width=16.15mm] (coreq) at (10.10,0.05) {input $x^{(\ell)}_t$};
\node[orangebox, text width=12.35mm, minimum height=21.39mm] (coreattn) at (12.30,0.72) {core\\attention};
\draw[arrow] (ids2) -- (gather);
\draw[arrow] (10.10,2.50) -- (gather.north);
\draw[arrow] (gather) -- (gather -| coreattn.west);
\draw[arrow] (coreq) -- (coreq -| coreattn.west);
\draw[arrow] (coreattn.east) -- node[above] {out} ++(0.75,0);
\begin{scope}[on background layer]
\node[draw=black!14, fill=softgray, rounded corners=4pt, fit=(ids2)(gather)(coreattn)(coreq), inner sep=3.6pt] {};
\end{scope}
\end{tikzpicture}
\caption{Dynamic sparse attention at a Full layer: (a) the indexer scores the prefix and keeps the top-$k$ positions plus sink and local positions; (b) core attention gathers the layer's own keys and values there.}
\label{fig:dsa-mechanics}
\end{figure*}

\paragraph{Indexer scores.}
At layer $\ell$ the indexer compares the current query with every candidate position through $H_{\idx}$ small score heads. Each head $r$ projects the normalized layer input $x^{(\ell)}$ into an index query for the query position, $\widetilde{\mathbf q}^{(\ell,r)}_t=W^{(\ell,r)}_Q\operatorname{LN}(x^{(\ell)}_t)$, and into an index key for every earlier position $p$, $\widetilde{\mathbf k}^{(\ell,r)}_p=W^{(\ell,r)}_K\operatorname{LN}(x^{(\ell)}_p)$; tilded bold vectors denote these index-head projections, and the index keys are cached as tokens arrive. A head's score for position $p$ is the scaled dot product of its query and key, passed through a ReLU so that only positive matches contribute. A query-dependent softmax gate $\mathbf g^{(\ell)}_t$ over heads then weights the per-head scores and sums them into one ranking value per candidate position:
\begin{equation}
I^{(\ell)}_{t,p}=\sum_{r=1}^{H_{\idx}}g^{(\ell,r)}_t\operatorname{ReLU}\left(\langle\widetilde{\mathbf q}^{(\ell,r)}_t,\widetilde{\mathbf k}^{(\ell,r)}_p\rangle/\sqrt{d_{\idx}}\right).
\label{eq:index-score-ext}
\end{equation}
This score adapts the DSA lightning indexer; Appendix~\ref{app:execution} lists the differences. A layer that selects keeps the highest-ranked candidates,
\begin{equation}
\mathcal T^{(\ell)}_t=\TopK{k}\left(\mathbf I^{(\ell)}_t\right),
\label{eq:topk}
\end{equation}
where the operator ranges over $\mathcal E_t$: it returns the $k$ best positions, or all of $\mathcal E_t$ when fewer than $k$ exist, so it keeps $k_t=\min\{k,|\mathcal E_t|\}$ positions, and it breaks ties toward the smaller position. Core attention then reads the attended set $\mathcal A^{(\ell)}_t=\mathcal T^{(\ell)}_t\cup\mathcal S_t\cup\mathcal W_t$, the selected candidates together with the available sinks and the recent window, from the layer's own key-value cache. A position that is both a sink and in the window is attended once, so $|\mathcal A^{(\ell)}_t|=k_t+|\mathcal S_t\cup\mathcal W_t|$. For a fixed number of attended positions, the core-attention work per query is the same whichever layer chose them. Figure~\ref{fig:dsa-mechanics} shows the two steps at one Full layer: scoring and top-$k$ selection, then the gather of the layer's own keys and values at the selected, sink, and recent positions.

\subsection{IndexCache execution semantics}
\label{sec:indexcache-semantics}

Running the indexer at every layer repeats much of the same ranking work, because adjacent layers of DSA models select largely overlapping positions. IndexCache therefore lets only some layers rank. Backbones differ in which layers can use this kind of attention, so we first name the layers that take part in index reuse. The eligible layers form an ordered set $\Lidx=\{\ell_1<\ell_2<\cdots<\ell_{\Nidx}\}\subseteq\{1,\ldots,N\}$ of size $\Nidx$. For the Qwen backbones every layer is eligible, so $\Lidx=\{1,\ldots,N\}$; for a model that interleaves sliding-window and global attention, only the global layers are eligible (Section~\ref{sec:models}). A layer pattern assigns a role $c_\ell\in\{\Full,\Shared\}$ to each eligible layer. A Full layer runs the indexer and selects positions with Equation~\ref{eq:topk}; a Shared layer reuses the positions of the nearest preceding eligible Full layer,
\begin{equation}
\begin{aligned}\mathcal T^{(\ell)}_t&=\mathcal T^{(f(\ell))}_t,\\ f(\ell)&=\max\{\ell'\in\Lidx:\ell'<\ell,\ c_{\ell'}=\Full\}.\end{aligned}
\label{eq:indexcache-ext}
\end{equation}
As in IndexCache, the first eligible layer $\ell_1$ is always Full, so every query receives at least one fresh selection; we write the maximization index as $\ell'$ because $j$ denotes agent steps here.

\begin{figure*}[ht]
\centering
\begin{tikzpicture}[
  air figure, x=0.93cm, y=0.93cm,
  layer/.style={draw=black!35, fill=white, rounded corners=2pt, minimum width=7.91mm, minimum height=7.44mm, align=center},
  full/.style={layer, draw=refreshblue, fill=softblue},
  shared/.style={layer, draw=reusegray, fill=white},
  reg/.style={draw=refreshblue, fill=softblue, rounded corners=2pt, minimum width=13.95mm, minimum height=6.51mm, align=center},
  op/.style={draw=black!35, fill=softgray, rounded corners=2pt, minimum height=8.37mm, align=center, inner sep=1.8pt},
  arrow/.style={-{Latex[length=1.8mm]}, draw=black!65, line width=0.45pt},
  readarrow/.style={arrow, draw=agentgreen}
]
\node[air title] at (3.15,3.70) {(a) Nearest-Full index buffer $\Tcache$};
\node[full] (f1) at (0.45,2.75) {\Full$_1$};
\node[shared] (s2) at (1.57,2.75) {\Shared$_2$};
\node[shared] (s3) at (2.69,2.75) {\Shared$_3$};
\node[full] (f4) at (3.81,2.75) {\Full$_4$};
\node[shared] (s5) at (4.93,2.75) {\Shared$_5$};
\node[shared] (s6) at (6.05,2.75) {\Shared$_6$};
\foreach \a/\b in {f1/s2,s2/s3,s3/f4,f4/s5,s5/s6} {\draw[arrow] (\a) -- (\b);}
\node[reg, minimum width=12.09mm] (r1) at (2.13,1.35) {$\mathcal T^{(1)}_t$};
\node[reg, minimum width=12.09mm] (r4) at (5.49,1.35) {$\mathcal T^{(4)}_t$};
\draw[arrow, draw=refreshblue] (f1.south) |- node[pos=0.25, right] {write} (r1.west);
\draw[readarrow] (r1.north -| s2) -- (s2.south);
\draw[readarrow] (r1.north -| s3) -- (s3.south);
\draw[arrow, draw=refreshblue] (f4.south) |- node[pos=0.25, right] {rewrite} (r4.west);
\draw[readarrow] (r4.north -| s5) -- (s5.south);
\draw[readarrow] (r4.north -| s6) -- (s6.south);
\node[anchor=north, align=center] at (3.25,0.70) {Full layers score the prefix;\\Shared layers read the latest position IDs.};
\draw[black!18, line width=0.7pt] (6.78,0.05) -- (6.78,4.00);
\node[air title] at (10.35,3.70) {(b) Shared positions, private KV tensors};
\node[full, text width=16.15mm, inner sep=1pt] (fl) at (7.92,2.85) {Full $\ell$\\append keys};
\node[op, text width=9.5mm] (idxop) at (9.62,2.85) {score\\prefix};
\node[reg, minimum width=9.3mm] (ir) at (10.95,2.85) {$\mathcal T^{(\ell)}_t$};
\node[op, text width=16.15mm] (g1) at (12.85,2.85) {attend own\\$(K,V)^{(\ell)}$};
\draw[arrow] (fl) -- (idxop);
\draw[arrow] (idxop) -- (ir);
\draw[arrow] (ir) -- (g1);
\node[shared, text width=16.15mm, inner sep=1pt] (sl) at (7.92,1.25) {Shared $\ell{+}1$\\append keys};
\node[op, text width=9.5mm] (skip) at (9.62,1.25) {skip\\scoring};
\node[reg, minimum width=9.3mm] (ir2) at (10.95,1.25) {$\mathcal T^{(\ell)}_t$};
\node[op, text width=16.15mm] (g2) at (12.85,1.25) {attend own\\$(K,V)^{(\ell+1)}$};
\draw[arrow] (sl) -- (skip);
\draw[readarrow] (ir) -- node[right] {reuse} (ir2);
\draw[arrow] (ir2) -- (g2);
\node[anchor=north, align=center] at (10.35,0.35) {Layers share position IDs and keep\\their own persistent KV state.};
\end{tikzpicture}
\caption{IndexCache execution inherited by \method: Full layers overwrite the position buffer $\Tcache$, Shared layers reuse it, and \method appends index keys at every eligible layer so any of them can later become Full.}
\label{fig:indexcache-semantics}
\end{figure*}
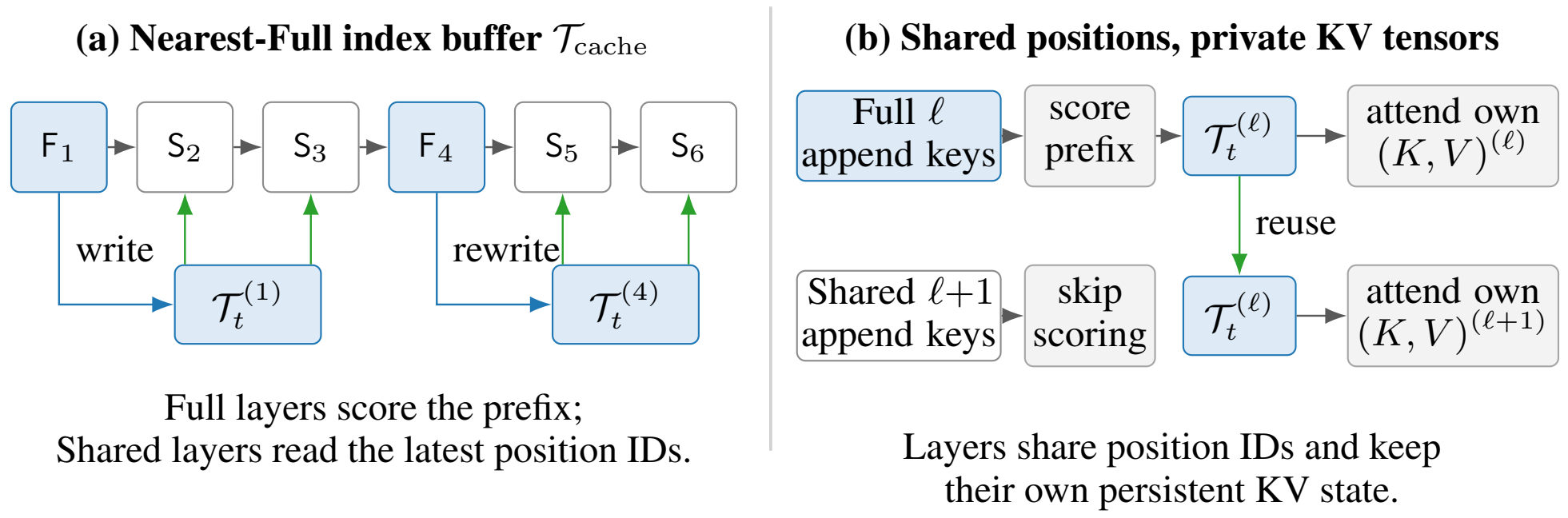

\paragraph{The position buffer and the persistent caches.}
Two kinds of state are involved, and they have different lifetimes. The selected positions travel through a temporary buffer $\Tcache$ that lives for one query position: each Full layer overwrites it, each Shared layer reads it, and it is released after the query's last layer. With five eligible layers and the pattern F, S, S, F, S, layer $\ell_1$ writes $\mathcal T^{(\ell_1)}_t$, layers $\ell_2$ and $\ell_3$ attend over $\mathcal T^{(\ell_1)}_t$, layer $\ell_4$ overwrites the buffer with $\mathcal T^{(\ell_4)}_t$, and layer $\ell_5$ attends over $\mathcal T^{(\ell_4)}_t$. The next query starts again at $\ell_1$, so every generated token makes its own selections. Layers outside $\Lidx$ keep their native attention and neither read nor overwrite the buffer. The persistent caches grow at every token and are never recomputed when the pattern changes: each layer keeps its own key-value cache, and each eligible layer keeps a cache of index keys. Reuse therefore shares only position IDs. Each layer gathers its own keys and values at the shared positions, so a Shared layer's output still depends on its own key-value tensors (Figure~\ref{fig:indexcache-semantics}).

\paragraph{Shared execution and later promotion.}
A Shared layer skips the query-time part of the indexer: the index query, the scoring of candidates, and the top-$k$ selection. In \method every eligible layer still projects and stores its index key for each new token, whatever its current role. This upkeep lets a layer that is Shared now become Full at a later segment: its index-key cache already covers the whole prefix, so it can score past positions without replaying past hidden states. Each cached entry was computed from the layer input when its token was processed, and that input depended on the selections made under the pattern then in force; this is the usual causal semantics of incremental decoding. A specialized static runtime whose pattern never changes can omit index keys at the layers it keeps Shared. \method keeps them at every eligible layer, and its cost model charges this upkeep (Section~\ref{sec:cost}). Table~\ref{tab:indexcache-comparison} sets out what \method shares with IndexCache and what it adds.

\begin{table*}[ht]
\centering
\small
\caption{What \method shares with IndexCache and what it adds. Both methods execute the same Full and Shared layer roles.}
\label{tab:indexcache-comparison}
\begin{tabular}{p{0.5\columnwidth}p{0.35\columnwidth}p{0.5\columnwidth}}
\toprule
Axis & IndexCache & \method \\
\midrule
Layer action & Full or Shared & Full or Shared \\
Patterns & One deployed pattern & One pattern per budget, nested \\
Decision state & None (global) & Current agent segment \\
Pattern selection & Calibration LM loss & Calibration LM loss \\
Controller training & None & NLL-regret warm start, then terminal reward and cost \\
Compute target & Fixed retention ratio & Expected trajectory budget \\
Index keys at Shared layers & Not stored & Stored at every eligible layer \\
Controller overhead & None & Included in the compute cost \\
Relation & Static schedule & Constant policy recovers it \\
\bottomrule
\end{tabular}
\end{table*}

\section{Method Details}
\label{app:ext-method}

This appendix adds to Section~\ref{sec:method} the details that the main text states only in outline.

\begin{figure*}[ht]
\centering
\begin{tikzpicture}[
  air figure, x=0.93cm, y=0.93cm,
  box/.style={draw=black!35, fill=white, rounded corners=3pt, minimum height=8.37mm, align=center, inner sep=1.8pt},
  bluebox/.style={box, draw=refreshblue, fill=softblue},
  greenbox/.style={box, draw=agentgreen, fill=softgreen},
  orangebox/.style={box, draw=warmorange, fill=softorange},
  violetbox/.style={box, draw=violet, fill=softviolet},
  layer/.style={draw=black!35, fill=white, rounded corners=1pt, minimum width=7.44mm, minimum height=6.51mm, align=center},
  full/.style={layer, draw=refreshblue, fill=softblue},
  shared/.style={layer, draw=reusegray, fill=white},
  kv/.style={draw=agentgreen!65, fill=softgreen, minimum width=5.12mm, minimum height=4.19mm, inner sep=0pt},
  ik/.style={draw=violet!65, fill=softviolet, minimum width=5.12mm, minimum height=4.19mm, inner sep=0pt},
  arrow/.style={-{Latex[length=2mm]}, line width=0.5pt, draw=black!65}
]
\node[anchor=west, air title] at (0.00,6.95) {Agent loop: one decision per segment};
\node[greenbox, text width=15.2mm] (obs) at (0.85,5.70) {observation\\$o_j$};
\node[bluebox, text width=13.3mm] (feat) at (3.20,5.70) {causal state\\$s_u$};
\node[violetbox, text width=13.3mm] (policy) at (5.65,5.70) {controller\\$b_u\sim\pi_\theta$};
\node[orangebox, text width=13.3mm] (library) at (8.10,5.70) {mask library\\$\mathbf c_u$};
\node[bluebox, text width=14.25mm] (model) at (10.55,5.70) {model\\segment $u$};
\node[greenbox, text width=15.2mm] (action) at (12.75,5.70) {tool action\\or text};
\draw[arrow] (obs) -- (feat);
\draw[arrow] (feat) -- (policy);
\draw[arrow] (policy) -- (library);
\draw[arrow] (library) -- (model);
\draw[arrow] (model) -- (action);
\draw[arrow, draw=agentgreen] (action.north) -- ++(0,0.32) -| node[pos=0.12, above] {environment transition} (obs.north);

\node[anchor=west, air title] at (0.00,4.62) {Token time: one mask per segment};
\foreach \x/\lab in {0/$y_1$,1/$y_2$,2/$\cdots$,3/$y_Q$} {
  \node[bluebox, minimum width=11.16mm] (tok\x) at (0.75+1.60*\x,3.10) {\lab};
}
\foreach \a/\b in {tok0/tok1,tok1/tok2,tok2/tok3} {\draw[arrow] (\a) -- (\b);}
\draw[decorate, decoration={brace,mirror,amplitude=4pt}, draw=violet] (0.10,2.48) -- (6.20,2.48)
  node[midway, below=4pt, text=violet] {one mask for the prefill and next $Q$ decode tokens};
\draw[arrow, dashed, draw=refreshblue] (model.south) -- (10.55,4.30) -| (tok2.north);
\node[anchor=west] at (8.95,3.85) {append-only caches};
\foreach \x in {0,...,6} {
  \node[kv] at (9.30+0.62*\x,3.30) {};
  \node[ik] at (9.30+0.62*\x,2.70) {};
}
\draw[arrow, draw=agentgreen] (tok3.east) -- (7.05,3.10) |- node[pos=0.75, above] {core KV} (8.95,3.30);
\draw[arrow, draw=violet] (7.05,3.10) |- node[pos=0.75, below] {index keys} (8.95,2.70);

\node[anchor=west, air title] at (0.00,1.55) {Depth: one query position};
\node[full] (l1) at (0.75,0.95) {$\Full$};
\node[shared] (l2) at (1.95,0.95) {$\Shared$};
\node[shared] (l3) at (3.15,0.95) {$\Shared$};
\node[full] (l4) at (4.45,0.95) {$\Full$};
\node[shared] (l5) at (5.65,0.95) {$\Shared$};
\foreach \a/\b in {l1/l2,l2/l3,l3/l4,l4/l5} {\draw[arrow] (\a) -- (\b);}
\node[bluebox, text width=19mm, minimum height=6.51mm] (r1) at (2.55,-0.25) {$\Tcache$ from\\first Full};
\node[bluebox, text width=19mm, minimum height=6.51mm] (r2) at (5.95,-0.25) {$\Tcache$ from\\next Full};
\draw[arrow, draw=refreshblue] (l1.south) |- (r1.west);
\draw[arrow, draw=agentgreen] (r1.north -| l2) -- (l2.south);
\draw[arrow, draw=agentgreen] (r1.north -| l3) -- (l3.south);
\draw[arrow, draw=refreshblue] (l4.south) |- (r2.west);
\draw[arrow, draw=agentgreen] (r2.north -| l5) -- (l5.south);
\node[orangebox, text width=47.5mm] at (10.60,0.45) {core KV and index keys persist across tokens;\\$\Tcache$ persists across depth\\for one query position only};
\end{tikzpicture}
\caption{Three execution timescales: one library mask per segment of the agent loop, cache appends per query within a segment, and the $\Tcache$ buffer per query across depth.}
\label{fig:timescales}
\end{figure*}
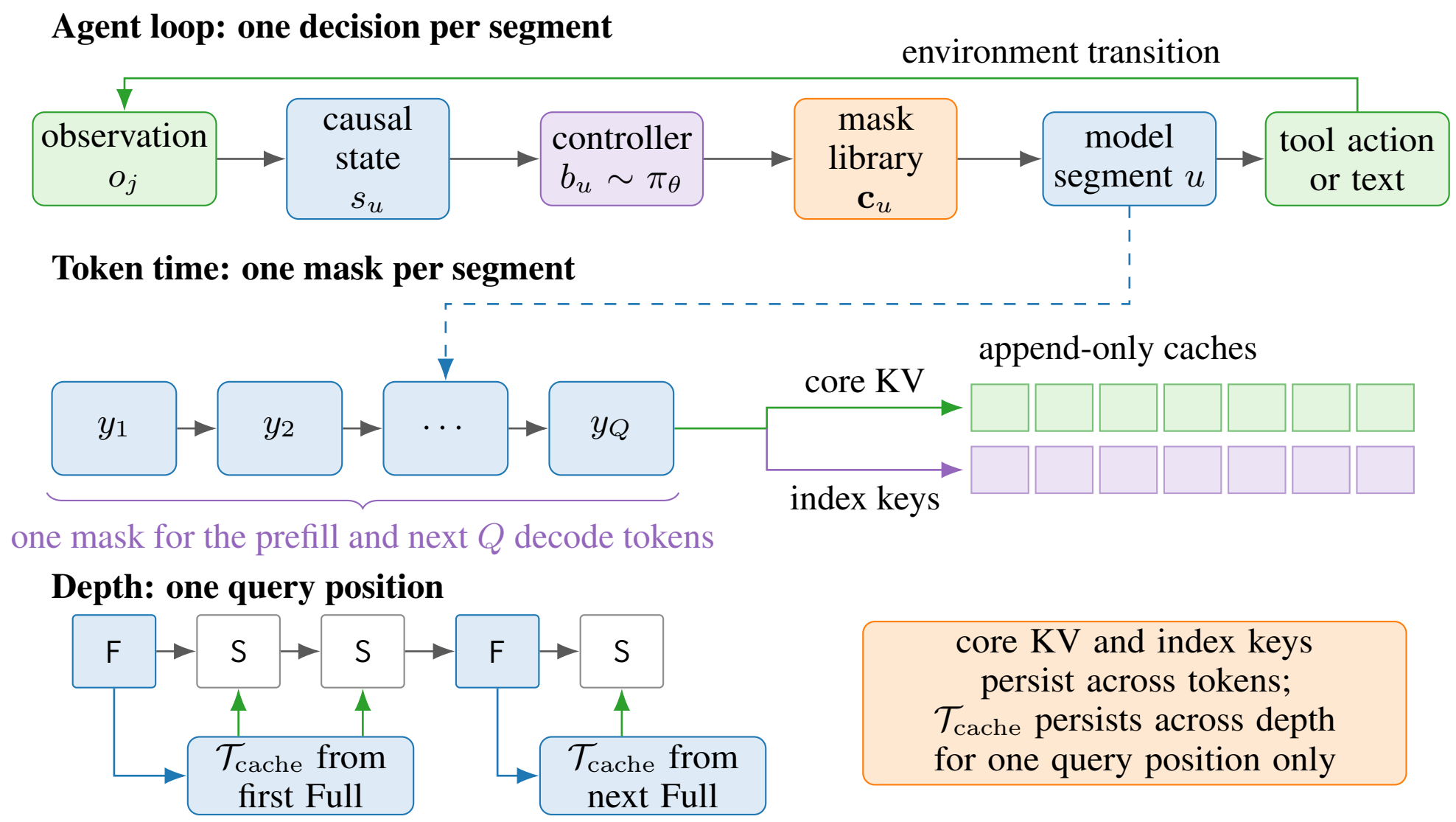

\paragraph{Segments.}
\label{app:ext-execution}
An agent episode alternates between environment observations $o_j$ and model responses $z_j$, where $j$ indexes agent steps. Suppose a test run returns a 1,200-token report and the model answers with 1,300 tokens. The new invocation opens segment $u$, which covers the prefill of the 1,200 report tokens and the first 512 decoded tokens; segments $u+1$ and $u+2$ cover the next 512 and the remaining 276 decoded tokens. Every prefilled or decoded token is a query position $t$ that passes through all $N$ layers $\ell$, so the decision operates on three nested timescales (Figure~\ref{fig:timescales}).

\paragraph{Menus.}
Each budget $b$ names a library pattern $\mathbf c^{(b)}$ with $R_b$ Full eligible layers (Figure~\ref{fig:library}). The primary backbone, with $\Nidx=36$ eligible layers, uses $R_b\in\{3,5,9,18,36\}$; other backbones scale these counts with
\begin{equation}
\begin{aligned}\mathcal R_m=\big\{&\max\left(1,\lceil \Nidx(m)\,r/36\rceil\right):\\ &r\in\{3,5,9,18,36\}\big\},\end{aligned}
\label{eq:menu-rule}
\end{equation}
taken as a set, so duplicate counts collapse before the controller's output size $|\Bset_m|=|\mathcal R_m|$ is fixed. The rule keeps all-refresh as the largest entry; Table~\ref{tab:models} reports the realized fractions $R_b/\Nidx$.

\paragraph{Paid probe.}
\label{sec:observations}
Statistics of the newly appended tokens, such as the indexer's score entropy on them, would require selector work. A paid-probe ablation buys them: before the decision, it runs the frozen indexer of the first eligible layer $\ell_1$, whose input does not depend on the budget, on the new tokens in temporary buffers, keeps three summary statistics, leaves the persistent caches untouched, and charges all of its computation and memory traffic to the controller (Appendix~\ref{app:features}).

\paragraph{Controller size.}
The actor has a hidden layer of width $h$ on the $d_s$-dimensional observation and one output per budget, so with biases it has
\begin{equation}
P_{\mathrm{actor}}=(d_s+1)h+(h+1)m
\label{eq:actor-params}
\end{equation}
parameters, where $m=|\Bset_m|$: five budgets for the Qwen backbones and MiniMax-M2 and four for Gemma. With $d_s=h=128$, the actor has 17,157 parameters (17,028 for Gemma), and a critic of the same width with one scalar output has 16,641. The critic and the two warm-start predictors are used during training only; the frozen feature projections run at deployment and are charged to $C_{\ctrl,u}$. The controller's size does not grow with context length, but the indexer state does: each eligible layer stores $H_{\idx}d_{\idx}$ index-key values per token, $H_{\idx}d_{\idx}/2048$ of the layer's core key-value cache, or one quarter for $H_{\idx}=8$ and $d_{\idx}=64$.

\paragraph{Policy-gradient estimator.}
\label{app:ext-rl}
\label{app:ext-cost}
The controller samples budgets at a fixed temperature $\tau_\pi$, and the critic $v_\psi(s)$ serves only as a baseline. With the backbone and indexer frozen, only the controller's sampling probabilities depend on $\theta$, so the likelihood-ratio identity gives an unbiased gradient
\begin{equation}
\begin{aligned}\widehat g_\theta=\frac{1}{|\mathcal K|}\sum_{\tau\in\mathcal K}\sum_{u}&\nabla_\theta\log\pi_\theta(b_u\mid s_u)\\ &\times\left(G_u-v_{\psi}(s_u)\right)\end{aligned}
\label{eq:policy-gradient-ext}
\end{equation}
over a batch $\mathcal K$ of on-policy episodes, with $G_u$ from Equation~\ref{eq:policy-gradient}. Costs incurred before decision $u$ cannot depend on $b_u$, which justifies summing cost from $u$. The actor takes one step per batch, the critic then regresses $v_\psi(s_u)$ onto $G_u$, and every $P_\lambda$ batches the price takes the projected step
\begin{equation}
\lambda\leftarrow\left[\lambda+\eta_\lambda\left(\tfrac{1}{|\mathcal K|}\textstyle\sum_{\tau\in\mathcal K}\overline C(\tau)-\overline B\right)\right]_+ ,
\label{eq:dual-update}
\end{equation}
so cost weighs more in $G_u$ after an overspending batch. The runtime executes every sampled budget as sampled, because the estimator would otherwise credit an action that never ran. If no static policy passes the validation bounds either, the operating point is declared infeasible and no policy is deployed at that budget (Appendix~\ref{app:rl}). Algorithm~\ref{alg:main} summarizes execution and training.

\begin{figure*}[ht]
\centering
\begin{tikzpicture}[
  air figure, x=0.93cm, y=0.93cm,
  state/.style={draw=black!25, fill=softgray, rounded corners=3pt, minimum height=7.44mm, align=center},
  full/.style={draw=refreshblue, fill=softblue, rounded corners=0.5pt, minimum width=1.86mm, minimum height=4.28mm, inner sep=0pt},
  shared/.style={draw=reusegray, fill=white, rounded corners=0.5pt, minimum width=1.86mm, minimum height=4.28mm, inner sep=0pt},
  arrow/.style={-{Latex[length=1.8mm]}, draw=black!60}
]
\node[air title] at (4.20,0.70) {nested masks over $\Nidx=36$ eligible layers, primary model};
\foreach \row/\r in {0/36,1/18,2/9,3/5,4/3} {
  \node[anchor=east] at (-0.20,-0.62*\row) {$R=\r$};
  \foreach \x in {0,...,35} {\node[shared] at (0.24*\x,-0.62*\row) {};}
}
\foreach \x in {0,...,35} {\node[full] at (0.24*\x,0) {};}
\foreach \x in {0,1,3,5,7,9,11,13,15,17,19,21,23,25,27,29,32,35} {\node[full] at (0.24*\x,-0.62) {};}
\foreach \x in {0,3,7,11,15,19,25,29,35} {\node[full] at (0.24*\x,-1.24) {};}
\foreach \x in {0,7,15,25,35} {\node[full] at (0.24*\x,-1.86) {};}
\foreach \x in {0,15,35} {\node[full] at (0.24*\x,-2.48) {};}
\node[anchor=north] at (0.00,-2.78) {$\ell_1$};
\node[anchor=north] at (8.40,-2.78) {$\ell_{36}$};
\node[state, text width=27mm] (eval) at (10.80,-0.45) {try each Full layer\\except $\ell_1$ as Shared};
\node[state, text width=27mm] (commit) at (10.80,-1.80) {commit the removal\\with the lowest loss};
\draw[arrow] (eval) -- (commit);
\draw[arrow] (commit.east) -- ++(0.25,0) |- (eval.east);
\node[full] at (9.75,-2.90) {};
\node[anchor=west] at (9.90,-2.90) {Full};
\node[shared] at (11.00,-2.90) {};
\node[anchor=west] at (11.15,-2.90) {Shared};
\end{tikzpicture}
\caption{Offline library construction: one greedy elimination path from all-refresh to aggressive reuse supplies a nested mask for each refresh count in the menu.}
\label{fig:library}
\end{figure*}
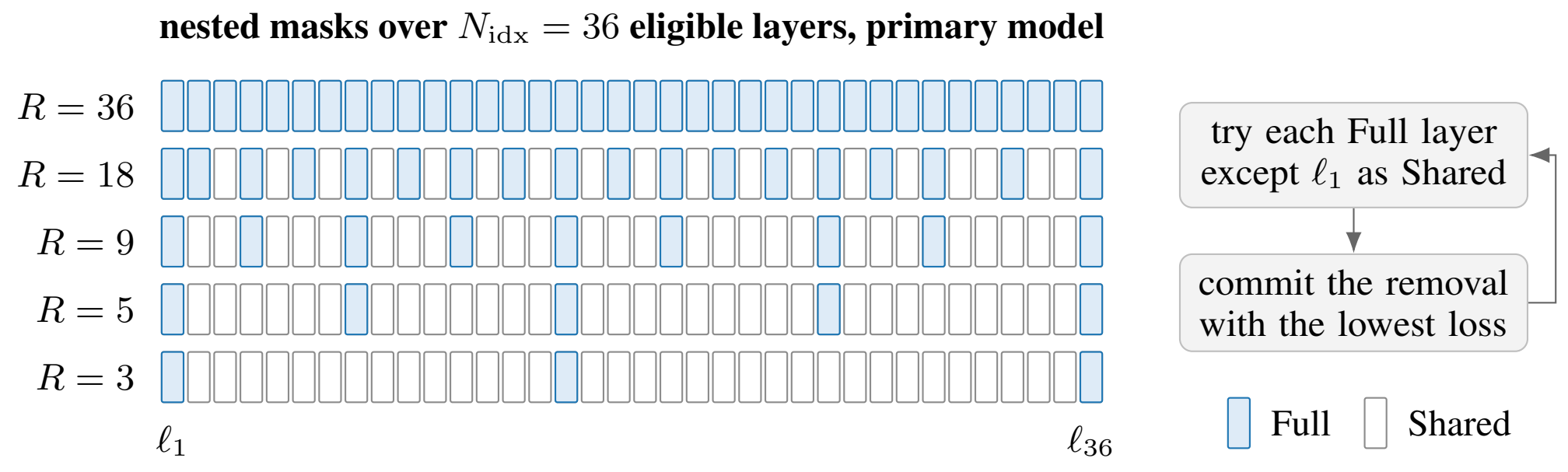

\begin{algorithm*}[tbp]
\caption{\method: segment-level budget selection inside constrained policy-gradient training}
\label{alg:main}
\small
\begin{algorithmic}[1]
\Require backbone $m$ with eligible layers $\Lidx$, frozen indexer, nested library $\{\mathbf c^{(b)}\}_{b\in\Bset_m}$, warm start $\theta_0$, $\overline B$
\Ensure predeclared controller checkpoints for validation-based selection
\Procedure{Forward}{$t,\mathbf c$} \Comment{one query position under pattern $\mathbf c$}
  \For{$\ell=1,\ldots,N$}
    \State Append layer $\ell$'s key and value for $t$
    \If{$\ell\notin\Lidx$}
      \State Run the layer's native attention \Comment{$\Tcache$ untouched}
    \Else
      \State Append layer $\ell$'s index key for $t$
      \If{$c_\ell=\Full$} \Comment{always at $\ell_1$}
        \State Select $\mathcal T^{(\ell)}_t$ (Eqs.~\ref{eq:index-score}--\ref{eq:topk}); $\Tcache\gets\mathcal T^{(\ell)}_t$
      \Else
        \State $\mathcal T^{(\ell)}_t\gets\Tcache$ \Comment{from layer $f(\ell)$}
      \EndIf
      \State Attend over $\mathcal A^{(\ell)}_t=\mathcal T^{(\ell)}_t\cup\mathcal S_t\cup\mathcal W_t$ with layer $\ell$'s own keys and values
    \EndIf
  \EndFor
  \State Release $\Tcache$
\EndProcedure
\State $\theta\gets\theta_0$; initialize the critic $\psi$ and the price $\lambda$
\Repeat
  \For{each episode $\tau$ of a batch $\mathcal K$}
    \For{each segment $u$ until the episode ends}
      \State Build $s_u$ from completed computation; sample $b_u\sim\pi_\theta(\cdot\mid s_u)$
      \State \Call{Forward}{$t,\mathbf c^{(b_u)}$} for each query position $t$ of the segment
      \State Add the segment cost $C_u$ (Section~\ref{sec:cost}); run the tool call if the response ended in one
    \EndFor
    \State Record the verifier score $V(\tau)$
  \EndFor
  \State Compute $G_u$; take one actor step along $\widehat g_\theta$ (Eq.~\ref{eq:policy-gradient}); fit the critic to $G_u$
  \State Update $\lambda$ every $P_\lambda$ batches (Eq.~\ref{eq:dual-update}); save checkpoints at predeclared positions
\Until{the rollout budget is exhausted}
\end{algorithmic}
\end{algorithm*}

\section{Notation}
\label{app:notation}

Please check Table~\ref{tab:notation} for full details.

\begin{table*}[ht]
\centering
\small
\caption{Core notation for agentic adaptive index reuse.}
\label{tab:notation}
\begin{tabular}{lp{0.78\textwidth}}
\toprule
Symbol & Meaning \\
\midrule
$N$ & Number of decoder layers ($N=36$ for the primary backbone) \\
$\Lidx$, $\Nidx$, $\ell_1$ & Ordered set of eligible layers, its size, and its first layer, which is always Full \\
$L$ & Sequence length \\
$\ell,\,j,\,u,\,t,\,p$ & Indices for layers, agent steps, controller decisions, query positions, and key positions \\
$k$, $k_t$ & Selected positions per query, and $k_t=\min\{k,|\mathcal E_t|\}$ \\
$\mathcal S$, $\mathcal S_t$ & Fixed sink positions, and those available at query $t$: $\mathcal S_t=\mathcal S\cap\{1,\ldots,t\}$ \\
$\mathcal W_t$ & Recent window: the last $w$ positions up to $t$ \\
$\mathcal E_t$ & Candidate positions: $\{1,\ldots,t\}\setminus(\mathcal S_t\cup\mathcal W_t)$ \\
$\mathbf I^{(\ell)}_t$ & Indexer score vector at layer $\ell$ for query position $t$ \\
$\mathcal T^{(\ell)}_t$ & Selected top-$k$ positions (at most $k_t=\min\{k,|\mathcal E_t|\}$) used by layer $\ell$ for query position $t$ \\
$\widetilde{\mathbf q}^{(\ell,r)}_t$, $\widetilde{\mathbf k}^{(\ell,r)}_p$ & Index query and index key of head $r$ (tilded bold: index-head projections) \\
$\mathcal A^{(\ell)}_t$ & Attended set at layer $\ell$: $\mathcal T^{(\ell)}_t\cup\mathcal S_t\cup\mathcal W_t$, of size $k_t+|\mathcal S_t\cup\mathcal W_t|$ \\
$\Tcache$ & Temporary index buffer, overwritten by each Full layer and released after the last layer \\
$\widetilde K^{(\ell)}_{\idx}$ & Persistent index-key cache at layer $\ell$ \\
$\mathbf c$, $c_\ell\in\{\Full,\Shared\}$ & Full/Shared layer pattern; $f(\ell)$ is the nearest preceding Full layer \\
$\Bset_m$, $\mathbf c^{(b)}$, $R_b$ & Budget menu of backbone $m$, library pattern for budget $b$, and its number of Full eligible layers ($\{3,5,9,18,36\}$ for the primary backbone) \\
$\rho_b=R_b/\Nidx$ & Realized refresh fraction \\
$s_u$, $b_u$, $\mathbf c_u$ & Controller observation before decision $u$, the selected budget, and the pattern $\mathbf c^{(b_u)}$ in force \\
$\mathbf p^{(\ell)}_t$, $\mathbf q^{(\ell)}_t$ & Aggregated teacher attention distribution and indexer output distribution over $\mathcal E_t$ \\
$J(\tau)$, $U(\tau)$ & Agent steps and controller decisions in episode $\tau$; $U(\tau)\leq U_{\max}$ \\
$\kappa$ & Fixed FLOP-per-byte conversion in the cost model \\
$C_u$, $C(\tau)$ & Compute cost of segment $u$ and of the episode \\
$C_{\idx,u}$ & Indexing cost: all-layer index-key maintenance and Full-only selection work \\
$C_{\mathrm{cache},u}$ & Cost of $\Tcache$ traffic and pattern switches for segment $u$ \\
$C_{\ctrl,u}$ & Controller overhead: features, probe, and policy inference for segment $u$ \\
$V(\tau)$ & Immutable terminal verifier score \\
$B$, $C_0$ & Expected episode compute budget, and the frozen mean all-refresh training cost \\
$\overline C(\tau),\overline B$ & Episode cost and budget normalized by $C_0$ \\
$\lambda$ & Nonnegative dual price for compute \\
$S_{\mathrm{RL}}$, $S_{\mathrm{ctrl}}$ & Independent RL training runs, and controller seeds per final task \\
\bottomrule
\end{tabular}
\end{table*}
\section{Execution Details}
\label{app:execution}

\subsection{Indexer specification}
\label{app:indexer-spec}
The score of Equation~\ref{eq:index-score} adapts the DSA lightning indexer, which sums ReLU-activated query-key products weighted by query-dependent head weights \citep{deepseek2025v32}. Our specification differs from DSA in two respects: each score head has its own key projection, whereas DSA shares one index key across heads, and the head weights are softmax-normalized gates, whereas DSA uses an unnormalized projection. The $1/\sqrt{d_{\idx}}$ scale is equivalent to the constant scaling that the DSA reference implementation folds into its head weights, because ReLU commutes with positive scaling. Per-head keys make the index-key cache $H_{\idx}$ times larger than a shared key of the same dimension would.

\subsection{Index-key maintenance and cache history}
\label{app:key-maintenance}
\begin{figure*}[tbp]
\centering
\begin{tikzpicture}[air figure, >=Latex, line width=.5pt, node distance=4mm and 4mm,
  box/.style={draw=black!30, rounded corners=2pt, align=center, inner sep=2.5pt},
  edge/.style={->, draw=black!65, line width=.7pt},
  note/.style={align=center, text=black!75},
  head/.style={air title, inner sep=0pt},
  half/.style={box, text width=6.1cm, minimum height=1.75cm},
  layer/.style={box, text width=2.1cm, minimum height=1.25cm, draw=refreshblue, fill=white},
  ids/.style={box, minimum width=1.2cm, minimum height=.45cm},
  third/.style={box, text width=3.9cm, minimum height=1.75cm},
  full/.style={box, text width=13.2cm}]
\node[head, anchor=north west] (hA) at (0,0) {A\quad A causal decision fixes one mask for the segment};
\node[half, fill=softgray, anchor=north west] (history) at (0,-.45)
  {\textbf{Available history $H_u$}\\latest tool event, trajectory state,\\previous budget and cost, and\\completed model and index statistics};
\node[half, fill=softblue, draw=refreshblue, anchor=north east] (choose) at (13.75,-.45)
  {\textbf{Controller and mask library}\\$s_u=\phi(H_u)$,\enspace $b_u\sim\pi_\theta(\cdot\mid s_u)$;\\use $c_u=c^{(b_u)}$ from the legal menu;\\the first eligible layer is Full};
\draw[edge] (history.east) -- (choose.west);
\node[half, dashed, draw=warmorange, fill=softorange, below=7mm of history] (probe)
  {\textbf{Optional paid probe (ablation)}\\scratch embeddings, native prefix,\\and first indexer; three summaries;\\no cache writes; charged in full};
\node[half, fill=softgreen, draw=agentgreen, below=7mm of choose] (segment)
  {\textbf{Run the segment under the mask}\\prefill newly appended tokens,\\then decode up to $Q=512$ tokens;\\choose again at the next boundary};
\draw[edge] (choose.south) -- node[right, note]{fixed mask} (segment.north);
\draw[edge, dashed, draw=warmorange] (probe.north) -- node[right, note]{summaries} (history.south);
\path (probe.south west) ++(0,-.35) coordinate (sepB);
\draw[black!15] (sepB) -- ++(13.75,0);
\node[head, anchor=north west] (hB) at ([yshift=-2mm]sepB) {B\quad One query shares selected IDs across eligible depth};
\node[note, anchor=north west, inner sep=0pt] (gl) at ([yshift=-2mm]hB.south west) {Eligible global layers L6, L12, L18, L24, and L30; legal mask $(\Full,\Shared,\Shared,\Full,\Shared)$ with $R=2$.};
\node[layer, fill=softblue, anchor=north west] (f6) at ([yshift=-3mm]gl.south west) {\textbf{L6 \textperiodcentered\ Full}\\score, select\\seed buffer};
\node[layer, right=5.3mm of f6] (s12) {\textbf{L12 \textperiodcentered\ Shared}\\read L6 IDs\\gather own KV};
\node[layer, right=5.3mm of s12] (s18) {\textbf{L18 \textperiodcentered\ Shared}\\read L6 IDs\\gather own KV};
\node[layer, right=5.3mm of s18, draw=warmorange, fill=softorange] (f24) {\textbf{L24 \textperiodcentered\ Full}\\score, select\\overwrite IDs};
\node[layer, right=5.3mm of f24, draw=warmorange] (s30) {\textbf{L30 \textperiodcentered\ Shared}\\read L24 IDs\\gather own KV};
\foreach \a/\b in {f6/s12,s12/s18,s18/f24,f24/s30} \draw[edge] (\a.east) -- (\b.west);
\node[ids, draw=refreshblue, fill=softblue, below=8mm of f6] (idf6) {$\mathcal T_t^{(6)}$};
\node[ids, draw=refreshblue, fill=softblue, below=8mm of s12] (ids12) {$\mathcal T_t^{(6)}$};
\node[ids, draw=refreshblue, fill=softblue, below=8mm of s18] (ids18) {$\mathcal T_t^{(6)}$};
\node[ids, draw=warmorange, fill=softorange, below=8mm of f24] (idf24) {$\mathcal T_t^{(24)}$};
\node[ids, draw=warmorange, fill=softorange, below=8mm of s30] (ids30) {$\mathcal T_t^{(24)}$};
\draw[edge, draw=refreshblue] (f6.south) -- node[right, note]{write} (idf6.north);
\draw[edge, draw=refreshblue] (ids12.north) -- node[right, note]{read} (s12.south);
\draw[edge, draw=refreshblue] (ids18.north) -- node[right, note]{read} (s18.south);
\draw[edge, draw=warmorange] (f24.south) -- node[right, note]{write} (idf24.north);
\draw[edge, draw=warmorange] (ids30.north) -- node[right, note]{read} (s30.south);
\draw[edge, draw=refreshblue] (idf6.east) -- node[above, note]{same IDs} (ids12.west);
\draw[edge, draw=refreshblue] (ids12.east) -- node[above, note]{same IDs} (ids18.west);
\draw[edge, draw=warmorange] (idf24.east) -- node[above, note]{same IDs} (ids30.west);
\node[full, fill=softgray, anchor=north] (native) at ([yshift=-4mm]ids18.south)
  {\textbf{Native attention stays native.} Layers 1--5, 7--11, 13--17, 19--23, 25--29, and 31--34 keep\\their sliding-window attention and never read or overwrite the selected-ID buffer.};
\path (native.south) ++(0,-.35) coordinate (c0);
\path (sepB |- c0) coordinate (sepC);
\draw[black!15] (sepC) -- ++(13.75,0);
\node[head, anchor=north west] (hC) at ([yshift=-2mm]sepC) {C\quad Query-local selections reset; persistent caches remain};
\node[third, fill=softorange, draw=warmorange, anchor=north west] (release) at ([yshift=-3mm]hC.south west)
  {\textbf{Finish query $t$}\\after the last layer,\\release $\Tcache$; no IDs\\carry over to $t+1$};
\node[third, fill=softblue, draw=refreshblue, right=5.4mm of release] (reset)
  {\textbf{Start query $t+1$}\\run the native prefix;\\the first eligible layer\\seeds fresh IDs};
\node[third, fill=softgreen, draw=agentgreen, right=5.4mm of reset] (continue)
  {\textbf{Continue through depth}\\Full layers overwrite IDs;\\Shared layers read the\\last Full layer's IDs};
\draw[edge] (release.east) -- (reset.west);
\draw[edge] (reset.east) -- (continue.west);
\node[full, fill=softgreen, draw=agentgreen!70, anchor=north] (persist) at ([yshift=-4mm]reset.south)
  {\textbf{Persistent state grows once per position during the ordinary forward pass.} Core KV is kept\\at every layer. Index keys are kept at every eligible layer, Shared layers included, so a later\\segment can promote a layer without replaying history. An eligible layer gathers its own KV\\at the selected IDs plus the sink and recent-window positions.};
\node[anchor=north, align=center, text=black!75] at ([yshift=-1mm]persist.south)
  {The paid probe runs on scratch state, and ordinary execution recomputes and appends.\\Cache entries keep their causal history, and changing the mask never recomputes earlier hidden states.};
\end{tikzpicture}%
\caption{One \method decision and one query on the Gemma-3-4B-it layout: (A) the mask is fixed from completed history; (B) Full layers write position IDs that Shared layers read, while sliding-window layers bypass the buffer; (C) IDs are released per query and caches persist.}
\label{fig:mechanism}
\end{figure*}
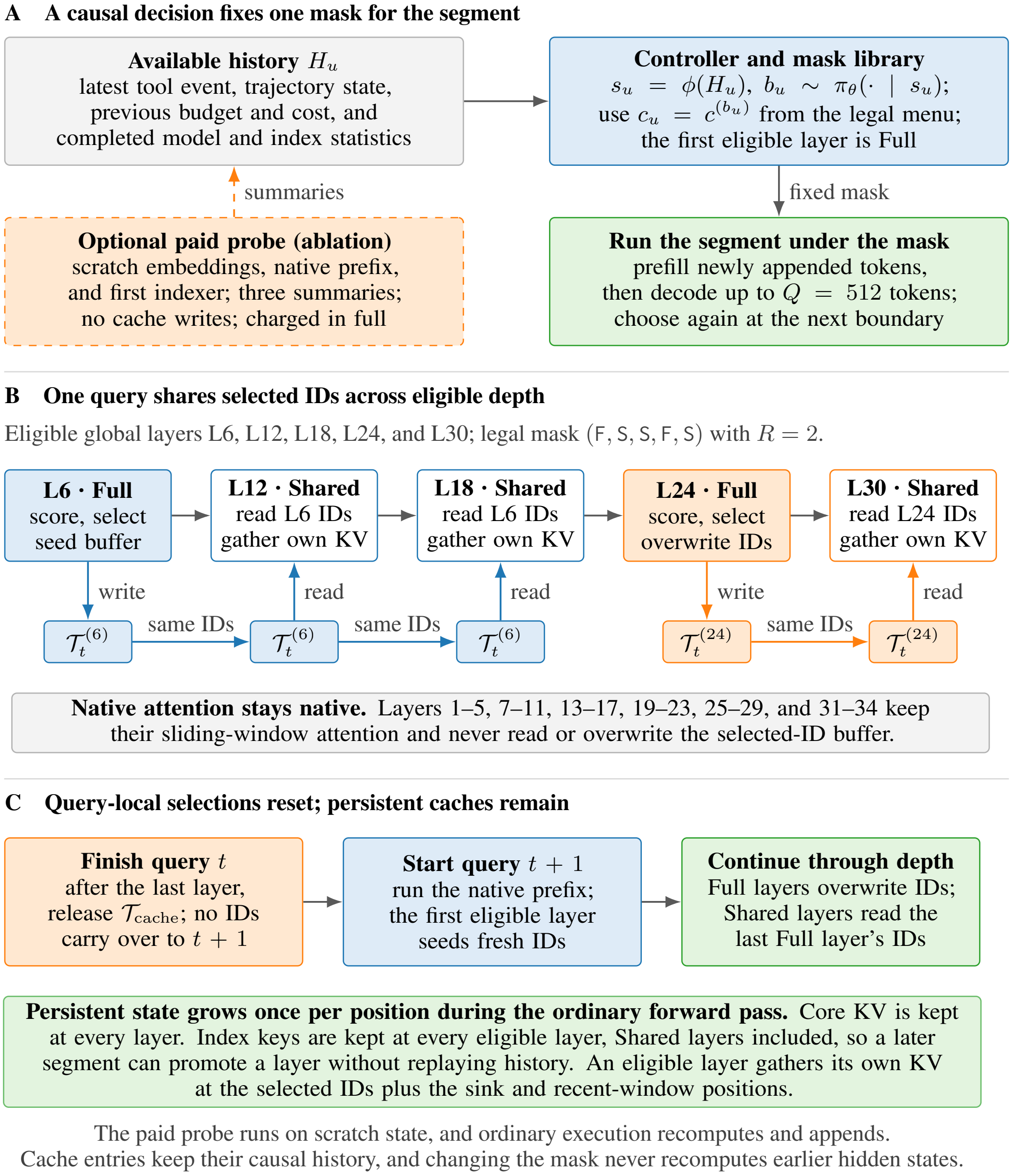

Section~\ref{sec:indexcache-semantics} explained why every eligible layer keeps its index keys while it is Shared: a layer promoted at a segment boundary finds keys for the whole prefix in its own cache and can score past positions without replaying past hidden states. This subsection records what that upkeep costs and what the cached values mean. Each eligible layer appends its core keys and values and its index keys at every token, whatever its current role. The upkeep costs $H_{\idx}d_{\idx}$ stored values per token and eligible layer, plus the index-key projection at eligible layers that are currently Shared. For comparison, the core cache holds $2n_{\mathrm{kv}}d_{\mathrm{head}}=2048$ key and value entries per token and layer for every backbone in Table~\ref{tab:models}. The alternative, reconstructing keys only at promotion time, would require either stored layer inputs, which take $d$ values per token and layer, or a replay of the prefix; storing the keys is the cheaper option. Every cached entry keeps the value computed when its token was processed. That value depended on the layer input at the time, and the layer input depended on the selections made under the pattern then in force, so two runs that chose different budgets earlier can hold different cache contents for the same prefix. This is the usual causal semantics of incremental decoding, and it is the reason the cost and feature definitions refer to the actual execution history. Figure~\ref{fig:mechanism} follows one decision and one query through this execution on the Gemma-3-4B-it layout, where sliding-window layers bypass the position buffer.

\subsection{Relation to IndexCache and the scope of the policy-class guarantee}
\label{app:relation}
\method inherits three elements from IndexCache: the observation that cross-layer top-$k$ overlap makes consecutive indexer calls partly redundant, the Full/Shared branch that skips those calls while every layer keeps its own KV tensors, and greedy loss-based search over patterns \citep{bai2026indexcache}. What changes is the policy class and its objective. IndexCache fixes one pattern from calibration loss; \method chooses among several precomputed patterns from the evolving agent state and learns that choice from terminal task feedback under a compute constraint. The setting differs too: IndexCache runs on models whose indexers were trained with sparse attention, whereas we inject an indexer into a dense checkpoint (Section~\ref{app:indexer}). In this paper, ``static IndexCache'' therefore means IndexCache's training-free search applied to our injected indexer.

\paragraph{What the policy class guarantees.}
A constant policy that always picks one library entry reproduces that static schedule exactly, including its selected positions and outputs, and a policy that ignores the state reduces to a fixed mixture of static schedules; both serve as controls (Section~\ref{sec:baselines}). Their execution costs match \method only in the dynamic-ready runtime, which maintains index keys at every eligible layer. A specialized static runtime omits those keys at permanently Shared layers and is the stronger systems comparator. The policy class contains every constant policy over the declared library. Under the same budget $\overline B$, execution in the dynamic-ready runtime, and the same cost model, in which a policy of the class pays its feature-extraction and controller-inference costs $C_{\ctrl,u}$ even when its output is constant, the best feasible policy in the class is therefore at least as good as the best feasible constant library policy. The argument covers neither a static schedule run without a controller, which avoids $C_{\ctrl,u}$, nor the specialized static runtime, which also avoids the index keys at permanently Shared layers. It also guarantees nothing empirical: training may reach a different policy, and the argument says nothing about held-out quality or latency. An empirical claim requires \method to beat the strongest feasible static schedule at matched realized cost (Section~\ref{sec:endpoints}).

\section{Schedule Library Construction}
\label{app:library}

The library separates where refresh happens from when it happens. It is built once per backbone, offline, before any controller exists, and it decides which layers are Full at each budget; the controller later decides only which budget to use.

\paragraph{Search objective.}
For each refresh count $R$ in the menu, we look for the pattern with $R$ Full eligible layers, $\ell_1$ among them, that gives the lowest next-token loss on agent-generated calibration sequences when the whole sequence, prefix included, runs under that static pattern. The loss is a token-weighted average whose weights balance environment-event categories and repositories. The calibration set $\D_{\mathrm{sched}}$ is collected once by all-refresh execution, and every candidate pattern is scored on this same reference corpus. This model-side surrogate decides only where to refresh; terminal reward enters later, at the outer level.

\paragraph{Greedy elimination.}
Exhaustive search over patterns is combinatorial, so we follow training-free IndexCache and remove layers greedily from the all-Full pattern \citep{bai2026indexcache}. Each step tries every remaining Full eligible layer except $\ell_1$ as Shared, evaluates the calibration loss on the same cached sequences, and commits the removal with the lowest loss. A path from $\Nidx$ Full layers down to the smallest menu entry $R_{\min}$ takes $\Nidx-R_{\min}$ steps and $\sum_{R=R_{\min}}^{\Nidx-1}R$ candidate evaluations: for the primary model, 33 steps and 627 evaluations from 36 to 3; for MiniMax-M2, 56 steps and 1,876 evaluations from 62 to 6; for Gemma's global layers, 4 steps and 10 evaluations. Because every budget's pattern is read off this single path, the library is nested. A held-out schedule-validation subset, drawn by owner group from the schedule-calibration partition with a fixed fraction and seed before the search, checks the stability of the path and selects nothing.

\section{Controller Observations and Feature Requirements}
\label{app:features}

The controller must decide before it pays for a refresh, so everything it observes comes from computation that has already finished. We keep three kinds of signal apart: lagged statistics that the default controller reads at no extra selector cost, paid probes that an ablation buys explicitly, and offline diagnostics that are never controller inputs.

\paragraph{Features by source.}
The controller receives a partial observation of the causal history $\mathcal H_u$, and every default feature exists before any model or selector computation governed by $b_u$. The observation $s_u=\phi(\mathcal H_u)$ concatenates five groups of features with missing-value indicators. Position features record context length, environment step, segment index, and tokens since the last tool return. Environment features derive from raw tool type, output length, traceback and test-result parser flags, file transitions, and elapsed tool steps, and a compact embedding of the latest tool output is the mean of the backbone's input embeddings over its tokens. Cost features record the normalized cumulative cost and its gap from the target. The remaining two groups are lagged: they summarize computation from the completed segment $u-1$. Lagged model features summarize the final token of that segment, including predictive entropy and a low-dimensional projection of its hidden state. Lagged index features include the previous mask, the layerwise distance to the most recent Full layer, and overlap, score entropy, and selection margin computed from queries that have already completed. The hidden-state and tool-output projections are principal-component maps fitted on controller-training states before reinforcement learning and frozen thereafter. Missing-value indicators distinguish the first segment, which has no completed predecessor, from observed zeros.

\paragraph{Timing and cost requirements.}
Table~\ref{tab:features} gives the requirements that each feature group meets. For every feature we record its source token, source layer, timestamp, cache version, formula, missing-value rule, dependence on earlier actions, runtime, memory traffic, and any duplicated selector work. Future tokens, terminal reward, counterfactual loss, repository identity, and any statistic produced after $b_u$ is chosen are excluded.

\begin{table*}[tbp]
\centering
\small
\caption{Controller observation groups and the timing and cost requirements each meets. Every value exists before the segment's budget is chosen.}
\label{tab:features}
\begin{tabular}{p{0.21\textwidth}p{0.28\textwidth}p{0.42\textwidth}}
\toprule
Block & Example fields & Requirement and cost treatment \\
\midrule
Trajectory position & Segment index, context length, elapsed tool steps & Read from history at the boundary and normalize with training-only statistics \\
Interface event & Traceback, test result, file read, edit, API reply & Derive from the latest completed tool event, with no verifier or future outcome \\
Budget history & Previous budget, cumulative normalized cost, target gap & Compute only from executed actions and the frozen cost model \\
Cache state & Previous mask, distance to last Full layer, completed index age & Record source token, layer, timestamp, cache version, and action-history dependence \\
Lagged uncertainty & Completed-segment entropy and logit margin & Read retained logits from segment $u-1$, with a missing flag at $u=0$ \\
Lagged index state & Completed-query overlap, score entropy, selection margin & Reuse statistics already produced in segment $u-1$, with no new prefix scoring \\
Lagged representation & Projected hidden state of the last completed token & Freeze source layer, PCA map, normalization, precision, and retention cost \\
Observation content & PCA-projected mean input embedding of the latest tool output & Remove raw owner and repository identity; fit no projection on final tasks \\
Paid current-prefix probe & Action-independent entropy, overlap, or drift proxy & Run once before $b_u$, discard selected IDs, and include all compute and memory traffic in the cost \\
\bottomrule
\end{tabular}
\end{table*}

\paragraph{Paid probes.}
Score entropy, top-$k$ margin, or index overlap for the new query would require selector work, so the default controller never receives these current-query values. A paid-probe ablation runs one frozen, action-independent probe on the newly appended tokens before $b_u$ is chosen. The probe uses the frozen indexer of the first eligible layer $\ell_1$, which is Full under every budget and whose input depends only on the embeddings and on any native, noneligible layers below it, and it reports score entropy, top-$k$ margin, and overlap with the most recent completed selection at $\ell_1$. The probe runs on temporary state. It computes the embeddings, the native layers below $\ell_1$, and the index queries and keys of $\ell_1$ for the newly appended tokens in scratch buffers, scores them against the cached index keys of the prefix and against one another under the causal mask, and keeps only the three statistics; nothing is appended to the persistent caches, and the selected positions are discarded. The segment's forward pass then computes these quantities again and appends every key and value, and every index key, exactly once per position, as Algorithm~\ref{alg:main} specifies. Within the paid-probe comparison, the runtime executes the probe once per segment under every policy and adds all of its compute and memory traffic to $C_{\ctrl,u}$; each probe operation is counted once, separately from the forward pass that follows (Table~\ref{tab:counting}). This rule prevents a low-budget action from receiving refresh-level information for free.

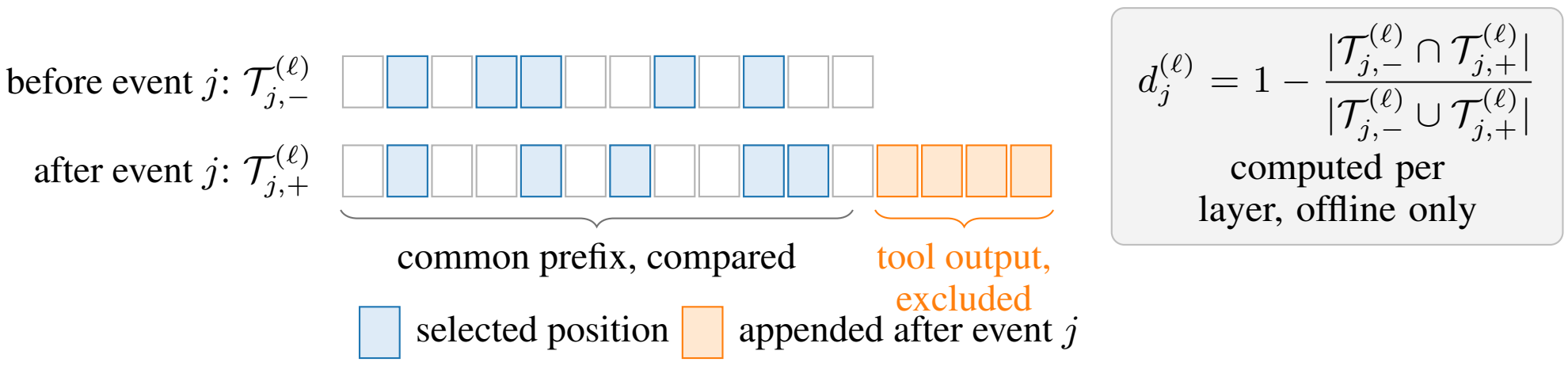
\begin{figure*}[ht]
\centering
\begin{tikzpicture}[
  air figure, x=0.93cm, y=0.93cm,
  cell/.style={draw=black!30, fill=white, minimum width=3.35mm, minimum height=4.28mm, inner sep=0pt},
  sel/.style={cell, draw=refreshblue, fill=softblue},
  new/.style={cell, draw=warmorange, fill=softorange},
  box/.style={draw=black!25, fill=softgray, rounded corners=3pt, align=center, inner sep=3.6pt}
]
\node[anchor=east] at (2.35,1.20) {before event $j$: $\mathcal T^{(\ell)}_{j,-}$};
\node[anchor=east] at (2.35,0.40) {after event $j$: $\mathcal T^{(\ell)}_{j,+}$};
\foreach \x in {0,...,11} {\node[cell] at (2.65+0.40*\x,1.20) {}; \node[cell] at (2.65+0.40*\x,0.40) {};}
\foreach \x in {12,...,15} {\node[new] at (2.65+0.40*\x,0.40) {};}
\foreach \x in {1,3,4,7,9} {\node[sel] at (2.65+0.40*\x,1.20) {};}
\foreach \x in {1,4,6,9,10} {\node[sel] at (2.65+0.40*\x,0.40) {};}
\draw[decorate, decoration={brace,mirror,amplitude=4pt}, draw=black!55] (2.45,0.05) -- (7.05,0.05)
  node[midway, below=4pt] {common prefix, compared};
\draw[decorate, decoration={brace,mirror,amplitude=4pt}, draw=warmorange] (7.25,0.05) -- (8.85,0.05)
  node[midway, below=4pt, text=warmorange, align=center] {tool output,\\excluded};
\node[box, text width=35.15mm] at (11.40,0.80) {$d^{(\ell)}_{j}=1-\dfrac{|\mathcal T^{(\ell)}_{j,-}\cap\mathcal T^{(\ell)}_{j,+}|}{|\mathcal T^{(\ell)}_{j,-}\cup\mathcal T^{(\ell)}_{j,+}|}$\\[2pt] computed per layer, offline only};
\node[sel] at (2.80,-1.05) {};
\node[anchor=west] at (2.98,-1.05) {selected position};
\node[new] at (5.70,-1.05) {};
\node[anchor=west] at (5.88,-1.05) {appended after event $j$};
\end{tikzpicture}
\caption{Offline structural-change diagnostic: selected positions of matched queries before and after event $j$ are compared on their common prefix by Jaccard distance.}
\label{fig:structural-diversity}
\end{figure*}

\paragraph{Offline structural diagnostic.}
For offline analysis only, we quantify structural change around an environment event $j$. Let $\mathcal T^{(\ell)}_{j,-}$ and $\mathcal T^{(\ell)}_{j,+}$ denote the selected IDs on matched diagnostic queries immediately before and after the event, restricted to their common prefix. Their Jaccard distance, one minus the ratio of intersection to union, computed for each layer, tests whether agent events alter retrieval structure across depth (Figure~\ref{fig:structural-diversity}). Because it uses post-event selector output, the diagnostic is never a controller input.

\paragraph{Controller architecture.}
The controller is a two-affine-layer network (hidden width $h$, a pointwise nonlinearity, and $|\Bset_m|$ outputs) whose logits, divided by a fixed temperature $\tau_\pi$, define a categorical distribution $\pi_\theta(b\mid s)$ over the backbone's $|\Bset_m|$ budgets; Equation~\ref{eq:actor-params} gives its parameter count, and $d_s$ and $h$ are fixed per backbone. We choose $\tau_\pi$ on controller-training data before reinforcement learning and keep it fixed afterward, including at deployment, so the deployed policy is the policy that was trained. Deterministic argmax routing is evaluated only as a secondary variant with its own feasibility check. A value network $v_\psi(s)$ with the same hidden width, a scalar output, and no shared parameters serves as the baseline during policy-gradient training and is discarded afterward. The controller is kept small to limit its latency and its capacity to overfit; its per-segment inference, $2(d_sh+h|\Bset_m|)$ FLOPs plus activations, and the feature computation that precedes it are charged to $C_{\ctrl,u}$ (Table~\ref{tab:counting}). Two ablations restrict the state: the model-only controller removes interface features, and the phase-only baseline keeps deterministic event metadata and drops model and index signals.

\section{Cost Model and Counting Rule}
\label{app:counting}

Two different quantities are called cost in this paper. The optimization cost is the analytic indexing and control cost of Section~\ref{sec:cost}: it is the axis of the constraint, the dual update, and the decision rules, and it is computed identically for every policy from tensor shapes. The resource measurements, such as full-model compute, wall-clock time, and memory, describe what a deployment would experience and are reported separately.

\paragraph{What the optimization cost leaves out.}

Core attention is excluded from the optimization cost because, for a given query, every budget attends to the same number of positions: $k_t$ selected positions plus the $|\mathcal S_t\cup\mathcal W_t|$ available sink and window positions, each counted once. Its total, like that of projections and feed-forward layers, still varies with episode length, so full-model FLOP equivalents and end-to-end latency are reported separately; a full-model saving cannot exceed the indexer's share of per-token work. A policy can also lower $C(\tau)$ by ending episodes early, which shows up as lower success when it harms the task, so we also report cost per processed token, generated tokens, tool calls, and episode length. Resource reports further include index-key and core KV-cache bytes, peak memory, and cache-allocation latency.

\paragraph{Normalization and the static reference.}
Raw counts would give a poorly scaled dual problem next to a terminal score in $[0,1]$. For each suite and context condition, we therefore divide costs by $C_0$, the mean episode cost of all-refresh execution on the controller-training partition, frozen before controller fitting; segment costs are normalized the same way, $\overline C_u=C_u/C_0$. A preregistered ratio $\beta$ sets the normalized budget $\overline B=\beta$. On a separate static-selection partition, the static reference $b_0$ is the library entry with the highest success rate among those whose mean normalized cost stays within $\overline B$, measured in the specialized static runtime, with ties broken toward lower cost and then fewer Full layers. Its constant policy $\pi_0$ is frozen before routing labels are built. If no library entry stays within budget, the operating point is invalid.

\paragraph{Counting rule.}
Table~\ref{tab:counting} lists the analytic counts behind Equation~\ref{eq:index-cost} and the buffer and controller terms of Section~\ref{sec:cost}. For query position $t$ and layer $\ell$, let $E_t=|\mathcal E_t|$, let $k_t=\min\{k,E_t\}$ be the number of positions actually selected, let $b_{\idx}$ be the bytes per index-key element, and let position IDs use 4 bytes. Weight reads are counted once per forward step and divided among the query tokens processed in that step. Each entry is converted to a FLOP equivalent $F^{\mathrm{op}}+\kappa M$ (Section~\ref{sec:cost}). We fix per backbone whether the scoring kernel materializes scores in accelerator memory; if it does, their writes and reads are added to the byte count.

\begin{table*}[ht]
\centering
\small
\caption{Cost breakdown by operation, per query position $t$ and layer $\ell$ unless stated otherwise.}
\label{tab:counting}
\begin{tabular}{p{0.19\textwidth}p{0.35\textwidth}p{0.35\textwidth}}
\toprule
Term & Floating-point operations & Bytes \\
\midrule
$F_{\mathrm{key}}(t,\ell)$ & $2dH_{\idx}d_{\idx}$ for key projections, plus input normalization & $H_{\idx}d_{\idx}b_{\idx}$ written, plus amortized weight reads \\
$F_{\mathrm{sel}}(t,\ell)$ & $2dH_{\idx}(d_{\idx}+1)$ for query and gate projections; $2H_{\idx}d_{\idx}E_t$ for dot products; $3H_{\idx}E_t$ for scaling, ReLU, and gating; $E_t$ for top-$k$ selection & $H_{\idx}d_{\idx}b_{\idx}E_t$ of index keys read, plus amortized weight reads \\
$F_{\mathrm w}(t,\ell)$, $F_{\mathrm r}(t,\ell)$ & None & $4k_t$ written at a Full layer or read at a Shared layer \\
$F_{\mathrm{sw}}(b_{u-1},b_u)$, per segment & None & $\lceil N/8\rceil$ for the new mask when $b_u\neq b_{u-1}$ \\
$F_\phi(\mathcal H_u)$, per segment & Entropy and margin over the retained final-token logits; PCA projections; overlap statistics & Logits, stored position sets, and projection weights read \\
$F_{\mathrm{probe}}(\mathcal H_u)$, per segment (paid-probe comparison only) & For the appended tokens: embeddings and native layers below $\ell_1$ (standard per-token counts); $F_{\mathrm{key}}$ at $\ell_1$; $F_{\mathrm{sel}}$ at $\ell_1$; entropy, margin, and overlap statistics & Weight and embedding reads; temporary writes and reads of layer inputs and index keys; prefix index keys of $\ell_1$ read; previous selection read \\
$F_\pi(s_u)$, per segment & Two-layer MLP: $2(d_sh+h|\Bset_m|)$ plus activations & Controller weights and $s_u$ read \\
\bottomrule
\end{tabular}
\end{table*}

\paragraph{Full-model and wall-clock measurements.}
Full-model FLOP equivalents, which are reported but excluded from the optimization cost, add the standard per-token counts for projections, feed-forward layers, and core attention over $|\mathcal A^{(\ell)}_t|$ positions.

\section{Sparse Attention and Indexer Preparation}
\label{app:indexer}

Every schedule in this paper runs on an injected indexer, so each backbone needs a working sparse-attention implementation before any library or controller can be built. Preparation attaches the indexer, distills it in two stages, and accepts or rejects it; Gemma-3-4B-it and MiniMax-M2 need further adaptations, and the training-aware IndexCache comparison adapts the indexers to one fixed pattern.

\subsection{Attaching the indexer}
\label{app:attach}

Every backbone in Table~\ref{tab:models} is dense in attention as released and contains no indexer, so \method runs on an injected selector. For each backbone we attach the indexer of Equation~\ref{eq:index-score} to every eligible layer and replace that layer's dense core attention with top-$k$ sparse attention through the model's attention interface, which keeps its rotary embeddings, query-key normalization, and grouped-query layout. Layers outside $\Lidx$ are not modified. All backbone parameters stay frozen. The indexer uses $H_{\idx}=8$ score heads and $k=2048$ selected positions, excluding sinks and the local window, on every backbone. $H_{\idx}$ is a shared experimental choice: the score heads have their own projections, separate from the backbone's key-value groups, of which the backbones have 8 (Qwen, MiniMax-M2) or 4 (Gemma-3-4B-it). The index dimension $d_{\idx}$, the sink set, the local-window width $w$, and the index-key precision are fixed per backbone. Each backbone stores $H_{\idx}d_{\idx}$ index-key values per token and eligible layer on top of its core cache of $2n_{\mathrm{kv}}d_{\mathrm{head}}=2048$ values per token and layer ($n_{\mathrm{kv}}=8$, $d_{\mathrm{head}}=128$ for Qwen and MiniMax-M2; $n_{\mathrm{kv}}=4$, $d_{\mathrm{head}}=256$ for Gemma).

DSA and the models evaluated by IndexCache train their indexers jointly with sparse attention \citep{deepseek2025v32,bai2026indexcache}. Our backbones were trained with dense attention only, so all-refresh sparse execution can lose quality relative to the native model. The sparse-attention check measures this loss on fitting validation before any controller is trained.

\subsection{Distillation}
\label{app:distill}
Each backbone's indexer is distilled from that backbone's own dense attention, as in the DSA warm-up stage \citep{deepseek2025v32}. For a query with a nonempty candidate set, the target $\mathbf p^{(\ell)}_t$ sums the dense attention weights of all $H_q$ query heads of layer $\ell$ ($H_q=16$, 32, 8, and 48 for Qwen3-1.7B, the 4B and 8B Qwen checkpoints, Gemma-3-4B-it, and MiniMax-M2) and renormalizes them over $\mathcal E_t$, and the indexer's output distribution $\mathbf q^{(\ell)}_t$ is the softmax of $\mathbf I^{(\ell)}_t$ over the same set. Only indexer parameters are trained, with
\begin{equation}
\mathcal L_{\mathrm{indexer}}
=\E_{t,\ell}\left[D_{\mathrm{KL}}\left(\operatorname{sg}[\mathbf p^{(\ell)}_t]\,\Vert\,\mathbf q^{(\ell)}_t\right)\right],
\label{eq:indexer-distill}
\end{equation}
where the expectation runs over eligible layers and over query positions with nonempty candidate sets in agent-generated prefixes from the backbone's indexer-training partition. Dense teacher passes first cache $\mathbf p^{(\ell)}_t$ and the layer inputs, and an initialization stage trains each indexer on these dense states. A second stage executes the all-refresh sparse student, recomputes the indexer inputs from the student's hidden states, stops gradients through those states, and keeps the same dense targets on the full candidate set; in DSA's sparse stage, the KL term covers only the selected set. Top-$k$ selection stays outside the gradient path, and backbone and KV-projection gradients remain disabled throughout.

The target sums the head attentions. Each head's dense attention at a global or full-attention layer is a distribution over the same causal positions $\{1,\ldots,t\}$, so the sum and the average differ only by the factor $1/H_q$, which cancels after restriction to $\mathcal E_t$ and renormalization. The restriction removes the sink and local positions, which the indexer never ranks, whereas IndexCache defines its aggregated distribution by averaging over heads without restriction \citep{bai2026indexcache}. Plain bold $\mathbf p$ and $\mathbf q$ denote distributions, while tilded bold vectors are index-head projections.

\subsection{Acceptance}
\label{app:accept}
Held-out acceptance tests on each backbone's indexer-validation subset, drawn by owner group from the indexer-training partition with a fixed fraction and seed before distillation, cover next-token loss, selected teacher mass, short-prefix execution, deterministic top-$k$ ties, incremental decoding, absolute position offsets, and equality of offline and online feature extraction. The acceptance decision precedes schedule search. After acceptance, the indexer is frozen, and every static schedule, router, bandit, and controller on that backbone shares its hash. A rejected indexer fails the sparse-attention check for every cell of that backbone.

\subsection{Backbone-specific adaptations}
\label{app:backbone-adapt}

\paragraph{Gemma-3-4B-it.}
The checkpoint is multimodal; we use only its language model. The layout is taken from Google DeepMind's reference implementation (\texttt{gemma/gm/nn/\_gemma.py} and \texttt{\_modules.py} at commit \texttt{0f7e742}) \citep{gemma_code}: the 4B model has 34 layers whose attention pattern repeats five local sliding-window layers and one global layer, with a 1,024-token window, 8 query and 4 key-value heads of dimension 256, query-key RMS normalization before rotary embeddings, and rotary bases of $10^4$ for local and $10^6$ (scaled by 8) for global layers. The global layers are therefore 1-indexed layers 6, 12, 18, 24, and 30, and only these are eligible. The reference code establishes the architectural layout, and agreement with our pinned checkpoint is checked separately. The configuration, tokenizer, and chat template of the pinned Hugging Face revision (\texttt{093f9f38}) match the reference layout, including the attention pattern, window size, and head geometry. The sliding-window layers never consume a selection made at a global layer and never overwrite $\Tcache$. The first global layer always refreshes, so every query receives one fresh selection; the menu $\{1,2,3,5\}$ therefore contains the anchor-only pattern. Because global and local layers use different rotary bases and the injected indexer reads the global layer's own input, the index keys are computed in the global layers' representation only.

\paragraph{MiniMax-M2.}
All 62 layers use full grouped-query softmax attention with 48 query and 8 key-value heads, partial rotary embeddings, and per-layer query-key normalization, and each block routes tokens to 8 of 256 experts. The released weights are FP8 with block scaling. The indexer is attached through an attention adapter in the serving engine, which computes index keys from each layer's input, keeps expert routing and expert weights unchanged, and runs with the tensor and expert parallel layout fixed for the backbone. The configuration lists multi-token-prediction modules, but the released checkpoint contains no weights for them; speculative and multi-token decoding are disabled for every method, and any later study of speculative decoding would be a separate condition. The adapter reproduces the unmodified engine's dense outputs when the indexer is bypassed, and a reference implementation's all-refresh sparse outputs, on fixed prompts within a fixed tolerance, as part of the infrastructure check. Hosted access to the model cannot implement these changes. Preserving the interleaved thinking history means that earlier \texttt{<think>} blocks remain in the prefix and are attended, selected, and costed like any other tokens.

\subsection{Training-aware IndexCache}
\label{app:ta-indexcache}
Training-aware IndexCache trains the indexers for one fixed pattern with a multi-layer distillation loss, and its sparse training phase also updates the remaining weights with the language-modeling loss \citep{bai2026indexcache}. We compare \method against a frozen-backbone variant that keeps only the indexer loss (Appendix~\ref{app:comparisons}). For a pattern $\mathbf c$, the group served by Full layer $\ell$ is $\mathcal G_\ell=\{\ell\}\cup\{\ell'\in\Lidx:c_{\ell'}=\Shared,\ f(\ell')=\ell\}$, and multi-layer distillation minimizes
\begin{equation}
\begin{aligned}\mathcal L_{\mathrm{multi}}
&=\sum_{\ell:c_\ell=\Full}
\frac{1}{|\mathcal G_\ell|}
\sum_{\ell'\in\mathcal G_\ell}
\sum_{t:\mathcal E_t\neq\varnothing}\\ &\qquad
D_{\mathrm{KL}}
\left(
\operatorname{sg}[\mathbf p^{(\ell')}_t]
\ \Vert\
\mathbf q^{(\ell)}_t
\right).\end{aligned}
\label{eq:multi-layer}
\end{equation}
Because each budget defines different served groups, an indexer adapted for one pattern is fitted to that pattern's groups only. \method therefore keeps one frozen indexer bank shared by every library entry, and training-aware IndexCache is evaluated as a separate comparison with one adapted bank per evaluated pattern.

\section{Supervised Warm Start}
\label{app:warm}

The warm start turns logged agent behavior into an initial controller whose predicted cost meets the budget, in three steps: replay produces counterfactual labels on fixed text, regressors predict those labels from the observation, and a priced soft target over budgets defines the policy that the controller imitates.

\paragraph{Replay labels.}
All-refresh execution collects agent segments on controller-training environments. For each logged pre-action history $\mathcal H_u$, we keep the continuation the behavior run produced and replay it with teacher forcing under every library pattern, from the same prefix and caches. Replay yields, for each budget $b$, the mean negative log-likelihood $\NLL_u(b)$ of the continuation's assistant tokens and its normalized compute cost $\overline C_u(b)$. These are counterfactual labels on fixed text: they do not show how another budget would have changed later text, tool calls, or terminal success.

\paragraph{Predictors and the soft target.}
Local regret $\delta_u(b)=\NLL_u(b)-\NLL_u(b_0)$ measures the loss increase relative to the static reference. Two action-conditioned regressors, $f_\omega(s,b)$ for regret and a nonnegative $g_\chi(s,b)$ for normalized cost, are fitted with a Huber loss averaged over all actions in the menu. Their predictions are cross-fitted, so the prediction for a record comes from models that never saw that record's split group (Section~\ref{sec:splits}). For a compute price $\nu\geq0$, the predictions define a soft target over budgets,
\begin{equation}
\bar\pi_\nu(b\mid s)\propto e^{-[f_\omega(s,b)+\nu\,g_\chi(s,b)]/\tau_0},
\label{eq:warm-target}
\end{equation}
which prefers budgets with low predicted regret and, as $\nu$ grows, budgets with low predicted cost. At every state the expected predicted cost under $\bar\pi_\nu$ is nonincreasing in $\nu$, because its derivative is minus the variance of $g_\chi$ under $\bar\pi_\nu$, divided by $\tau_0$. We therefore find by bisection the smallest price $\nu_0\geq0$ at which the expected predicted episode cost over the logged trajectories meets $\overline B$; if no price suffices, the operating point is infeasible. The warm-start parameters $\theta_0$ minimize the KL divergence from $\bar\pi_{\nu_0}$ to $\pi_\theta$, averaged over logged states. Because those states come from all-refresh execution, the warm policy's own cost can differ from $\overline B$. The reinforcement-learning stage adjusts the price on its own trajectories, and the validation bounds of Appendix~\ref{app:rl} decide whether any trained policy meets the budget. The frozen warm policy also serves as the supervised baseline.

\section{Constrained Reinforcement Learning Details}
\label{app:rl}

\paragraph{Gradient estimator and update schedule.}
The policy gradient of Equation~\ref{eq:policy-gradient} is an unbiased estimate of $\nabla_\theta\mathcal L$ at fixed $\lambda$ when episodes come from the current policy, have bounded length and bounded score and cost, $\pi_\theta$ is positive and differentiable, and the baseline depends only on $s_u$ and on parameters fixed before the batch. Costs incurred before decision $u$ cannot depend on $b_u$, which justifies summing cost from $u$, and the constant $\lambda\overline B$ has no gradient. An inaccurate critic, including one fitted under an earlier $\lambda$, adds variance but no bias. Episodes end at submission, environment termination, or a token, tool-step, or wall-time limit, so they contain at most $U_{\max}$ decisions. During training the runtime executes every sampled budget as sampled, so the gradient uses the logged budget that actually ran; work forced by the frozen cache-rebuild rule after a changed prefix is counted in the cost of the segment in which it occurs. The actor takes exactly one step per batch, the critic takes $K_\psi$ squared-error steps on the same batch, and the price is updated every $P_\lambda$ batches with a smaller step size, so it moves on a slower timescale \citep{borkar2005actor,tessler2019rcpo}. We run $S_{\mathrm{RL}}$ independent training runs from $\theta_0$ with different controller seeds.

\paragraph{Validation bounds and deployment.}
Section~\ref{sec:rl} explains why feasibility is tested on held-out data. Before training, we freeze a cost tolerance $\epsilon_C$, a confidence level $1-\alpha$, and a finite candidate set containing every predeclared checkpoint of every run and every static library policy. On the controller-validation partition, a grouped bootstrap gives each candidate a one-sided upper bound on expected normalized cost and a lower bound on expected success, each at level $1-\alpha/(2M)$ for $M$ candidates, so by the Bonferroni inequality all bounds hold jointly with probability at least $1-\alpha$, up to bootstrap approximation error. Among learned checkpoints whose cost bound is at most $\overline B+\epsilon_C$, we deploy the one with the largest success bound. If none qualifies, we deploy the qualifying static policy with the largest success bound, and if no static policy qualifies either, the operating point is declared infeasible. The deployed controller samples from the trained policy with the same $\tau_\pi$, using a counter-based random stream keyed by task and decision index.

\begin{figure*}[tbp]
\centering
\begin{tikzpicture}[
  air figure, x=0.93cm, y=0.93cm,
  box/.style={draw=black!30, fill=white, rounded corners=3pt, minimum width=20.46mm, minimum height=8.37mm, align=center, inner sep=1.8pt},
  pbox/.style={box, draw=violet, fill=softviolet},
  gbox/.style={box, draw=agentgreen, fill=softgreen},
  obox/.style={box, draw=warmorange, fill=softorange},
  arrow/.style={-{Latex[length=1.8mm]}, draw=black!65, line width=0.45pt}
]
\node[pbox] (policy) at (0.95,0) {policy\\$\pi_\theta$};
\node[gbox] (rollout) at (3.80,0) {on-policy\\agent episodes};
\node[obox] (measure) at (6.65,0) {terminal $V$\\normalized $\overline C$};
\node[pbox, text width=23.75mm] (actor) at (9.75,0) {one actor step,\\$K_\psi$ critic steps};
\node[gbox, text width=22.8mm] (gate) at (11.95,-1.55) {validation\\feasibility check};
\node[obox, minimum width=33.48mm] (dual) at (6.65,-1.55) {slower dual update\\$\lambda\leftarrow[\lambda+\eta_\lambda(\overline C-\overline B)]_+$};
\draw[arrow] (policy) -- (rollout);
\draw[arrow] (rollout) -- (measure);
\draw[arrow] (measure) -- (actor);
\draw[arrow, dashed] (actor.east) -| node[pos=0.75, right, align=left] {saved\\checkpoints} (gate.north);
\draw[arrow] (measure) -- (dual);
\draw[arrow] (dual.east) -| (9.75,-0.48);
\draw[arrow, draw=violet] (10.40,-0.48) -- (10.40,-2.45) -| (policy.south);
\end{tikzpicture}
\caption{Primal-dual training: each batch of episodes gives one actor step and several critic steps, the price moves on a slower timescale, and validation bounds select the deployed policy.}
\label{fig:rl-loop}
\end{figure*}
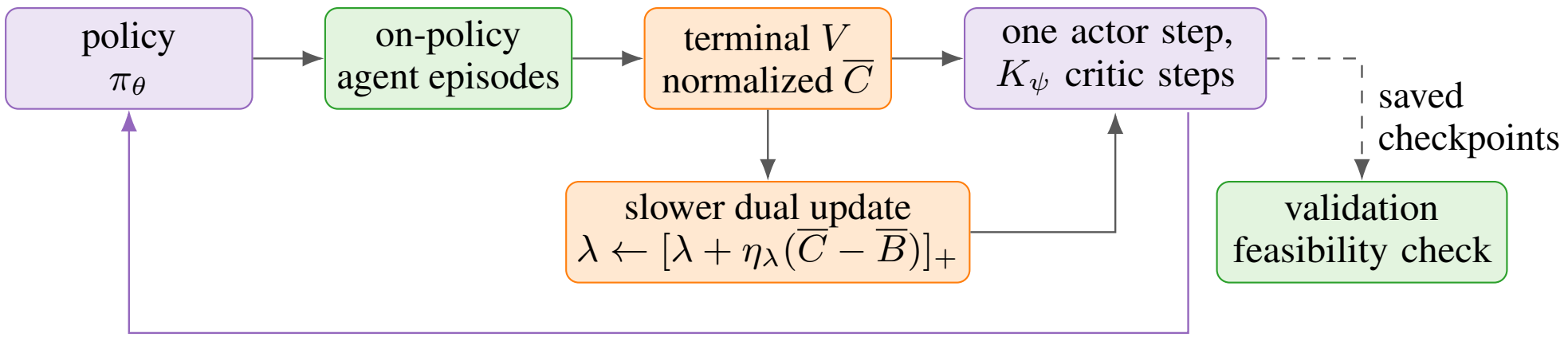

Figure~\ref{fig:rl-loop} sketches the training loop. Algorithm~\ref{alg:episode} gives the full online execution of one episode, Algorithm~\ref{alg:prep} the offline preparation and warm start, and Algorithm~\ref{alg:air} the training loop with policy selection.

\begin{algorithm*}[tbp]
\caption{Online execution of one episode under a controller $\pi$ (deployment, rollouts, and validation)}
\label{alg:episode}
\small
\begin{algorithmic}[1]
\Require backbone with frozen indexer and nested library, feature map $\phi$, policy $\pi$, limits, controller seed
\Ensure trajectory log, terminal score $V(\tau)$, compute cost $C(\tau)$
\State Empty all caches; queue the system and task prompt for prefill; $u\gets0$
\While{the episode has not terminated}
  \State $s_u\gets\phi(\mathcal H_u)$ from completed computation only; sample $b_u\sim\pi(\cdot\mid s_u)$ from the stream keyed by (task, $u$)
  \State \Call{Forward}{$t,\mathbf c^{(b_u)}$} (Algorithm~\ref{alg:main}) for each query position $t$ of segment $u$: queued prefill, then up to $Q$ decoded tokens
  \State Add $C_u=C_{\idx,u}+C_{\mathrm{cache},u}+C_{\ctrl,u}$ (Section~\ref{sec:cost})
  \State If the response is complete, parse its action; if it is a tool call, execute it and queue the observation for prefill
  \State $\mathcal H_{u+1}\gets\mathcal H_u$ extended with $b_u$ and the segment's outcome; $u\gets u+1$
\EndWhile
\State \Return the log, $V(\tau)$, and $C(\tau)=\sum_u C_u$
\end{algorithmic}
\end{algorithm*}

\begin{algorithm*}[tbp]
\caption{Offline preparation and supervised warm start. Every data collection and the measurement of $C_0$ use all-refresh execution.}
\label{alg:prep}
\small
\begin{algorithmic}[1]
\Require backbone, grouped partitions (Section~\ref{sec:splits}), menu $\Bset_m$, ratio $\beta$, temperature $\tau_0$
\Ensure frozen indexer, nested library, $C_0$, static reference $b_0$, and warm start $\theta_0$; or an infeasibility verdict
\State Train the indexer with Eq.~\ref{eq:indexer-distill}; run its acceptance tests; freeze it
\State Collect $\D_{\mathrm{sched}}$; build the library by greedy elimination (Section~\ref{app:library}); freeze it
\State Measure $C_0$ and set $\overline B\gets\beta$; select and freeze $b_0$; run the capability check (Sections~\ref{sec:cost} and~\ref{sec:tasks})
\State Collect $\D_{\mathrm{route}}$
\For{each logged pre-action history $\mathcal H_u$ in $\D_{\mathrm{route}}$ and every $b\in\Bset_m$}
  \State Replay the fixed continuation under $\mathbf c^{(b)}$; record $\NLL_u(b)$ and $\overline C_u(b)$
\EndFor
\State Cross-fit the regret predictor $f_\omega$ and the cost predictor $g_\chi$ by split group
\State By bisection, find the smallest $\nu_0\ge0$ at which the predicted cost of $\bar\pi_{\nu_0}$ (Eq.~\ref{eq:warm-target}) meets $\overline B$; if none exists, \Return operating point infeasible
\State Fit $\theta_0$ by minimizing the KL divergence from $\bar\pi_{\nu_0}$ to $\pi_\theta$
\end{algorithmic}
\end{algorithm*}

\begin{algorithm*}[tbp]
\caption{Constrained policy-gradient training and policy selection}
\label{alg:air}
\small
\begin{algorithmic}[1]
\Require outputs of Algorithm~\ref{alg:prep}, $\tau_\pi$, $\lambda_0$, step sizes, $K_\psi$, $P_\lambda$, $S_{\mathrm{RL}}$, rollout budget, $\epsilon_C$, $\alpha$
\Ensure one selected learned or static policy, or an infeasibility verdict
\For{run $r=1,\ldots,S_{\mathrm{RL}}$}
  \State $\theta\gets\theta_0$; initialize $\psi$; $\lambda\gets\lambda_0$
  \Repeat
    \State Run a batch $\mathcal K$ of episodes with Algorithm~\ref{alg:episode} under $\pi_\theta$
    \State Compute $G_u$ (Eq.~\ref{eq:policy-gradient}) for every decision with the current $\lambda$
    \State Take one actor step along $\widehat g_\theta$ (Eq.~\ref{eq:policy-gradient}), then $K_\psi$ critic steps regressing $v_\psi(s_u)$ onto $G_u$
    \State Update $\lambda$ (Eq.~\ref{eq:dual-update}) if the batch count is a multiple of $P_\lambda$
    \State Save the checkpoint if its position is predeclared
  \Until{the rollout budget is exhausted}
\EndFor
\State Compute simultaneous cost and success bounds on controller validation for every candidate
\State $\mathcal F\gets$ candidates whose cost bound is at most $\overline B+\epsilon_C$
\If{$\mathcal F$ contains a learned checkpoint}
  \State \Return the learned checkpoint in $\mathcal F$ with the largest success bound
\ElsIf{$\mathcal F\neq\emptyset$}
  \State \Return the static policy in $\mathcal F$ with the largest success bound
\Else
  \State \Return operating point infeasible
\EndIf
\end{algorithmic}
\end{algorithm*}

\section{Extended Experimental Design}
\label{app:ext-design}

\subsection{Backbones and scale}
\label{sec:models}

Each backbone answers a different question about where adaptive allocation helps (Table~\ref{tab:models}). Qwen3-4B-Instruct-2507 is the primary model \citep{qwen2025instruct2507}: it is small enough for repeated agent rollouts with reinforcement learning, and it supports long contexts natively. Qwen3-1.7B and Qwen3-8B form the size comparison within the Qwen3 family \citep{yang2025qwen3,qwen3_1p7b_card,qwen3_8b_card}. Gemma-3-4B-it adds a second attention family, in which global attention layers are interleaved with sliding-window layers \citep{gemma3report,gemma3_4b_it_card}. MiniMax-M2 is the mixture-of-experts extension, with about 230B total and 10B active parameters, and tests whether the method carries over to a large sparse model served by an inference engine \citep{minimax2026m2series,minimax_m2_card}. Qwen3-4B-Instruct-2507 differs from Qwen3-1.7B and Qwen3-8B in release, context length, and reasoning mode as well as in size, so a difference among the three cannot be attributed to parameter count alone. The original Qwen3-4B \citep{qwen3_4b_card}, which has the primary model's size and the release of the other two, serves as a supplementary control for this size comparison.

Every checkpoint is dense in attention as released: none contains an indexer, so \method always runs on an injected selector. For the Qwen checkpoints every layer is eligible. Gemma-3-4B-it repeats five sliding-window layers followed by one global layer, according to Google DeepMind's reference implementation \citep{gemma_code}; \method is applied only to the five global layers (1-indexed layers 6, 12, 18, 24, and 30), and the sliding-window layers keep their native 1,024-token window and never touch $\Tcache$. The infrastructure check confirms this layout against the pinned checkpoint's own configuration. All 62 layers of MiniMax-M2 use full grouped-query softmax attention, so all are eligible. Equation~\ref{eq:menu-rule} then gives the menus in Table~\ref{tab:models}: five budgets for the Qwen checkpoints and MiniMax-M2, and four for Gemma, whose deduplicated menu includes the anchor-only budget $R=1$.

\begin{table*}[htbp]
\centering
\scriptsize
\caption{Backbones. ``Supported context'' follows each model card; caps, menus, and settings are our choices (Hugging Face repositories pinned to fixed revisions). $N$: decoder layers; $\Nidx$: eligible layers; menus follow Equation~\ref{eq:menu-rule}.}
\label{tab:models}
\setlength{\tabcolsep}{3pt}
\begin{tabular}{P{0.13\textwidth}P{0.06\textwidth}P{0.135\textwidth}P{0.05\textwidth}P{0.115\textwidth}P{0.115\textwidth}P{0.08\textwidth}P{0.095\textwidth}P{0.065\textwidth}}
\toprule
Checkpoint (role) & Params (total/ active) & Attention & $N$ / $\Nidx$ & Menu $\mathcal R_m$ (fractions) & Supported context & Context caps & Reasoning mode & Precision \\
\midrule
Qwen3-4B-Instruct-2507 (primary) & 4.0B & GQA, 32 Q / 8 KV heads & 36 / 36 & 3, 5, 9, 18, 36 (.08, .14, .25, .50, 1) & 262,144 native & 32K, 128K & non-thinking only & BF16 \\
Qwen3-1.7B (smaller) & 1.7B & GQA, 16 Q / 8 KV & 28 / 28 & 3, 4, 7, 14, 28 (.11, .14, .25, .50, 1) & 32,768 & 32K & hybrid; thinking off & BF16 \\
Qwen3-8B (larger) & 8.2B & GQA, 32 Q / 8 KV & 36 / 36 & 3, 5, 9, 18, 36 (.08, .14, .25, .50, 1) & 32,768; 131,072 with YaRN & 32K; 128K with YaRN $\times4$ & hybrid; thinking off & BF16 \\
Gemma-3-4B-it (second family) & 4B & 5 local (1,024) : 1 global; 8 Q / 4 KV & 34 / 5 & 1, 2, 3, 5 (.20, .40, .60, 1) & 128K (131,072) & 32K, 128K & none & BF16 \\
MiniMax-M2 (MoE extension) & 230B / 10B & GQA, 48 Q / 8 KV; 256 experts, 8 active & 62 / 62 & 6, 9, 16, 31, 62 (.10, .15, .26, .50, 1) & 196,608 & 32K, 128K & interleaved thinking, retained & FP8 weights \\
\midrule
Qwen3-4B (release control) & 4.0B & GQA, 32 Q / 8 KV & 36 / 36 & as primary & 32,768; 131,072 with YaRN & 32K & hybrid; thinking off & BF16 \\
\bottomrule
\end{tabular}
\end{table*}

\paragraph{Contexts and decoding.}
Two context conditions are defined by their total-context cap, which bounds prompt and generation together. All cross-model comparisons use a common 32K cap, which every checkpoint supports natively. The 128K cap tests allocation when the candidate set is four times larger, and it is run only where the checkpoint supports it: natively for Qwen3-4B-Instruct-2507, Gemma-3-4B-it, and MiniMax-M2, and for Qwen3-8B only through the YaRN extension its model card describes, which we identify as an extension and apply only at 128K. Qwen3-1.7B and the original Qwen3-4B are evaluated at 32K only. Every cap reserves a fixed allowance for generation. Full episodes start from the original task state and run under the cap; continuations of long agent histories isolate scheduling under long state (Appendix~\ref{app:longctx}). Each checkpoint uses its own chat template and tool-call format, and its reasoning and decoding settings are fixed across all methods compared on it. The hybrid Qwen3 checkpoints run with thinking disabled, matching the primary model's non-thinking mode. MiniMax-M2 runs in its native interleaved-thinking mode and keeps every earlier \texttt{<think>} block in the history, as its model card requires, with the card's sampling parameters and its own history format. The Qwen and Gemma checkpoints decode greedily.

\paragraph{Per-backbone sparse attention.}
Each backbone receives its own injected indexer ($H_{\idx}=8$ score heads, separate from the backbone's key-value heads, and $k=2048$), acceptance tests, schedule library, normalizations, and controller; nothing trained on one backbone is reused on another. MiniMax-M2 runs through an attention adapter in its serving engine with speculative and multi-token decoding disabled (Appendix~\ref{app:backbone-adapt}). A hosted API cannot be used, because \method changes attention internally.

\subsection{Evaluation populations and fitting data}
\label{sec:tasks}
\label{sec:splits}

Table~\ref{tab:benchmarks} lists the five final evaluation populations. SWE-bench Verified carries the primary claim \citep{jimenez2024swebench,openai2024sweverified}. Terminal-Bench 2.0 \citep{merrill2026terminalbench} and the retail, airline, and telecom domains of $\tau^2$-bench \citep{barres2025tau2} test whether a controller fitted on repository repair transfers unchanged to terminal work and to dual-control customer service. The benchmark releases, task lists, and scoring definitions are fixed to those in Table~\ref{tab:benchmarks}, and every reported score names its release, because later releases change tasks and scoring under the same names: $\tau^2$-bench v1.0.0, for example, keeps the database check for retail but adds LLM-judged natural-language assertions to 112 of its 114 retail tasks, drops the required-output check there, and edits the gold actions or user scenarios of some retail and airline tasks with unchanged identifiers. Comparisons and pooling stay within one release (Appendix~\ref{app:tau2}).

\begin{table*}[htbp]
\centering
\scriptsize
\caption{Final evaluation populations. Counts are the published size and the size of the pinned task list; the pinned count is authoritative for analysis. Releases and scoring definitions are fixed.}
\label{tab:benchmarks}
\begin{tabular}{P{0.13\textwidth}P{0.10\textwidth}P{0.21\textwidth}P{0.25\textwidth}P{0.15\textwidth}}
\toprule
Population & Tasks (published / pinned) & Release, harness, and scorer & Endpoint & Role \\
\midrule
SWE-bench Verified & 500 / 500 & Pinned dataset revision; official \texttt{swebench} harness (version and image digests fixed) & Official resolved rate: all FAIL\_TO\_PASS and PASS\_TO\_PASS tests pass & Primary claim; repair controller \\
Terminal-Bench 2.0 & 89 / 89 & Harbor, dataset \texttt{terminal-bench@2.0} & Official task-verifier success & Transfer \\
$\tau^2$ retail & 115 / 114 & \texttt{tau2-bench} v0.2.0, base tasks & pass@1 and pass$^4$ over 4 trials & Transfer \\
$\tau^2$ airline & 50 / 50 & \texttt{tau2-bench} v0.2.0, base tasks & pass@1 and pass$^4$ over 4 trials & Transfer \\
$\tau^2$ telecom & 114 / 114 & \texttt{tau2-bench} v0.2.0, base tasks, default dual control & pass@1 and pass$^4$ over 4 trials & Transfer \\
\bottomrule
\end{tabular}
\end{table*}

\paragraph{$\tau^2$ trials.}
In $\tau^2$-bench an LLM-simulated user converses with the agent; in telecom the user also has its own tools and acts on a shared environment, so solving a task requires coordinating actions on both sides \citep{barres2025tau2}. We keep this dual-control setting and fix the user-simulator checkpoint and configuration across all compared methods. Each task is run four times with the same deployed policy. pass@1 is the mean one-trial success over the four trials. pass$^4$, the four-trial reliability, requires success in all four trials of that policy; it is stricter than the conventional pass@4, which counts a task as solved if any of four attempts succeeds. The randomness of a trial has four separate sources, each with its own fixed seed stream: the controller-training seed of the selected policy, the controller's sampling at evaluation, the agent's own decoding when it samples, and the user simulator.

\begin{figure*}[tbp]
\centering
\begin{tikzpicture}[
  air figure, x=0.93cm, y=0.93cm,
  stage/.style={draw=black!30, fill=white, rounded corners=3pt, text width=31.35mm, minimum height=12.09mm, align=center, inner sep=2.7pt},
  blue/.style={stage, draw=refreshblue, fill=softblue},
  green/.style={stage, draw=agentgreen, fill=softgreen},
  orange/.style={stage, draw=warmorange, fill=softorange},
  violetbox/.style={stage, draw=violet, fill=softviolet},
  arrow/.style={-{Latex[length=2mm]}, draw=black!65, line width=0.5pt}
]
\node[blue] (indexer) at (2.10,1.85) {indexer preparation\\dense teacher to frozen sparse selector};
\node[orange] (library) at (6.85,1.85) {schedule calibration\\one nested mask per budget};
\node[green] (traces) at (11.60,1.85) {agent data collection\\causal states and fixed continuations};
\node[violetbox] (router) at (9.25,-0.65) {controller training\\supervised warm start and on-policy RL};
\node[blue] (test) at (4.45,-0.65) {group-disjoint evaluation\\verifier success and net cost};
\draw[arrow] (indexer) -- (library);
\draw[arrow] (library) -- (traces);
\draw[arrow] (traces.south) -- (11.60,0.55) -| (router.north);
\draw[arrow] (router) -- (test);
\node[draw=lossred, fill=softred, rounded corners=2pt, minimum width=31.62mm, minimum height=5.58mm] at (6.40,0.55)
  {freeze backbone, indexer, $k$, masks};
\node[draw=lossred, fill=softred, rounded corners=2pt, minimum width=38.13mm, minimum height=5.58mm] at (6.85,-1.85)
  {freeze features, policy, dual, RNG, evaluator};
\end{tikzpicture}
\caption{Experimental pipeline per backbone: separate fitting partitions train the indexer, calibrate the masks, and fit and select the controller, each frozen before the next stage.}
\label{fig:experiment-pipeline}
\end{figure*}
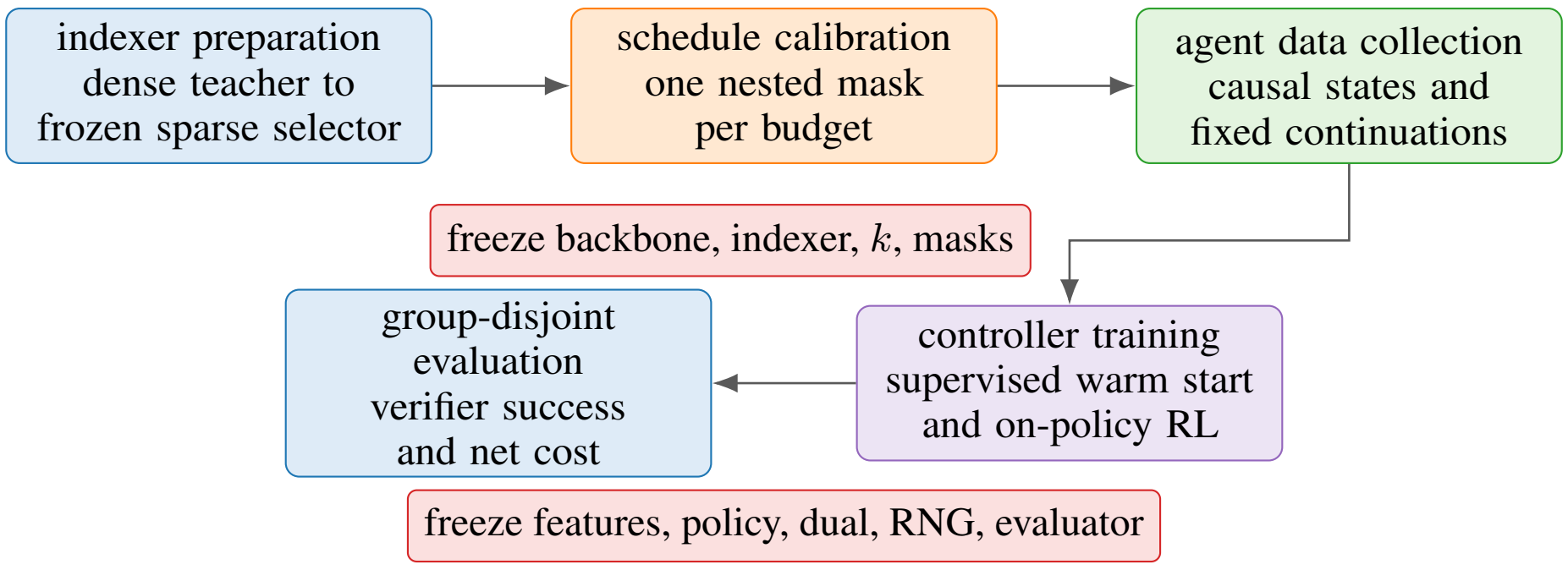

\paragraph{Fitting pipeline.}
For each backbone, fitting proceeds through indexer preparation, indexer acceptance, schedule calibration, selection of the static reference, warm-start fitting, constrained reinforcement learning, held-out controller selection, and one frozen final evaluation (Figure~\ref{fig:experiment-pipeline}). In the main protocol, which evaluates a repair-trained policy for transfer on the full populations, every stage uses fitting data, and the five final populations are used only for the final evaluation. The in-domain studies (Section~\ref{sec:transfer}) are the one exception: they train and select their controllers on designated training and validation subsets of Terminal-Bench 2.0 and of the $\tau^2$ domains, and they are evaluated only on the held-out subsets.

The repair fitting pool combines SWE-Gym and SWE-smith \citep{pan2025swegym,yang2025swesmith}. After mapping every task to its canonical upstream repository and owner, we remove every task whose owner, or a former name of that owner, owns a SWE-bench Verified repository, keep Python tasks with a nonempty issue statement, and remove SWE-smith tasks duplicated in SWE-Gym. The resulting list of eligible tasks has 37,968 tasks from 122 repositories and 115 owners, and the owners serve as split groups for five partitions (Appendix~\ref{app:splits}, Table~\ref{tab:partitions}). Disjointness from Verified rests on this owner-level filter; SWE-bench's own training split is not used.

The repair pool supplies fitting data for repository repair only. Terminal and tool-use cells also need fitting data for their pre-evaluation checks and for method qualification. These data come from a source disjoint from Terminal-Bench 2.0 and from the $\tau^2$ base tasks, which remain reserved, in full, for transfer evaluation. In the transfer protocol, final benchmark data never influence indexer training, schedule search, feature or cost normalization, heuristic thresholds, controller fitting, check outcomes, or checkpoint selection.

\paragraph{Checks before evaluation.}
A comparison of refresh schedules means little if the injected sparse attention has already lost most of the backbone's ability, or if a task is so easy or so hard that every schedule scores the same. Three checks on fitting data therefore decide, before any final run, which cells of the backbone-by-population matrix are evaluated. The sparse-attention check requires that the backbone's indexer pass its acceptance tests and that all-refresh sparse execution stay within a fixed margin of native dense execution on fitting validation. The capability check requires that the static reference's success on fitting validation lie within a fixed interval, so that a floor or ceiling does not decide the comparison. The infrastructure check requires that the chat template, tool-call parsing, harness, serving backend, and memory at the context cap be qualified. Every checkpoint in Table~\ref{tab:models} is evaluated on all five populations at each of its caps. A cell that fails the sparse-attention or the infrastructure check is not evaluated. A cell that fails only the capability check is evaluated and reported descriptively, outside the confirmatory and secondary tests. Every failed cell stays in the reported matrix, marked with the check it failed.

\subsection{Comparisons}
\label{sec:baselines}

The comparisons are arranged so that each one isolates one ingredient of \method. Appendix~\ref{app:comparisons} is the authoritative inventory of every arm, with the components it shares, its calibration, its added costs, and the question it answers (Tables~\ref{tab:baselines}, \ref{tab:ablations}, \ref{tab:methods}, and~\ref{tab:methods-cost}); Table~\ref{tab:t17_comparison_roles} summarizes the tracks.

Native dense execution runs the unmodified backbone and measures how much task quality the injected sparse attention keeps. All other arms share the injected indexer, $k$, the sink and local sets, and the cost model. Static IndexCache runs one library pattern per episode in two runtimes: a dynamic-ready runtime that maintains index keys at every eligible layer, as \method does, and a specialized runtime that omits them at permanently Shared layers; the strongest feasible specialized policy is the primary comparator. The controls for placement, state-free timing, rules, local learning, and restricted state, the alternative selectors, the composition experiments, and the retrained-backbone track are defined in Appendix~\ref{app:comparisons}.

\paragraph{Qualification before evaluation.}
Every method follows the same sequence. It is first implemented on the primary model and checked for correctness and compatibility with fitting data from repository repair, terminal, and tool-use tasks, the latter two from the disjoint fitting source of Section~\ref{sec:tasks}. The checks of Section~\ref{sec:tasks} then run on fitting data, and the extension of each qualified method to the other backbones is decided from its compatibility with the backbone's attention layout and its resource cost (Appendix~\ref{app:extension}). Only after these steps is each method evaluated, once, on SWE-bench Verified, Terminal-Bench 2.0, and the three $\tau^2$ domains. Within a comparison, the agent scaffold, checkpoint, decoding, population, and resource limits are held fixed. A method that changes the attended support or the model parameters is compared on end-to-end success and resource use, with the change reported, because the indexing cost of Equation~\ref{eq:index-cost} covers only the refresh decision.

\begin{table}[t]\centering\small
\setlength{\tabcolsep}{3pt}\renewcommand{\arraystretch}{1.14}
\begin{tabular}{>{\raggedright\arraybackslash}p{2.0cm}>{\raggedright\arraybackslash}p{4.2cm}>{\raggedright\arraybackslash}p{2.6cm}>{\raggedright\arraybackslash}p{4.2cm}}
\toprule
Track & Methods & Extension rule & What must be controlled \\
\midrule
Shared indexer & Anchor, static, periodic, staggered; mixture, temporal, event, staleness, supervised, bandit, AIR & Every accepted backbone & Frozen backbone and indexer; paired budget and state controls \\
Alternative indexer & Training-aware IndexCache; LatentIndex; MISA; HISA; LongCat CLI / HI & Every compatible injected indexer & Separate calibration; key, pooling, routing and training costs \\
Temporal adaptation & ReTopK with injected indexer & Every compatible injected indexer & Exact prefill; decode reuse; cache invalidation on prefix rebuild \\
Native-score methods & Published ReTopK; HShare; SFI; TidalDecode; Quest; Kascade & Qwen family; other families after adapter check & Different selector; own scores, caches and complete runtime charged \\
AIR composition & AIR with HISA, MISA-4 or LongCat HI & Where both components qualify & Own all-refresh reference, static library and fitted controller \\
Retrained backbone & Full LongCat Sparse Attention & Primary model only & Continual training and changed support; upfront cost disclosed \\
Related architecture & CLA; YOCO; You Only Index Once; HySparse / HySparse2; MiniMax Sparse Attention & Related work only & Different pretrained checkpoint; no frozen-backbone score assigned \\
\bottomrule\end{tabular}

\caption{Comparison tracks: each answers its own question and holds different factors fixed; read numerical rows within their track.}\label{tab:t17_comparison_roles}
\end{table}

\subsection{Transfer and in-domain studies}
\label{sec:transfer}

Repository repair is the only domain with a large fitting pool, so the main question about other domains is transfer. The main transfer experiment selects one policy per backbone on repair validation tasks and evaluates it unchanged on Terminal-Bench and on all three $\tau^2$ domains. Transfer evaluation uses the cost normalization $C_0$ and the budget $\overline B$ fixed on repair-training data, so a transfer result measures how well a repair-trained allocation rule carries over to terminal work and to dual-control customer service. Separate in-domain studies then ask how much of any gap comes from the training domain. They train terminal and tool-use controllers on training and validation tasks drawn from the benchmarks themselves (Appendix~\ref{app:indomain}). The pinned benchmarks provide no training split, so these studies partition the benchmark tasks by group and report only the held-out subset, with its size. Their results are labelled as in-domain results on held-out subsets, and they are reported apart from unchanged-policy transfer, which covers the full populations.

\subsection{Endpoints and the primary comparison}
\label{sec:endpoints}

\paragraph{Endpoints.}
Quality is the population's official success measure (Table~\ref{tab:benchmarks}), reported as an absolute rate and as a percentage-point difference. Efficiency is reported at four levels that are kept apart: the indexing and control work per attempt from Section~\ref{sec:cost}; full-model work and model-only inference time; end-to-end time per attempt, with tool, environment, and user-simulator time shown separately; and prefill latency, decode throughput, peak accelerator memory, and cache memory. Input and output tokens, tool calls, and episode length are reported alongside. Cost per success divides the total measured cost of all attempts, failed ones included, by the number of successful attempts; when a method has no success, the ratio is reported as undefined. Reductions in index work are reported in their own units, separately from measured wall-clock time.

\paragraph{Three static selections.}
Three selections involve static policies, and each has its own role. The development static reference $b_0$ is chosen on the static-selection partition before any routing label is built; it anchors the warm-start labels and the capability check, and it gives a secondary contrast. The deployed controller is chosen on controller validation by the procedure of Section~\ref{sec:rl} and frozen before final evaluation; when no learned checkpoint is feasible, it is the static fallback. The analysis comparator is chosen on the final sample itself: the static library entry with the highest success among those within the cost tolerance, run in the specialized runtime, and reselected inside every bootstrap replicate. Because it sees the final sample, the analysis comparator favors the static side, which makes the primary test conservative. Table~\ref{tab:t24_reference_roles} lists the three selections for every backbone and cap.

\begin{table}[t]\centering\small
\setlength{\tabcolsep}{4pt}\renewcommand{\arraystretch}{1.14}
\begin{tabular}{lrrrrrr}
\toprule
Backbone & Cap & $b_0$ & Dev. \% & Dev. cost & Deployment & Final R values \\
\midrule
Qwen3-1.7B & 32K & R=14 & 9.4 & 0.485 & Learned & 14 \\
Qwen3-4B-Instruct-2507 & 32K & R=18 & 23.0 & 0.485 & Learned & 18, 36 \\
Qwen3-4B-Instruct-2507 & 128K & R=18 & 23.0 & 0.485 & Learned & 18 \\
Qwen3-8B & 32K & R=18 & 30.6 & 0.485 & Learned & 18 \\
Qwen3-8B & 128K & R=18 & 30.6 & 0.485 & Learned & 18 \\
Gemma-3-4B-it & 32K & R=2 & 12.2 & 0.388 & Static R=2 & 2, 3 \\
Gemma-3-4B-it & 128K & R=2 & 12.2 & 0.388 & Learned & 2, 3 \\
MiniMax-M2 & 32K & R=31 & 57.0 & 0.485 & Learned & 31 \\
MiniMax-M2 & 128K & R=31 & 57.0 & 0.485 & Learned & 31 \\
Qwen3-4B (original) & 32K & R=18 & 18.1 & 0.485 & Learned & 18 \\
\bottomrule\end{tabular}

\caption{The three static selections come from different data: the development reference $b_0$ (static-selection partition), the deployed policy (validation), and the analysis comparator (each final population; last column).}\label{tab:t24_reference_roles}
\end{table}

\paragraph{The primary comparison.}
The primary claim concerns Qwen3-4B-Instruct-2507 on SWE-bench Verified at the 128K cap and compares the deployed controller with the analysis comparator. \method can succeed in two ways: it can reach higher success at matched cost, or it can match success, within a noninferiority margin, while spending clearly less. Let $\Delta_V$ and $\Delta_{\overline C}$ be the differences in expected success and normalized cost between the deployed controller and the analysis comparator, with one-sided confidence bounds $L_V$ (lower, on $\Delta_V$) and $U_C$ (upper, on $\Delta_{\overline C}$) at level $1-\alpha/2$. The adaptive claim holds if either preregistered rule holds,
\begin{equation}
\begin{aligned}&\text{(i)}\ \ L_V>0\ \text{and}\ U_C\leq\epsilon_{\mathrm{match}},\\ &\text{(ii)}\ \ L_V\geq-\epsilon_V\ \text{and}\ U_C\leq-\epsilon_{\mathrm{save}},\end{aligned}
\label{eq:decision-rules}
\end{equation}
where rule (i) is higher success at matched cost and rule (ii) is noninferior success with a minimum saving; splitting $\alpha$ between the rules bounds the familywise error by $\alpha$. Tasks are the resampling unit; repeated trials and controller seeds stay inside their task's cluster. The same comparison is computed for every other evaluated cell as a secondary analysis, with Holm adjustment within each predeclared family (populations for one backbone; backbones for one population). Training curves of verifier reward, validation success, normalized cost, constraint violation, the dual price, budget frequencies, and controller overhead are reported per backbone and labelled as repair training or in-domain training (Appendix~\ref{app:protocol}).

\section{Experimental Protocol Details}
\label{app:protocol}

\subsection{Per-backbone settings}
\label{app:backbone-settings}

\paragraph{Contexts.}
A context cap $L_{\mathrm{cap}}$ bounds prompt plus generation for one episode and includes a fixed generation reserve $L_{\mathrm{reserve}}$. The common cross-model cap is $L_{\mathrm{cap}}=32{,}768$. The 128K cap, $L_{\mathrm{cap}}=131{,}072$, is used for Qwen3-4B-Instruct-2507 (262,144 native), Gemma-3-4B-it (131,072 native), and MiniMax-M2 (196,608 native), and for Qwen3-8B only with the YaRN rope scaling of factor 4 described on its model card; the YaRN setting is labelled as a context extension and applied only in that model's 128K runs, so its 32K and 128K results come from two configurations of the same weights. Qwen3-1.7B, whose card states a 32,768-token context and no extension, and the original Qwen3-4B release control run at 32K only. Each backbone's candidate-set size $|\mathcal E_t|$ grows with $t$ up to its cap, so at 32K the fixed $k=2048$ covers a larger fraction of the prefix than at 128K.

\paragraph{Templates, tools, and reasoning.}
Each checkpoint uses the chat template shipped with its pinned revision. All four Qwen3 templates emit tool calls as JSON objects inside \texttt{<tool\_call>} tags and receive tool results as \texttt{<tool\_response>} blocks in a user turn; the templates of Qwen3-1.7B, Qwen3-4B, and Qwen3-8B are identical, and that of Qwen3-4B-Instruct-2507 differs only in how it handles thinking. We parse this format with the Hermes-style parser that the Qwen documentation names, since the model cards name none. The hybrid Qwen3 checkpoints run with thinking disabled through the template. This matches the primary model, which supports only non-thinking mode, and follows their cards' instruction to leave thinking content out of the history. MiniMax-M2's template emits XML tool calls inside \texttt{<minimax:tool\_call>} tags and returns tool results in dedicated tool turns; it runs with the model's own reasoning and tool-call parsers, keeps every earlier \texttt{<think>} block in the history in its original form, as its model card requires, and samples with the card's temperature (1.0), top-$p$ (0.95), and top-$k$ (40). Gemma-3-4B-it has no thinking mode, and its chat template, as reported by public mirrors of the gated repository, has no tool role and requires strict alternation of user and model turns; Google's documentation describes tool use through prompting. Its tool calls therefore use a fixed text protocol: tool schemas are described in the first user turn, the model answers with a Python-style function call in a fenced block, which a deterministic parser extracts, and tool results are returned in the next user turn. The parse-failure rate of this protocol is part of the infrastructure check. The Qwen and Gemma checkpoints decode greedily. Within a checkpoint, every compared method uses the same template, parser, reasoning setting, and decoding configuration.

\subsection{Full episodes and long-history continuations}
\label{app:longctx}

\paragraph{Full episodes.}
A full episode starts from the original task state and runs under the cap of its context condition until the task ends or a limit is reached. Full episodes measure end-to-end success and resource use, and they are the unit of the primary comparison.

\paragraph{Long-history continuations.}

Most full episodes spend little time near the context cap, so continuations of long agent histories are evaluated as well, to isolate scheduling under long agent-generated state. A cap condition $c$ has the total-context cap $L_{\mathrm{cap}}(c)\in\{32{,}768,131{,}072\}$ that its 32K or 128K label names. A continuation must leave backbone $m$'s fixed generation reserve $L_{\mathrm{reserve}}(m)$ free, so its prefix holds at most $L_{\mathrm{use}}(m,c)=L_{\mathrm{cap}}(c)-L_{\mathrm{reserve}}(m)$ tokens, counted with $m$'s tokenizer. The long-history band for $(m,c)$ is $\left[\lceil\varphi\,L_{\mathrm{use}}(m,c)\rceil,\ L_{\mathrm{use}}(m,c)\right]$, where the fraction $\varphi\in(0,1)$ is fixed before construction. A frozen behavior agent runs each eligible fitting or evaluation task with all-refresh sparse execution under the declared tools and limits and the cap of condition $c$, and we retain the first nonterminal controller boundary whose serialized causal prefix length $L$ lies in the band. A trajectory that terminates first, or whose prefix grows past the band between two boundaries, yields no snapshot, and the number of such trajectories is reported for each condition. The rule is defined when $0<L_{\mathrm{reserve}}(m)<L_{\mathrm{cap}}(c)$ and $0<\varphi<1$; the band is then nonempty, because $\lceil\varphi\,L_{\mathrm{use}}\rceil\leq L_{\mathrm{use}}$ for every positive integer $L_{\mathrm{use}}$. Both $L_{\mathrm{reserve}}(m)$ and $\varphi$ are set on fitting data before any evaluation history is constructed. Snapshots are built once, by the frozen behavior agent, and the same snapshots serve every compared method.  The snapshot stores the transcript, sandbox state, remaining tool and token budgets, environment and user-simulator random state, and split-group ID. Population construction reads only prefix length and termination; verifier output, candidate losses, and later task outcomes play no role.

Every compared policy starts from the same snapshot. An all-refresh burn-in rebuilds the core key-value cache and the index-key cache over the common prefix, and $\Tcache$ is empty before the first evaluated query. Burn-in work is identical across policies, reported separately, and excluded from the post-branch cost contrast. From the branch point onward, each policy controls every model-call and $Q$-token boundary until termination. The runtime inserts no filler tokens, duplicates no tool output, and preserves the original event order. A condition with fewer eligible trajectories than the fixed minimum fails this population check and is not evaluated; the construction rule stays fixed. All snapshots from one source task inherit its split group, so a shorter branch cannot enter training while a longer branch of the same task enters evaluation.

\subsection{Fitting data and partitions}
\label{app:splits}

\paragraph{Sources and eligibility.}
The pool is built from SWE-Gym (Hugging Face revision \texttt{bb94ed9e}, 2,438 tasks from 11 repositories) and SWE-smith (revision \texttt{ea6d7173}, 59,136 tasks from 222 repositories) \citep{pan2025swegym,yang2025swesmith}; SWE-Gym-Lite is a subset of SWE-Gym and adds no tasks. Every task is first mapped to its canonical upstream \texttt{owner/repository}. SWE-smith stores tasks under its own mirror namespace (\texttt{swesmith/owner\_\_repo.commit}); we recover the upstream owner and repository from each mirror name, confirm them against a GitHub snapshot, which also resolves renamed repositories, and group tasks by that upstream owner. The SWE-bench Verified revision \texttt{c104f840} contains twelve repositories under twelve owners (astropy, django, matplotlib, mwaskom, pallets, psf, pydata, pylint-dev, pytest-dev, scikit-learn, sphinx-doc, sympy).

Each task receives at most one exclusion reason, assigned in the order below, and SWE-Gym loses no task under any rule. Owner-level matching is case-insensitive and removes 4,449 SWE-smith tasks from eight repositories owned by Verified owners (\texttt{django/channels}, \texttt{django/daphne}, \texttt{pallets/click}, \texttt{pallets/jinja}, \texttt{pallets/markupsafe}, \texttt{pydata/patsy}, \texttt{pylint-dev/astroid}, \texttt{pytest-dev/iniconfig}); no training repository coincides with a Verified repository itself. Two historical namespaces are excluded as whole owner groups because one repository in each moved to a Verified owner: \texttt{PyCQA/pylint} became \texttt{pylint-dev/pylint}, which the GitHub API confirms by redirect, and \texttt{kennethreitz/requests} became \texttt{psf/requests}, which the API no longer confirms because the old name now belongs to a 2021 fork of \texttt{psf/requests}. Neither moved repository appears in the sources; the exclusion removes 488 tasks from the other repositories in those namespaces, \texttt{PyCQA/flake8} and \texttt{kennethreitz/records}. Because SWE-smith also contains other languages, we keep repositories whose tasks mostly modify Python files, as in Verified, which removes 8,228 tasks. SWE-smith tasks that match a SWE-Gym task by repository and pull-request number, identical issue text, or identical patch are duplicates; 347 are dropped in favor of the SWE-Gym copies. Finally, 10,094 SWE-smith tasks whose issue statement is empty or whitespace-only are dropped, because an agent cannot be posed such a task.

The eligible pool contains 37,968 tasks: 2,438 from SWE-Gym (11 repositories, 11 owners) and 35,530 from SWE-smith (120 repositories, 113 owners), together 122 repositories under 115 owners, since some repositories appear in both sources. Eligibility depends on task metadata only.

\paragraph{Container qualification.}
Before a partition is used, each task must also pass a container check under the pinned harnesses. The task's environment image must be available, the reference fix must make every FAIL\_TO\_PASS and PASS\_TO\_PASS test pass, and an inert patch, which leaves the bug in place, must leave at least one FAIL\_TO\_PASS test failing. Each task's check runs once. A task is run a second time, under a lighter load, when its reference-fix run leaves no parsed test results, including a timeout at the harness's 90-second limit for a test run, or when that run passes every FAIL\_TO\_PASS test and fails only PASS\_TO\_PASS tests; the second outcome is final. One property of the pinned SWE-smith harness excludes tasks for a reason that lies outside the task. After applying a patch, the harness restores the files named by the task's test identifiers to their original version, and for a doctest that name is a source module. When the reference fix edits such a module, the harness undoes the fix before the tests run, and the task fails its reference check. A static check of the frozen data finds 896 such candidates among the 35,530 eligible SWE-smith tasks. Candidates that fail their reference check are excluded. Those tasks are excluded like every other task that fails the check, which keeps the fitting pool consistent with the harness that will score it, and the summary reports them as a separate failure class next to the candidate count. Tasks that fail are dropped without reassigning owners, and retained counts are kept per partition and source, apart from the eligibility counts.

\paragraph{Partitions.}
The canonical owner is the split group. This paper fixes the assignment: complete owner groups go to five partitions with target task shares of 20\% indexer training, 10\% schedule calibration, 10\% static selection, 45\% controller training, and 15\% controller validation, applied both to the whole pool and within each source. Each owner is described by its pair of SWE-Gym and SWE-smith task counts. Owners are placed in decreasing order of size, with ties broken by a permutation drawn with seed 20261007. Each owner goes to the partition that minimizes the sum over partitions of the squared deviations of the overall share and of each source's share from the target, each divided by the target. A deterministic local search over single-owner moves and pairwise swaps then lowers this objective from 0.0029 to 0.0006. Table~\ref{tab:partitions} gives the result. An earlier rule assigned each owner by its majority source alone; it placed pandas-dev (737 SWE-Gym and 1,716 SWE-smith tasks) by its SWE-smith count and left controller training and validation with 18\% and 6\% of SWE-Gym. The vector rule brings every partition within 0.7 percentage points of its target in each source.

\begin{table*}[tbp]
\centering
\small
\caption{Repair-pool partitions by owner group (seed 20261007). Counts are eligible tasks. Shares are of the whole pool and of each source.}
\label{tab:partitions}
\setlength{\tabcolsep}{4pt}
\begin{tabular}{lrrrrrrrr}
\toprule
 & & \multicolumn{3}{c}{All tasks} & \multicolumn{2}{c}{SWE-Gym} & \multicolumn{2}{c}{SWE-smith} \\
\cmidrule(lr){3-5}\cmidrule(lr){6-7}\cmidrule(lr){8-9}
Partition & Target & Tasks & Repos & Owners & Tasks & Share & Tasks & Share \\
\midrule
Indexer training & 20\% & 7,592 & 24 & 23 & 484 & 19.9\% & 7,108 & 20.0\% \\
Schedule calibration & 10\% & 3,788 & 13 & 12 & 228 & 9.4\% & 3,560 & 10.0\% \\
Static selection & 10\% & 3,801 & 12 & 12 & 251 & 10.3\% & 3,550 & 10.0\% \\
Controller training & 45\% & 17,087 & 52 & 51 & 1,101 & 45.2\% & 15,986 & 45.0\% \\
Controller validation & 15\% & 5,700 & 21 & 17 & 374 & 15.3\% & 5,326 & 15.0\% \\
\midrule
Total & & 37,968 & 122 & 115 & 2,438 & & 35,530 & \\
\bottomrule
\end{tabular}
\end{table*}

Whole-owner grouping still limits source diversity. SWE-Gym has only eleven owners, so each partition's SWE-Gym tasks come from one to three of them, and the 374 SWE-Gym tasks in controller validation all come from Project-MONAI/MONAI. Controller validation therefore reports success and normalized cost separately for each source alongside the pooled values. The pooled bounds of Appendix~\ref{app:rl} remain the selection criterion, and the SWE-Gym portion of validation is reported as a single-repository diagnostic. No owner or repository appears in more than one partition, and validation tasks never share an owner with training tasks. These guarantees concern the repair pool and SWE-bench Verified; the terminal and tool-use benchmarks are separated from fitting data by the requirement stated in Section~\ref{sec:tasks}. An MBPP-derived repair suite remains a development suite, used only for instrumentation and leakage tests \citep{austin2021program}.

\paragraph{Per-backbone partitions.}
The same group assignment is used for every backbone, but each backbone collects its own behavior trajectories, fits its own indexer, library, feature and cost normalization, and controller, and selects its own checkpoint. The cost normalization $C_0$ of a backbone is measured on its repair controller-training partition and is then used unchanged for that backbone's transfer evaluation. Every mutation, environment seed, replay row, and continuation snapshot inherits its source group.

\subsection{Transfer and in-domain studies}
\label{app:indomain}

\paragraph{Transfer.}
Each backbone's transfer policy is the repair policy selected on repair controller validation. It is evaluated unchanged on Terminal-Bench 2.0 and on the three $\tau^2$ domains, with the repair cost normalization $C_0$ and the repair budget $\overline B$, and with no tuning on the target population.

\paragraph{In-domain studies.}
Two further studies ask what training in the target domain adds, and their results are reported apart from transfer.

The terminal study partitions the 89 Terminal-Bench 2.0 tasks into training, validation, and held-out groups, stratified by task category, with a fixed seed. Results are reported on the held-out subset, labelled with its size. The tool-use study trains a separate controller for each $\tau^2$ domain. The pinned release has no training split. Release v1.0.0 publishes one (74/40 for retail and telecom, 30/20 for airline) whose identifiers all occur in the pinned v0.2.0 base tasks and together cover them exactly. We therefore apply these identifier lists to the pinned v0.2.0 task definitions and scorer, carve validation tasks from the training identifiers by group, and report results on the 40, 20, and 40 held-out tasks under the v0.2.0 scorer. For telecom, the pinned release also contains a larger generated task set beyond the base tasks, which may supplement training after its disjointness from the held-out identifiers is verified. Each in-domain study measures its own cost normalization on its training partition, and each in-domain policy is evaluated only on its held-out tasks; the transfer results cover the full populations.

\subsection{\texorpdfstring{$\tau^2$}{tau2}-bench release and scoring}
\label{app:tau2}

All $\tau^2$ results use release v0.2.0 (tag commit \texttt{f8de30c}, 2025-10-06) and its base task sets of 114 retail, 50 airline, and 114 telecom tasks; the paper's table lists 115 retail tasks, and the pinned task list is authoritative. In v0.2.0, retail and airline tasks carry no explicit reward basis, so the default product of a database check (the final database must equal the one produced by the gold actions) and a required-output check applies, and telecom tasks are scored by environment assertions (94 tasks) or environment assertions plus actions (20 tasks). In v1.0.0 (tag \texttt{17e07b1}, 2026-03-18), 112 retail tasks are scored by the database check together with natural-language assertions judged by an LLM, of which only 40 tasks carry nonempty assertions, and the remaining 2 retail tasks by the database check alone; airline keeps the database and required-output checks, and telecom's base-task scoring is unchanged. The same release also edits gold actions or user scenarios for some retail and airline tasks without changing their identifiers. Because these changes alter both what is scored and some tasks themselves, results are labelled with their release, and comparisons stay within one release. The v0.2.0 default user model is the unversioned alias \texttt{gpt-4.1} at temperature 0; we pass the dated checkpoint \texttt{gpt-4.1-2025-04-14} at temperature 0 explicitly, which is the simulator used in the benchmark paper and pinned by later releases, and keep it fixed for every method.

\subsection{Checks before evaluation}
\label{app:gates}

The three checks are run per backbone and population on fitting data, before any final run, and their outcomes are recorded in the evaluation matrix.
The \emph{sparse-attention check} requires that the backbone's indexer pass the acceptance tests of Appendix~\ref{app:indexer} and that all-refresh sparse execution stay within a fixed success margin of the native model on fitting validation from the corresponding domain.
The \emph{capability check} requires that the static reference's success on fitting validation lie within a fixed interval $[p_{\min},p_{\max}]$; a cell outside the interval is reported descriptively, without a confirmatory contrast.
The \emph{infrastructure check} requires a qualified chat template and tool-call parser with a fixed maximum parse-failure rate, a harness that reproduces the benchmark's reference scoring on its provided reference solutions where they exist, a serving backend that reproduces reference outputs (for MiniMax-M2, the attention-adapter parity tests of Appendix~\ref{app:indexer}: dense bypass against the unmodified engine and all-refresh sparse execution against the sparse reference implementation), and enough memory to run the backbone at its cap.
If the primary cell itself fails a check, the failure is reported and the primary claim moves to the predeclared fallback, the primary model's in-domain $\tau^2$ retail study, evaluated on its held-out tasks.

\subsection{Endpoints and reporting}
\label{app:endpoints}

\paragraph{Success.}
SWE-bench Verified counts an instance as resolved when all of its FAIL\_TO\_PASS and PASS\_TO\_PASS tests pass under the official harness. Terminal-Bench 2.0 uses the task verifier's binary outcome. In $\tau^2$, a trial succeeds when the pinned scorer's reward, the product of the components listed for the task, equals one. For a task with $n=4$ trials of one trained policy and $c$ successes, pass$^k$ is estimated by $\binom{c}{k}/\binom{n}{k}$ and averaged over tasks, so the $k=1$ case is the mean success rate, which we write pass@1, and pass$^4$, the all-four success, is the fraction of tasks solved in every trial. The four trials of a task use four controller-evaluation seeds and four user-simulator seeds and always the same controller checkpoint, so pass$^4$ measures success in all four trials of one trained policy. When results are reported for several controller-training runs, pass$^k$ is first computed for each run and then averaged across runs. Scores are compared only within the pinned benchmark release. Success is reported as an absolute rate and as a percentage-point difference from each comparator.

\paragraph{Resource use.}
For every attempt we record the indexing and control work of Section~\ref{sec:cost}; full-model work; model-only inference time; end-to-end time, split into model time and tool, environment, and user-simulator time; prefill latency and decode throughput; peak accelerator memory and the sizes of the core key-value and index-key caches; input and output tokens; tool calls; and episode length. Cost per success and index-work reductions follow the definitions of Section~\ref{sec:endpoints}.

\paragraph{Primary comparator and decision rules.}
After final evaluation, we choose $b^\star_D$, the static library entry with the highest success on the final sample among those whose mean normalized cost is at most $\overline B+\epsilon_C$, run in the specialized static runtime, with ties broken toward lower cost and then fewer Full layers. Because this choice sees the final sample, it favors the static baseline; every bootstrap replicate repeats it. The analogous differences against $\pi_0$, the reference fixed during development, form a secondary preregistered contrast. Before final evaluation, we freeze $1-\alpha$, $\epsilon_V$, $\epsilon_{\mathrm{match}}$, and $\epsilon_{\mathrm{save}}$; each decision rule in Equation~\ref{eq:decision-rules} is an intersection of two one-sided conditions, and allotting $\alpha/2$ to each rule keeps the familywise error of their disjunction at most $\alpha$. Both contrasts are repeated with measured end-to-end time as the cost axis as secondary analyses. Every endpoint includes controller and probe overhead.

\subsection{Statistical aggregation}
\label{app:statistics}

Task-level comparisons pair environment images, initial states, decoding configuration, limits, and task order, and they keep the four random streams of Appendix~\ref{app:environment} separate. On SWE-bench Verified and Terminal-Bench 2.0 each task is run once, with one controller-evaluation seed; in $\tau^2$ each task has four trials of the deployed policy, which first yield the task's pass@1 and pass$^4$ contributions. Reward and cost are then averaged within a task before resampling, so the task is the resampling unit and repeated trials and seeds stay in their task's cluster.

On SWE-bench Verified, the bootstrap resamples tasks. Verified has only twelve repositories, one of which contributes 46\% of the instances, so repository-clustered resampling rests on a coarse cluster structure and serves only as a sensitivity analysis. On Terminal-Bench 2.0 tasks are resampled directly, and in each $\tau^2$ domain tasks are resampled with their four trials. Validation bounds on the fitting pool use the owner-clustered bootstrap. Policy fitting, normalization, schedule search, dual tuning, and controller runs happen once, before resampling, and every replicate reuses their results. Secondary cells use Holm adjustment within two predeclared families: the five populations for one backbone, and the backbones evaluated on one population. Variation across the $S_{\mathrm{RL}}$ training runs is reported for the validation candidates; the final contrast concerns the single selected policy.

\subsection{Training diagnostics}
\label{app:diagnostics}

For every backbone and training domain we report, against rollout count, the terminal verifier reward of training episodes, success on controller validation, the mean normalized cost and its excess over $\overline B$, the dual price $\lambda$, the frequency of each budget, and the controller's overhead as a share of indexing cost. Curves are labelled by training domain: the repair curves belong to the policies evaluated for transfer, and the terminal and tool-use curves belong to the separate in-domain studies.

\subsection{Evaluation matrix}
\label{app:matrix}

All five checkpoints and the release control are evaluated on all five populations at the 32K cap, and the three 128K-capable checkpoints plus Qwen3-8B with YaRN at the 128K cap. Each cell records the outcomes of the three pre-evaluation checks before final evaluation. A cell enters the confirmatory and secondary analyses only if all three checks pass; a cell that fails only the capability check is evaluated and reported descriptively, a cell that fails the sparse-attention or infrastructure check is not evaluated, and every failed cell is reported with the check it failed. Table~\ref{tab:evaluation-matrix} lists the questions isolated by each comparison.

\begin{table*}[tbp]
\centering
\small
\caption{Comparisons and the question each isolates. Rows apply to every evaluated backbone and population unless stated otherwise.}
\label{tab:evaluation-matrix}
\begin{tabular}{P{0.27\textwidth}P{0.29\textwidth}P{0.34\textwidth}}
\toprule
Comparison & Held fixed & Question \\
\midrule
Native vs.\ all-refresh sparse & Backbone, prompts, tools, limits & Does the injected sparse attention preserve agent capability? \\
All-refresh vs.\ static IndexCache & Backbone, indexer, $k$, runtime & How much indexer work does static reuse remove, and at what quality? \\
Dynamic-ready vs.\ specialized static runtime & Masks and model outputs & What does keeping all index keys cost? \\
Anchor-only, periodic, staggered vs.\ greedy static & Refresh count & Does layer placement matter within a budget? \\
Static vs.\ mixture and temporal schedule & Action frequencies or time sequence & Does varying the budget without state help? \\
Heuristics vs.\ supervised router & State, menu & Do simple rules capture the allocation signal? \\
Supervised router vs.\ bandit vs.\ \method & State, menu, budget & What do chosen-action feedback and delayed credit add? \\
\method vs.\ strongest feasible static & Everything except the scheduler & Primary contrast (Equation~\ref{eq:decision-rules}) \\
\method vs.\ alternative indexers and caches & Backbone, harness, population & End-to-end success and resource frontiers \\
Flat/HISA $\times$ static/\method & Final attended budget & Does allocation help after cheaper selection? \\
Repair-trained policy on terminal and $\tau^2$ & Policy, normalization, budget & Does allocation transfer unchanged? \\
In-domain vs.\ transferred policy (held-out subsets) & Held-out tasks & What does in-domain training add? \\
Across backbones at 32K & Cap, populations, protocol & Does the result depend on scale, family, or MoE? \\
Qwen3-4B vs.\ Qwen3-4B-Instruct-2507 & Size and architecture & How much of a Qwen scale difference is release? \\
32K vs.\ 128K cap & Backbone, selector, menu & Does lower selected-key density change refresh value? \\
\bottomrule
\end{tabular}
\end{table*}

\section{Comparison Protocols}
\label{app:comparisons}

The comparisons fall into four groups, which differ in what they hold fixed and therefore in how a difference can be interpreted: refresh controls on the shared indexer (Appendix~\ref{app:baselines}), alternative indexers and cache representations (Appendix~\ref{app:external}), composition experiments (Appendix~\ref{app:composition}), and architectural and retrained-backbone comparisons (Appendix~\ref{app:architectural}). For each method outside the first group, we describe its mechanism, how we adapt it to our injected indexer or backbones, the state and cost it adds, and the qualification test it must pass. Source versions, code revisions, and differences between papers and their released code are collected in Appendix~\ref{app:code}.

\subsection{Refresh controls on the shared indexer}
\label{app:baselines}

Table~\ref{tab:baselines} lists every controlled comparison and Table~\ref{tab:ablations} every ablation. Unless a table says otherwise, a method shares the backbone, injected indexer, $k$, sink and local sets, agent harness, prompts, tool schemas, decoding configuration, limits, and cost model of its backbone, and differs only in how many and which eligible layers refresh. Calibration and training use the fitting partitions of Appendix~\ref{app:splits}; in the transfer protocol nothing is tuned on a final population, and in an in-domain study only that study's designated training and validation subsets are used. Every controller and heuristic is charged for its own feature extraction and for any probe it runs. These controls answer the paper's central question at matched indexer cost, and only they enter the decision rules of Equation~\ref{eq:decision-rules}.

Several entries need more detail. Anchor-only execution keeps only $\ell_1$ Full, so each query receives one fresh selection and no refresh deeper in the network; it is the $R=1$ pattern and bounds the indexer cost from below. The periodic depth mask places Full layers at a fixed interval within a query, and the fixed temporal schedule varies the budget over time, cycling through a fixed sequence of budgets at segment boundaries without reading the state. The lagged-staleness heuristic raises the budget when the overlap between the selections of the two most recent completed Full queries falls below a threshold; because it uses only completed computation, it needs no extra selector work, and a variant with a paid probe on the new tokens counts the probe in its cost. For Gemma, whose menu has four entries on five global layers, all five static refresh counts $R\in\{1,\ldots,5\}$ also run as static controls. The deployed controller's static fallback (Appendix~\ref{app:rl}) and the strongest feasible static comparator of the primary analysis are separate policies, selected on different data (Section~\ref{sec:endpoints}).

The phase-only ablation reads the phase from observable event metadata at the segment boundary: the tool name, whether the last tool output contained a traceback or a test summary, the output length, and whether a new file was opened. The model-only, index-only, model-plus-index, and full-state ablations add the feature blocks of Appendix~\ref{app:features} in that order. The remaining ablations vary the decision interval $Q$, replace the greedy nested library with uniform, staggered, or independently searched patterns, sweep the budget ratio $\beta$, and add feature blocks one at a time with their counted overhead.

The adaptive claim fails if the selected policy satisfies neither decision rule. Every endpoint includes controller and probe overhead, so overhead that removes the saving makes the rule fail. Whether one repository or task family accounts for the difference is a sensitivity analysis, reported with leave-one-repository-out and per-family contrasts, and it adds no rejection criterion beyond the decision rules. If supervised routing matches \method, an adaptive result can stand while the delayed-credit claim is rejected.

\begin{table*}[ht]
\centering
\footnotesize
\caption{Native reference and refresh controls on the shared indexer. ``Fixed'' lists what is held constant beyond the shared setup described in the text.}
\label{tab:baselines}
\setlength{\tabcolsep}{4pt}
\begin{tabular}{P{0.17\textwidth}P{0.22\textwidth}P{0.14\textwidth}P{0.18\textwidth}P{0.19\textwidth}}
\toprule
Method & What changes & Fixed & Calibration or training & Question \\
\midrule
Native backbone & Unmodified attention, no indexer & Harness, decoding & None & Does the injected sparse attention preserve agent success? \\
All-refresh sparse & Every eligible layer Full & Indexer, $k$ & None & Upper end of refresh cost \\
Anchor-only ($R=1$) & Only $\ell_1$ Full & Indexer, $k$ & None & Is any depth refresh needed? \\
Periodic depth mask & Full layers at a fixed interval, same $R_b$ & Refresh count & None & Does greedy placement matter? \\
Staggered depth mask & Full layers at seeded random positions, same $R_b$ & Refresh count & Preregistered mask seeds & Is the greedy pattern better than chance? \\
Static IndexCache & One greedy library pattern for the episode & Library & Greedy calibration search & Static reuse at each budget \\
Static runtime variants & Dynamic-ready vs.\ specialized runtime & Pattern, outputs & None & Cost of keeping all index keys \\
Strongest feasible static & Best specialized static entry within tolerance & Selection rule & Reselected on the final sample and in every bootstrap replicate & Primary comparator \\
State-independent mixture & Budgets sampled from fixed frequencies & Marginal action rates & Frequencies frozen on controller validation & Does state matter beyond the action mix? \\
Fixed temporal schedule & Budget follows a fixed time sequence & No state input & Sequence fixed on fitting data & Does timing without state help? \\
Event-triggered refresh & Budget raised after tool events & Thresholds & Thresholds set on fitting data & Do simple event rules suffice? \\
Lagged-staleness threshold & Budget raised when completed-query overlap drops & Threshold, probe cost & Threshold set on fitting data & Does a cheap staleness signal suffice? \\
Supervised router & Warm-start policy, frozen & Features, menu & Counterfactual replay labels & Value of local prediction alone \\
Contextual bandit & Policy trained on immediate replay feedback & Features, menu & Replay rewards, priced cost & Value of chosen-action feedback \\
\method & On-policy constrained policy gradient & Features, menu & Terminal reward under the budget & Value of delayed terminal credit \\
\bottomrule
\end{tabular}
\end{table*}

\begin{table*}[ht]
\centering
\footnotesize
\caption{Controller and library ablations. Each varies one factor of \method and keeps the rest of its recipe.}
\label{tab:ablations}
\setlength{\tabcolsep}{4pt}
\begin{tabular}{P{0.17\textwidth}P{0.22\textwidth}P{0.14\textwidth}P{0.18\textwidth}P{0.19\textwidth}}
\toprule
Method & What changes & Fixed & Calibration or training & Question \\
\midrule
Phase / model / index / model+index / full state & Controller inputs & Training recipe & As \method & Which signals carry allocation value? \\
Decision interval & $Q$ and model-call-only decisions & Menu & As \method & Responsiveness vs.\ overhead \\
Library construction & Uniform, staggered, or independently searched patterns & Refresh counts & As \method & Outer policy vs.\ inner schedule \\
Budget sweep & Ratio $\beta$ & Everything else & As \method & Shape of the trade-off \\
Overhead & Feature blocks added one at a time & Everything else & As \method & Does added overhead remove the saving? \\
\bottomrule
\end{tabular}
\end{table*}

\subsection{Alternative indexers and cache representations}
\label{app:external}

These methods keep the frozen backbone and change how a selection is computed, what is cached, or which positions are attended. Each runs under the same agent harness, prompts, decoding, limits, and final populations as \method. Because they change the selector or the attended support, the indexer-only matched-cost rule does not apply to them; it is reserved for comparisons on the shared indexer. We compare them with \method on end-to-end success, the complete indexing cost including their own state maintenance, full-model compute, model-only and end-to-end latency, index-cache bytes, and peak memory. Prefill and decode are measured separately for every method, and any context length at which a method starts to pay off comes from our own profiling at the 32K and 128K caps on our hardware and backbones.

Every adaptation passes a qualification test on the primary model. Each reimplementation reproduces the published algorithm's selections on reference inputs, or, where the published method runs on a different selector, the documented relation to our flat indexer (for example, exact equality of selected sets below a length threshold). All calibration uses fitting data only.

\paragraph{Training-aware IndexCache.}
This comparison tests whether adapting the indexers to one fixed pattern matches a learned allocation over unadapted indexers. Starting from the frozen backbone and the accepted indexer, multi-layer distillation (Equation~\ref{eq:multi-layer}) adapts one indexer bank per evaluated pattern on indexer-training data \citep{bai2026indexcache}. This frozen-backbone variant is ours; the original method also updates the remaining weights with the language-modeling loss. Because it changes indexer parameters, we report it separately from the shared-indexer controls, together with its adaptation compute and the memory of keeping one bank per pattern. We reimplement the loss on our injected indexer.

\paragraph{LatentIndex.}
This comparison tests whether sharing a compressed index-key representation across layers, while keeping a separate selection at every layer, gives a better success-to-cost trade-off than refreshing selections only at some layers. For each group of consecutive layers, LatentIndex caches one latent per token computed from the group's first layer, folds each layer's key decoder into that layer's query, and keeps layer-specific head weights and top-$k$ selection \citep{wang2026latentindex}. Its training-free variant fits the latent maps in closed form by reduced-rank ridge regression onto each layer's native indexer keys, and its hierarchical variant lets the group's first layer keep a candidate set from which the other layers select. Our backbones have no native indexer, so the regression targets are the index keys of our accepted injected indexer, and calibration uses our fitting data; the paper calibrates on general-text corpora. Groups are formed over consecutive eligible layers; for Gemma, the five global layers form one group of four and one of one. Because the method changes the indexer representation, we compare success, complete indexing cost, latency, index-cache bytes, and peak memory. Our reimplementation reproduces the paper's closed-form fit and its layer-specific selections on reference inputs.

\paragraph{Head routing (MISA).}
This comparison tests whether activating fewer indexer heads per query reduces the cost of each selection enough to rival refreshing fewer layers. MISA scores each indexer head with block-pooled index keys and keeps only the highest-scoring heads per query \citep{zhou2026misa}. Our indexer has $H_{\idx}=8$ heads, so we preregister active-head settings $\{1,2,4,8\}$, with eight active heads as the control that reproduces the flat indexer. The router's pooling and scoring and the maintenance of pooled keys are counted. Our implementation follows the paper's router, which scores each head by its gate-weighted mean affinity to block-pooled index keys (Eq.~7 of the paper), in both prefill and decode, and it keeps pooled keys as a persistent per-block cache; the released decode path differs from the paper (Appendix~\ref{app:code}). Savings reported for native DSA indexers were measured with 32 or 64 heads, so we measure the savings for our eight-head indexer directly.

\paragraph{Hierarchical selection (HISA).}
This comparison measures how much a cheaper block-then-token search changes success and cost by itself. HISA ranks blocks of mean-pooled index keys, keeps a fixed number of blocks, always including the first and last, and runs token-level scoring only inside them \citep{xu2026hisa}. With $k=2048$ and the paper's defaults (block size 128, 64 kept blocks), the result is dense for $t\leq k$, equals flat selection for $k<t\leq 8192$, and approximates it beyond. We follow version~4 of the paper, which describes HISA as approximating DSA and reports per-layer overlap with flat selection of about 93--95\% that falls to about 79\% near 18K positions, and we report selection overlap with the flat indexer by context length. The pooled block keys are a persistent per-layer cache appended block by block, and their maintenance is counted. We reimplement HISA over our indexer's per-head keys, and the reimplementation reproduces flat selection exactly for $t\leq 8192$.

\paragraph{Similarity-guided temporal reuse (ReTopK).}
This comparison tests whether reusing supports across decoding steps when queries are similar can match \method's trade-off, which reuses supports only across depth. ReTopK keeps, for each attention head, a small FIFO cache of recent normalized queries and their selected supports. A new query retrieves its most similar cached queries, merges their supports with the most recent tokens, and reranks the union with exact query-key scores; it falls back to exact top-$k$ when the best similarity is below a threshold, and it refreshes exactly every fixed number of decode steps \citep{yao2026retopk}. We run two versions. The published version works per attention head on the backbone's own query-key scores, as in the paper. The adapted version replaces exact query-key scoring with our injected indexer so that it shares \method's selector, and it is labelled as an adaptation. Both keep candidate reranking, the similarity fallback, and the periodic refresh, with the paper's default settings fixed before any run. The method is defined for decoding. During prefill of a tool output, both ReTopK versions run exact selection, and the newly appended positions enter later supports through the recent window, reranking, and refresh; \method and the other schedules keep their own prefill execution, in which the segment's mask decides which layers select. The query-support cache stays valid while the prefix is append-only, because cached supports refer to fixed positions; when the frozen cache-rebuild rule changes or truncates the prefix, the cache is cleared and reseeded. Costs include similarity lookup, cache storage, candidate scoring, reranking, fallbacks, and periodic refreshes. We reimplement both versions and qualify them before use.

\paragraph{LongCat Sparse Attention components.}
LongCat Sparse Attention combines three components that require different amounts of training \citep{zan2026longcat}. Cross-layer indexing shares one layer's indexing result with the next layers of a fixed group and relies on cross-layer distillation of the indexer. As a targeted comparison, we apply it to our injected indexer with fixed groups of two eligible layers and distill the indexer while keeping the backbone frozen; this tests whether a uniform, trained sharing structure matches learned allocation over a frozen indexer. Hierarchical indexing is training-free: it ranks pages of tokens by sub-block mean keys before token-level scoring, and it is evaluated alone and in composition with \method where its page structure is compatible with our cache layout. Streaming-aware indexing changes the attended support to a fixed sink and sliding window plus selected middle positions and needs training of the model; it belongs to the full method, which we treat as a retrained-backbone comparison (Appendix~\ref{app:architectural}). Public code targets LongCat's own DSA indexer, so on our backbones every component runs through an adapter to our injected indexer, and we implement the distillation and hierarchical indexing ourselves (Appendix~\ref{app:code}).

\paragraph{Other adapted methods.}
HShare shares critical-token indices across layers, heads, and consecutive decoding steps with a configuration chosen per input batch after prefill \citep{wu2025hshare}; Slow-Fast Inference refreshes a reused token set near sentence boundaries and after a fixed refresh budget \citep{xie2026slowfast}; TidalDecode selects tokens with exact scores at a few fixed layers \citep{yang2025tidaldecode}; Quest selects key-value pages by query-aware bounds \citep{tang2024quest}; and Kascade reuses exact top-$k$ indices from anchor layers chosen on a development set \citep{deshmukh2025kascade}. These methods select with the backbone's own attention scores, and each runs in its published form where its released code supports the backbone, with an adapter otherwise (Appendix~\ref{app:code}). Kascade's anchor selection and head mapping are recomputed on our fitting data, and HShare's sharing ratio and SFI's maximum refresh interval are fixed before final evaluation.

\subsection{Composition experiments}
\label{app:composition}

The composition experiments test whether learned refresh allocation remains useful after each selection has become cheaper. The main composition crosses two selectors, the flat injected indexer and HISA, with two schedulers, the strongest feasible static schedule and \method. For each selector we also run all-refresh execution as a reference, so the experiment has six arms per backbone and population. The final number of attended positions, $k$ plus the sink and local positions, is the same in every arm. Each selector receives its own static library, calibrated on fitting data with its own costs, and its own \method controller, trained and selected under that selector's cost model; the HISA controller therefore learns how much a refresh costs when selection is hierarchical. \method with HISA is our proposed combination, and neither paper reports it. Compositions of \method with MISA head routing and with LongCat hierarchical indexing follow the same design where the qualification tests pass.

\subsection{Architectural and retrained-backbone comparisons}
\label{app:architectural}

Some methods cannot run on a frozen checkpoint. Full LongCat Sparse Attention combines its three components with a changed attended support and requires training of the model \citep{zan2026longcat}. We treat it as a retrained-backbone comparison on the primary model only, run after continual training whose data, compute, and changes to the attended support are disclosed, and compare it on end-to-end success and resource use. Architectures that share key-value tensors across layers or are pretrained with sparse attention, such as Cross-Layer Attention, YOCO, You Only Index Once, HySparse and HySparse2, and MiniMax Sparse Attention \citep{brandon2024cla,sun2024yoco,sun2026yoio,gao2026hysparse,wei2026hysparse2,lai2026msa}, require different pretrained checkpoints and are discussed as related work only.

\subsection{Qualification and extension to other backbones}
\label{app:extension}

Every method in this appendix is first qualified on Qwen3-4B-Instruct-2507 with fitting data from repository repair, terminal, and tool-use tasks, the latter two from the fitting source of Section~\ref{sec:tasks}, which is disjoint from the transfer populations. A fixed rule then decides whether the method extends to the other backbones, and the method is evaluated on SWE-bench Verified, Terminal-Bench 2.0, and the three $\tau^2$ domains. Shared-indexer controls run on every backbone whose sparse-attention check passes. Methods that operate inside our injected indexer (training-aware IndexCache, LatentIndex, head routing, HISA, and the LongCat indexing components) extend to every such backbone, on Gemma restricted to its global layers, and on MiniMax-M2 within its compute allocation. Methods that score with the backbone's own attention (ReTopK in its published form, HShare, Slow-Fast Inference, TidalDecode, Quest, and Kascade) extend to Qwen3-1.7B and Qwen3-8B, which share the primary model's attention layout, and to Gemma and MiniMax-M2 through adapters for their attention modules. Full LongCat Sparse Attention runs on the primary model only.

\subsection{Source versions and code availability}
\label{app:code}

Table~\ref{tab:methods} records, for each alternative method, the source version, the code we start from, and whether we reproduce the published method or adapt it; Table~\ref{tab:methods-cost} records its calibration or training, its persistent state and invalidation rule, and the extra cost it is charged. Where released code is missing, covers only part of a method, or differs from the paper, the details are as follows.

\emph{Training-aware IndexCache.} The released code patches SGLang and vLLM for natively sparse DSA models only \citep{indexcache_code}, so the multi-layer loss is our reimplementation.

\emph{LatentIndex.} The linked MuLabPKU/TransArch repository contains no LatentIndex code, so the method is our reimplementation.

\emph{MISA.} In the released vLLM patch (TransArch, directory \texttt{MISA\_\allowbreak preprint}, commit \texttt{ec9e6cb}), prefill recomputes block means on every call, and the one-stage decode path has pooled routing commented out and routes heads by the current token's own index key. We use the released code only to check prefill behavior.

\emph{HISA.} The paper's claim changed across revisions: version~1 stated that HISA preserves the identical token-level top-$k$ pattern, whereas version~4 states that it preserves the top-$k$ index interface; the arXiv abstract page still carries the earlier wording. The released kernels and serving integration target DSA models \citep{xu2026hisa}.

\emph{ReTopK.} The paper gives no code link.

\emph{LongCat Sparse Attention.} Public code covers the components unevenly across the released model \texttt{meituan-longcat/\allowbreak LongCat-Flash-Lite-Sparse} (revision \texttt{35de83a}), SGLang pull request \#32918, and SGLang main at \texttt{9ddbba5}. For model inference, SGLang main contains the LongCat sparse-attention paths, and pull request \#32918 adds the fixes the Lite-Sparse checkpoint needs; we implement the three components on our injected indexer. For streaming-aware indexing, SGLang main contains the selection rule that forces the first 16 and last 1,024 positions into the selected set; we implement the fused sparse and sliding-window attention kernel and the training code ourselves. For cross-layer indexing, SGLang main contains inference-time reuse (every second layer recomputes, set by \texttt{cli\_factor}=2), and we implement the cross-layer distillation and training ourselves. We implement hierarchical indexing from the paper's description. All of this code targets LongCat's own DSA indexer.

\emph{Native-score methods.} Released code for Slow-Fast Inference supports the Qwen3 family, including Qwen3-4B-Instruct-2507, and released code for Kascade provides head mappings for Qwen3-8B; the released code for HShare targets Llama and Mistral checkpoints, and that for TidalDecode and Quest targets Llama-family checkpoints; each needs an adapter for Qwen3's attention modules \citep{sfi_code,kascade_code,hshare_code,tidaldecode_code}.

\begin{table*}[tbp]
\centering
\footnotesize
\caption{Alternative-method documentation. ``Ours'' marks a reimplementation or adaptation. Table~\ref{tab:methods-cost} continues with calibration, persistent state, and counted cost.}
\label{tab:methods}
\setlength{\tabcolsep}{4pt}
\begin{tabular}{P{0.17\textwidth}P{0.15\textwidth}P{0.36\textwidth}P{0.22\textwidth}}
\toprule
Method & Source & Code we start from & Reproduction \\
\midrule
Training-aware IndexCache & arXiv 2603.12201 v1 & THUDM/\allowbreak IndexCache 08d22d69 (DSA models only) & Ours, on injected indexer \\
LatentIndex & arXiv 2610.04635 v1 & None in the linked TransArch repository & Ours; targets are our indexer's keys \\
MISA (head routing) & arXiv 2605.07363 v1 & TransArch, directory \texttt{MISA\_\allowbreak preprint} (decode router differs from paper) & Ours, paper's router \\
HISA & arXiv 2603.28458 v4 & TileLang example \texttt{dsa\_\allowbreak hisa}; SGLang branch \texttt{hisa\_\allowbreak pr} (DSA models) & Ours, on injected indexer \\
ReTopK & arXiv 2607.27692 v1 & No public code & Published per-head version and indexer-based adaptation, both ours \\
LongCat cross-layer indexing & arXiv 2608.01662 v2 & Inference reuse in SGLang main \texttt{9ddbba5}; distillation code unpublished & Adapter to our indexer and distillation, ours \\
LongCat hierarchical indexing & arXiv 2608.01662 v2 & No public code & Ours \\
HShare & ICLR 2025 & wuhuaijin/\allowbreak HShare 523bea0c (Llama, Mistral) & Adapter for Qwen3 \\
Slow-Fast Inference & arXiv 2603.12038 & LV-NUS/\allowbreak SFI 1415706f (vLLM, Qwen3) & Published code on Qwen3 \\
TidalDecode & ICLR 2025 & DerrickYLJ/\allowbreak TidalDecode 81878af2 (Llama-3) & Adapter for Qwen3 \\
Quest & ICML 2024 & mit-han-lab/\allowbreak Quest (Llama family) & Adapter for Qwen3 \\
Kascade & arXiv 2512.16391 v2 & microsoft/\allowbreak kascade d4463fcb (Qwen3-8B mappings) & Published code; mappings refit \\
\bottomrule
\end{tabular}
\end{table*}

\begin{table*}[tbp]
\centering
\footnotesize
\caption{Alternative-method documentation, continued from Table~\ref{tab:methods}: calibration or training, persistent state and its invalidation, and the extra cost each method is charged.}
\label{tab:methods-cost}
\setlength{\tabcolsep}{4pt}
\begin{tabular}{P{0.17\textwidth}P{0.25\textwidth}P{0.25\textwidth}P{0.23\textwidth}}
\toprule
Method & Calibration or training & Persistent state and invalidation & Extra cost counted \\
\midrule
Training-aware IndexCache & Multi-layer distillation per pattern on indexer-training data & One indexer bank per pattern; caches as flat indexer & Adaptation compute; bank memory \\
LatentIndex & Closed-form reduced-rank ridge fit on fitting data & One latent per token per layer group; append-only & Latent projections, latent cache, per-layer selection, candidate rescoring (hierarchical) \\
MISA (head routing) & None; active heads $\{1,2,4,8\}$ preregistered & Block-pooled keys, appended per block & Router scoring and pooled-key maintenance \\
HISA & None & Block-pooled keys, appended per block & Block scoring, pooled-key maintenance \\
ReTopK & None; defaults fixed in advance & Per-head FIFO of queries and supports; cleared on prefix change & Similarity lookup, cache storage, candidate scoring, fallbacks, refreshes \\
LongCat cross-layer indexing & Cross-layer distillation of the indexer & As flat indexer & Distillation compute \\
LongCat hierarchical indexing & None & Page and sub-block mean keys, appended & Page scoring and pooled-key maintenance \\
HShare & Sharing configuration per input batch after prefill; ratio fixed in advance & Shared critical-token indices across layers, heads, steps; full KV kept & Configuration search, approximate scoring \\
Slow-Fast Inference & None; interval fixed in advance & Reused token set refreshed at boundaries & Dense refresh steps, support construction \\
TidalDecode & Selection layers fixed on fitting data & Token set reused across later layers per step & Exact scoring at selection layers \\
Quest & None & Per-page min/max key metadata & Page bounds and scoring \\
Kascade & Anchors and head mapping on fitting data & Anchor indices reused across layers & Exact anchor attention \\
\bottomrule
\end{tabular}
\end{table*}

\section{Agent Runtime, Logging, and Random Streams}
\label{app:environment}
\label{app:runtime}

Schedules can be compared only if every arm faces the same task under the same conditions and leaves a record that can be replayed. The runtime, the logged record, and the random streams below are shared by every method.

\paragraph{Environment and agent loop.}
Each task starts from a versioned environment image and a fixed initial state. The agent follows a ReAct-style loop that interleaves reasoning with tool actions \citep{yao2023react}. Coding tasks expose file search, file read, patch editing, shell execution, and test tools inside a versioned sandbox; terminal tasks expose the benchmark's shell interface; tool-use tasks expose only the benchmark's domain APIs and policy documents. Every method receives the same system prompt, task prompt, tool schema, context policy, decoding rule, token limits, tool-step limit, wall-time limit, and environment seed; model decoding follows each backbone's fixed rule (Appendix~\ref{app:backbone-settings}). In $\tau^2$-bench, a simulated user with a fixed checkpoint and configuration converses with the agent; in telecom the user also acts through its own tools on a shared environment, and the default dual-control setting is kept for every method \citep{barres2025tau2}.

\paragraph{Cache reuse at decision boundaries.}
At a new model call, the runtime checks the prefix hash against the existing core KV and indexer-key caches. A matching append-only prefix keeps both caches. The runtime prefills only the newly appended system, user, or tool-observation tokens under the newly selected mask and then decodes the first $Q$ tokens under the same mask. A $Q$-token boundary inside a response keeps both caches and selects a new mask for the later decode queries. A changed or truncated prefix follows one frozen cache-rebuild rule for every method, and $\Tcache$ resets for every query position.

\paragraph{Logging and replay.}
Every run records the model revision and weight hash, indexer hash, prompt hash, tool-schema hash, benchmark revision, container digest, decoding configuration, controller seed, timeout, and cost-profiler revision in a run configuration record that is frozen before controller fitting, so replay can separate model-side schedule changes from changes in the task runtime. Every model call logs the hash of the full causal prefix, the incoming cache version, the segment boundary, the chosen budget, the exact layer mask, the selected-index lineage, the generated tokens, the parsed tool action, the environment response, the verifier output, the controller counter, and the per-segment cost breakdown. Hidden tests and future environment state never enter controller features. A replay tool reproduces model-side computation from the prefix hash and checkpoint hashes.

\paragraph{Random streams and pairing.}
Four random streams are kept separate and recorded: controller-training seeds, which index the $S_{\mathrm{RL}}$ training runs; controller-evaluation seeds, which drive the deployed controller's sampling through a counter-based stream keyed by task, trial, and decision index; agent-sampling seeds, used only by checkpoints that sample; and user-simulator seeds in $\tau^2$. Because budget sampling has its own stream, it leaves model decoding unchanged. Paired comparisons share task order, decoding configuration, environment seed, timeout policy, and step limit. The trajectory depends on actions, so generated text generally differs across methods, and common random numbers align only the exogenous streams. The final analysis treats each task as the paired unit.

\section{Reproducibility and Leakage Tests}
\label{app:tests}

Three failure modes could produce an apparent benefit of adaptive allocation without one existing: an execution error that makes schedules differ in ways the cost model does not see, a feature that leaks information from after the decision, and overlap between fitting and evaluation data. The tests below address each in turn.

\paragraph{Execution tests.}
Deterministic unit tests check incremental decoding against full-prefix evaluation under every schedule, as well as absolute offsets, short-prefix selection with fewer than $k$ candidate positions, causal masks, index-buffer overwrite, and Shared-to-Full switches. A switch test confirms byte-identical indexer keys under continuous execution and frozen-prefix replay. Candidate actions at one state share the same prefix and differ only in the selected schedule.

\paragraph{Feature leakage tests.}
Feature tests perturb future observations, terminal reward, the candidate action, the repository path, and the task identity. Valid pre-action features must stay unchanged under future or candidate-action perturbations, and identity perturbations must not reveal an owner or repository through raw paths or text keys.

\paragraph{Split audits and artifact hashes.}
Split audits require zero split-group overlap across indexer training, schedule calibration, static selection, controller training, controller validation, and final evaluation, and zero repository or owner overlap between the repository-repair fitting pool and SWE-bench Verified. Artifact hashes cover the backbone, indexer, schedule library, feature definitions, policy weights, dual variable, prompts, tool schemas, benchmark version, and analysis code. The final evaluator writes one immutable result after all shards pass the population audit.

\section{Extended Limitations}
\label{app:limitations}

These caveats expand the Limitations section without adding results.

\paragraph{Injected sparse attention.} The indexer is injected into dense pretrained backbones that were never trained with sparse attention, so the sparse attention may lose quality that a natively sparse model would keep; a native sparse checkpoint with production kernels would give stronger systems evidence.

\paragraph{Expressiveness and cost.} Index keys at every eligible layer let any layer be promoted and cost memory that static IndexCache avoids, which can reduce the feasible batch size at long contexts; the specialized static runtime omits that memory. The finite library keeps every mask valid by construction but cannot reach patterns outside it.

\paragraph{Learning.} The expected-cost constraint lets individual episodes exceed the budget, while safety-sensitive serving may require hard per-request limits. Terminal reward has high variance; value baselines, the supervised warm start, and grouped batches reduce it, and the method uses no reward shaping. The controller reads tool-interface metadata, which ties the policy to one tool interface; the restricted-state ablations measure this dependence.

\paragraph{Evaluation.} Each setting reports one deployed policy, and variation across the three controller-training seeds enters only through validation selection. Populations of 50 to 114 tasks limit the precision of the $\tau^2$ and Terminal-Bench 2.0 comparisons, continuations cover only the snapshots that reach each occupancy band, and in-domain results cover held-out subsets of 20 to 40 tasks. SWE-bench Verified concentrates nearly half of its instances in one repository, so repository-level conclusions rest on few clusters, and the MBPP-derived development suite is too short for the long-context conditions. Each backbone has its own indexer, library, normalization, and controller, so cross-backbone differences mix the injected sparse attention with allocation. Qwen3-8B reaches 128K only through a context extension, MiniMax-M2 runs through an adapter with FP8 weights, and Gemma-3-4B-it offers only five eligible layers. Results use fixed benchmark releases and one user simulator. Serving-side memory tiering, such as the expert cache and tiered KV storage of Strata \citep{niko1221strata}, lies outside the protocol; combining it with \method would require measuring where the persistent index-key cache lives.

\section{Complete Results}
\label{app:results}
\label{app:ext-results}

These tables and figures hold the evidence behind Section~\ref{sec:results}; each subsection follows the matching results subsection. Unless a caption says otherwise, displays use Qwen3-4B-Instruct-2507 at the 128K cap. Success is the percentage of attempts solved, with the four $\tau^2$ trials of each task averaged; time is mean end-to-end time per attempt, counting model, tool, environment, and user-simulator time for failed and successful attempts alike; speedup divides the all-refresh mean time by the method's mean time on the same backbone, cap, and population; and successful tasks per hour is 3,600 times all successful attempts divided by all elapsed seconds. The comparator is the analysis comparator of Appendix~\ref{sec:endpoints}, reselected for each backbone, cap, and population on the final sample. Displays report point estimates for one deployed policy per setting. Heatmap scales are centered at 0 for differences and at 1 for ratios, with blue for gains and red for losses. Training curves show the three seeds (11, 29, 47) as faint lines and seed means, smoothed with a causal exponential moving average ($\alpha=0.18$), as solid lines. In tables, bold marks the best value in the stated comparison set (ties at the reported precision included), a tint marks \method-based rows, and gray marks unranked references.

\subsection{Primary model at 128K}
\label{app:results-primary}
\label{app:ext-results-primary}

\begin{figure*}[ht]
\centering
\includegraphics{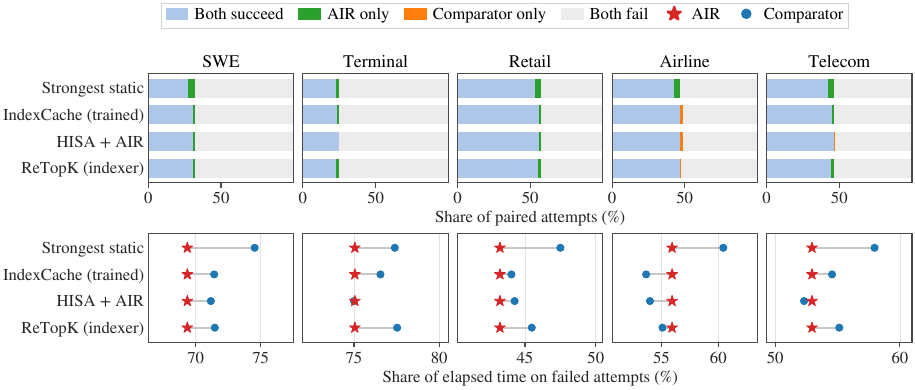}
\caption{Paired by task and trial, \method solves attempts that the strongest static schedule fails and spends a smaller share of its time on failed attempts. Top: outcome shares against each comparator; bottom: share of time spent on failed attempts.}
\label{fig:paired-outcomes}
\end{figure*}

\begin{figure*}[ht]
\centering
\includegraphics{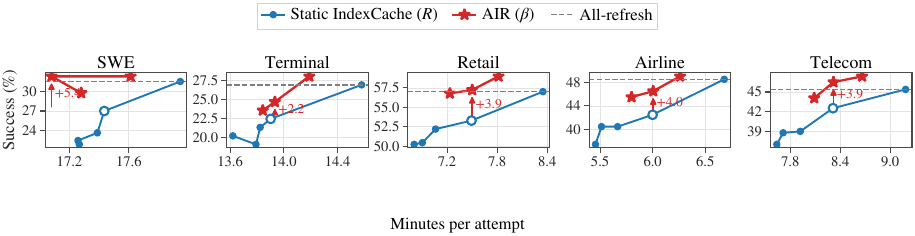}
\caption{At similar time per attempt, \method reaches higher success than static schedules on all five benchmarks (Qwen3-4B-Instruct-2507, 128K). Blue: static $R\in\{3,\dots,36\}$, comparator open; red: \method at $\beta\in\{0.35,0.50,0.70\}$; dashed: all-refresh; arrows: \method minus the static comparator at the deployed budget (pp).}
\label{fig:tradeoff}
\end{figure*}

Validation deploys the learned controller of seed 47 after 12,288 rollouts (Table~\ref{tab:t09_deployment_selection}). Its normalized indexing and control cost is 0.524, against 0.502 for the comparator ($R=18$ in the specialized runtime); both lie within the tolerance $\overline B+\epsilon_C=0.53$, and the controller itself accounts for 0.023 of \method's cost. On SWE-bench Verified, the predeclared test of rule~(i) in Equation~\ref{eq:decision-rules} supports higher success at matched cost (Table~\ref{tab:primary-contrast}). The development reference $b_0$ is also $R=18$ on this backbone and cap (Table~\ref{tab:t24_reference_roles}), so the secondary contrast against it has the same values. Paired by task and trial, \method solves attempts that the static comparator fails and spends a smaller share of its time on failed attempts; against the other three comparators of Figure~\ref{fig:paired-outcomes}, these shares differ by at most 2.4 points, and on airline all three spend less than \method. Raising the budget from $\beta=0.50$ to $0.70$ raises success on four benchmarks, leaves SWE-bench Verified at 32.4\%, and lengthens every episode by 1.8 to 4.2\%; lowering it to $0.35$ costs 0.4 to 2.6 points. At each budget, \method reaches higher success than static schedules with similar time per attempt (Figure~\ref{fig:tradeoff}).

\begin{figure*}[tbp]
\centering
\includegraphics{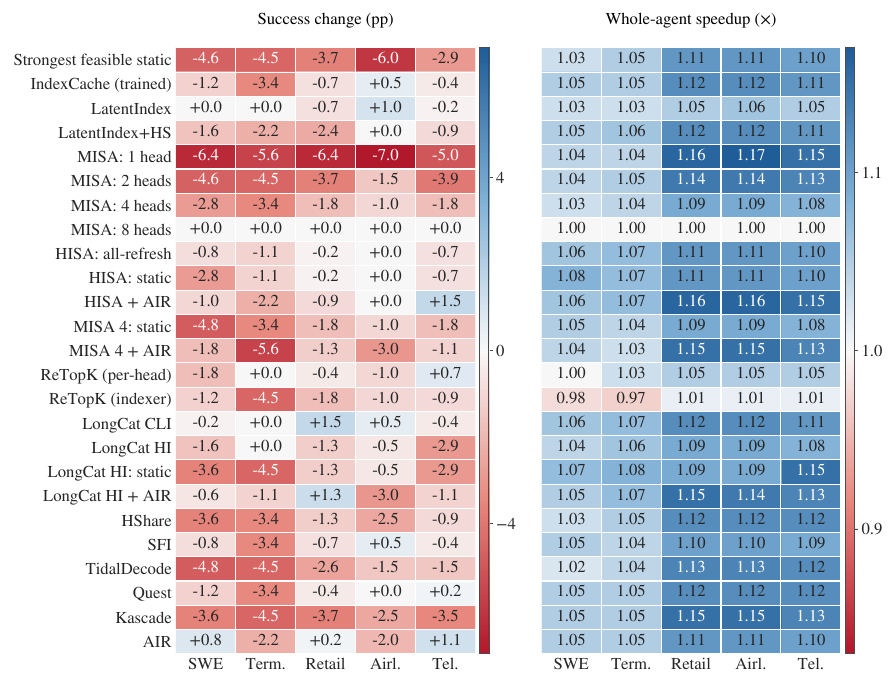}
\caption{Few settings are both faster than all-refresh and at least as successful; on SWE-bench Verified only \method and LatentIndex are. Left: success change (pp); right: speedup over all-refresh.}
\label{fig:quality-speedup}
\end{figure*}

\begin{figure*}[tbp]
\centering
\includegraphics{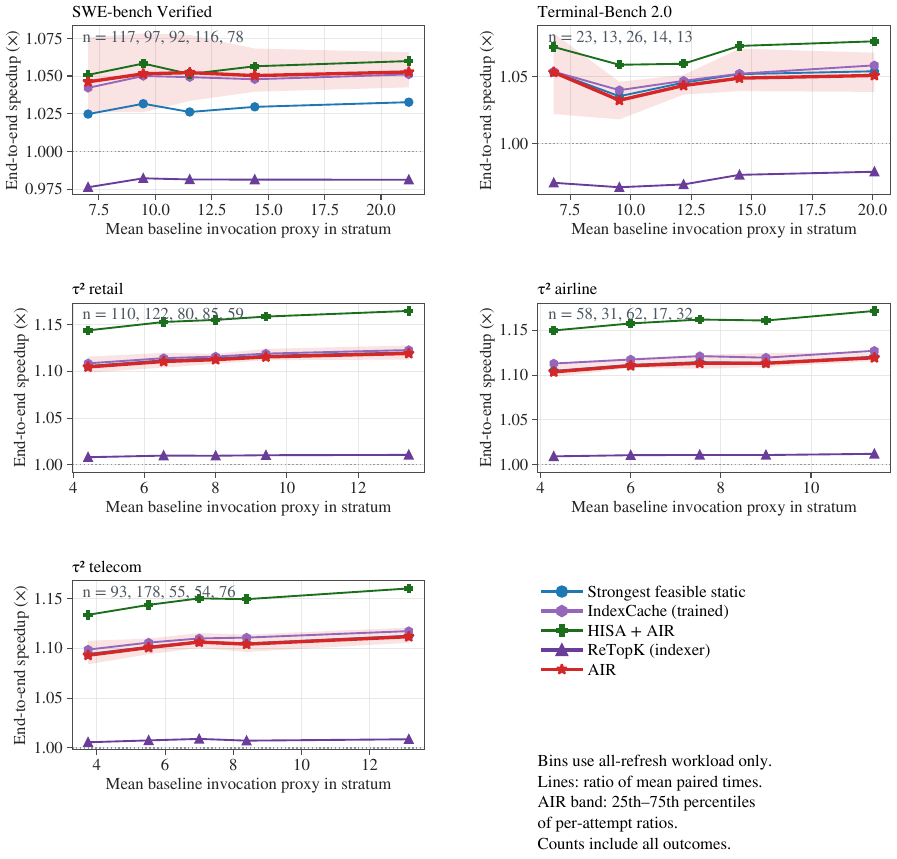}
\caption{\method's speedup over all-refresh changes little with episode length: across five workload strata it stays at 1.03 to 1.05 on SWE-bench Verified and Terminal-Bench 2.0 and 1.09 to 1.12 on $\tau^2$. Ribbon: interquartile range of per-attempt ratios.}
\label{fig:strata}
\end{figure*}

Across all methods (Table~\ref{tab:elapsed}, Figure~\ref{fig:quality-speedup}), \method gains 0.2 to 1.1 points over all-refresh on SWE-bench Verified, retail, and telecom with speedups of 1.05 to 1.11, and loses 2.2 and 2.0 points on Terminal-Bench 2.0 and airline; on SWE-bench Verified only \method and LatentIndex are both faster than all-refresh and at least as successful. HISA and LongCat hierarchical settings are the fastest on SWE-bench Verified and Terminal-Bench 2.0, one-head MISA and HISA with \method are among the fastest on $\tau^2$, and native attention is the slowest everywhere. Across five workload strata, defined by quintiles of the all-refresh episode's tool-call count, \method's ratio of mean times stays at 1.03 to 1.05 on SWE-bench Verified and Terminal-Bench 2.0 and 1.09 to 1.12 on $\tau^2$ (Figure~\ref{fig:strata}). Prefill accounts for about 80\% of model time for every method, so selection work during prefill drives most of the index cost.

\begin{table*}[ht]
\centering
\small
\caption{\method's success gain over the analysis comparator is positive on every population at a cost difference of at most 0.022. $\Delta_V$: success difference (pp); $\Delta_{\overline C}$: normalized cost difference; differences are calculated before rounding.}
\label{tab:primary-contrast}
\setlength{\tabcolsep}{4pt}
\begin{tabular}{lrcrr}
\toprule
Population & Tasks & Comparator & $\Delta_V$ (pp) & $\Delta_{\overline C}$ \\
\midrule
SWE-bench Verified & 500 & $R=18$ & $+5.4$ & $+0.022$ \\
Terminal-Bench 2.0 & 89 & $R=18$ & $+2.2$ & $+0.020$ \\
$\tau^2$ retail & 114 & $R=18$ & $+3.9$ & $+0.017$ \\
$\tau^2$ airline & 50 & $R=18$ & $+4.0$ & $+0.013$ \\
$\tau^2$ telecom & 114 & $R=18$ & $+3.9$ & $+0.017$ \\
\bottomrule
\end{tabular}
\end{table*}

\begin{table}[tbp]\centering\small
\setlength{\tabcolsep}{4pt}\renewcommand{\arraystretch}{1.14}
\begin{tabular}{lrrrrr}
\toprule
Backbone & Cap & Deployed type & Candidate & Success LB \% & Cost UB \\
\midrule
Qwen3-1.7B & 32K & Learned & s47\_n12288 & 6.2 & 0.477 \\
Qwen3-4B-Instruct-2507 & 32K & Learned & s11\_n12288 & 22.0 & 0.481 \\
Qwen3-4B-Instruct-2507 & 128K & Learned & s47\_n12288 & 22.6 & 0.473 \\
Qwen3-8B & 32K & Learned & s11\_n12288 & 27.2 & 0.476 \\
Qwen3-8B & 128K & Learned & s11\_n6144 & 29.0 & 0.488 \\
Gemma-3-4B-it & 32K & \cellcolor{AIRFallback}Static fallback & static\_r2 & 5.7 & 0.412 \\
Gemma-3-4B-it & 128K & Learned & s11\_n12288 & 14.2 & 0.503 \\
MiniMax-M2 & 32K & Learned & s11\_n12288 & 55.9 & 0.490 \\
MiniMax-M2 & 128K & Learned & s47\_n12288 & 55.4 & 0.477 \\
Qwen3-4B (original) & 32K & Learned & s47\_n9216 & 17.9 & 0.493 \\
\bottomrule\end{tabular}

\caption{Validation deploys a learned controller in nine of ten configurations and the static fallback $R=2$ (shaded) for Gemma-3-4B-it at 32K. Feasible: cost upper bound at most 0.53; candidate names give seed and rollout count.}\label{tab:t09_deployment_selection}
\end{table}

\begin{table}[tbp]\centering\footnotesize
\caption{Mean minutes per attempt (speedup over all-refresh) for every method; bold marks the fastest per column. \method is 4 to 5\% faster than all-refresh on SWE-bench Verified and Terminal-Bench 2.0 and 9 to 10\% faster on $\tau^2$.}\label{tab:elapsed}
\setlength{\tabcolsep}{4pt}
\begin{tabular}{lrrrrr}
\toprule
& \multicolumn{5}{c}{Minutes per attempt (speedup over all-refresh, $\times$)} \\
\cmidrule(lr){2-6}
Method & SWE & Terminal & Retail & Airline & Telecom \\
\midrule
\addlinespace[2pt]\multicolumn{6}{l}{\textcolor{black!65}{References}} \\
\quad Native attention & 22.9 (0.78) & 18.9 (0.77) & 12.0 (0.70) & 9.6 (0.70) & 12.9 (0.71) \\
\quad All-refresh sparse & \textcolor{black!60}{17.9 (1.00)} & \textcolor{black!60}{14.6 (1.00)} & \textcolor{black!60}{8.3 (1.00)} & \textcolor{black!60}{6.7 (1.00)} & \textcolor{black!60}{9.2 (1.00)} \\
\quad Strongest feasible static & 17.4 (1.03) & 13.9 (1.05) & 7.5 (1.11) & 6.0 (1.11) & 8.3 (1.10) \\
\addlinespace[2pt]\multicolumn{6}{l}{\textcolor{black!65}{Alternative injected selectors}} \\
\quad Training-aware IndexCache & 17.1 (1.05) & 13.9 (1.05) & 7.5 (1.12) & 6.0 (1.12) & 8.3 (1.11) \\
\quad LatentIndex & 17.4 (1.03) & 14.1 (1.03) & 7.9 (1.05) & 6.3 (1.06) & 8.7 (1.05) \\
\quad LatentIndex + HS & 17.2 (1.05) & 13.7 (1.06) & 7.5 (1.12) & 6.0 (1.12) & 8.3 (1.11) \\
\quad MISA, 1 head & 17.3 (1.04) & 14.0 (1.04) & \textbf{7.2} (1.16) & \textbf{5.7} (1.17) & \textbf{8.0} (1.15) \\
\quad MISA, 2 heads & 17.3 (1.04) & 13.9 (1.05) & 7.3 (1.14) & 5.8 (1.14) & 8.1 (1.13) \\
\quad MISA, 4 heads & 17.4 (1.03) & 14.0 (1.04) & 7.7 (1.09) & 6.1 (1.09) & 8.5 (1.08) \\
\quad MISA, 8 heads & 18.0 (1.00) & 14.6 (1.00) & 8.4 (1.00) & 6.7 (1.00) & 9.2 (1.00) \\
\quad HISA, all refresh & 16.9 (1.06) & 13.7 (1.07) & 7.5 (1.11) & 6.0 (1.11) & 8.3 (1.10) \\
\quad HISA, static schedule & \textbf{16.6} (1.08) & 13.7 (1.07) & 7.5 (1.11) & 6.0 (1.11) & 8.3 (1.10) \\
\quad MISA 4, static schedule & 17.1 (1.05) & 14.0 (1.04) & 7.7 (1.09) & 6.1 (1.09) & 8.5 (1.08) \\
\quad LongCat CLI & 16.9 (1.06) & 13.7 (1.07) & 7.4 (1.12) & 6.0 (1.12) & 8.3 (1.11) \\
\quad LongCat HI & 17.3 (1.04) & 13.8 (1.06) & 7.7 (1.09) & 6.1 (1.09) & 8.5 (1.08) \\
\quad LongCat HI, static schedule & 16.9 (1.07) & \textbf{13.6} (1.08) & 7.7 (1.09) & 6.1 (1.09) & \textbf{8.0} (1.15) \\
\quad ReTopK, indexer adapted & 18.3 (0.98) & 15.0 (0.97) & 8.3 (1.01) & 6.6 (1.01) & 9.1 (1.01) \\
\addlinespace[2pt]\multicolumn{6}{l}{\textcolor{black!65}{Native-score methods}} \\
\quad ReTopK, per head & 17.9 (1.00) & 14.2 (1.03) & 7.9 (1.05) & 6.4 (1.05) & 8.7 (1.05) \\
\quad HShare & 17.4 (1.03) & 13.8 (1.05) & 7.4 (1.12) & 6.0 (1.12) & 8.2 (1.12) \\
\quad Slow-Fast Inference & 17.1 (1.05) & 14.0 (1.04) & 7.6 (1.10) & 6.1 (1.10) & 8.4 (1.09) \\
\quad TidalDecode & 17.5 (1.02) & 14.0 (1.04) & 7.4 (1.13) & 5.9 (1.13) & 8.2 (1.12) \\
\quad Quest & 17.0 (1.05) & 13.9 (1.05) & 7.4 (1.12) & 6.0 (1.12) & 8.2 (1.12) \\
\quad Kascade & 17.1 (1.05) & 13.9 (1.05) & 7.3 (1.15) & 5.8 (1.15) & 8.1 (1.13) \\
\addlinespace[2pt]\multicolumn{6}{l}{\textcolor{black!65}{\method-based methods}} \\
\rowcolor{AIRRow}\quad HISA + \method & 17.0 (1.06) & 13.7 (1.07) & \textbf{7.2} (1.16) & 5.8 (1.16) & \textbf{8.0} (1.15) \\
\rowcolor{AIRRow}\quad MISA 4 + \method & 17.3 (1.04) & 14.1 (1.03) & 7.3 (1.15) & 5.8 (1.15) & 8.1 (1.13) \\
\rowcolor{AIRRow}\quad LongCat HI + \method & 17.1 (1.05) & \textbf{13.6} (1.07) & 7.3 (1.15) & 5.8 (1.14) & 8.1 (1.13) \\
\rowcolor{AIRRow}\quad \method & 17.1 (1.05) & 13.9 (1.05) & 7.5 (1.11) & 6.0 (1.11) & 8.3 (1.10) \\
\bottomrule
\end{tabular}
\par\smallskip{\scriptsize\textit{Key:} bold marks the best value in each column of the stated comparison set, ties at the reported precision included; \AIRswatch{AIRRow}~\method-based rows; gray: the all-refresh reference.}\par
\end{table}

\subsection{Backbones and caps}
\label{app:results-backbones}
\label{app:ext-results-backbones}

\begin{figure*}[tbp]
\centering
\includegraphics{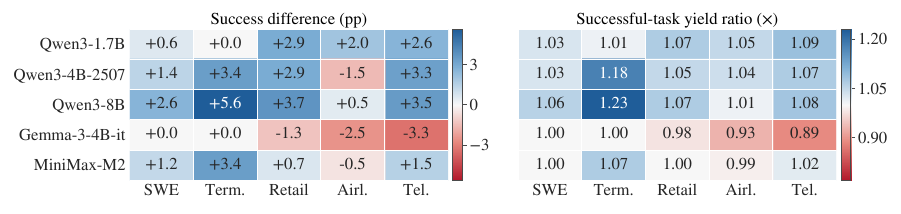}
\caption{At 32K, \method matches or exceeds the comparator in successful tasks per hour in 21 of 25 cells, by up to 23\%. Left: success difference (pp); right: ratio of successful tasks per hour.}
\label{fig:yield-atlas}
\end{figure*}

\method's success exceeds the comparator's in 37 of 50 backbone, cap, and benchmark settings, ties in 7, and falls below it in 6 (Figure~\ref{fig:gains-main}). Validation selects a learned controller in nine of ten backbone and cap configurations; for Gemma-3-4B-it at 32K no learned candidate meets the cost bound and the static fallback $R=2$ is deployed (Table~\ref{tab:t09_deployment_selection}). \method's time per attempt ranges from 3.9\% longer than the comparator's (Qwen3-8B at 128K on SWE-bench Verified, a tie in success) to 7.0\% shorter (Qwen3-4B-Instruct-2507 at 32K on airline). In successful tasks per hour at 32K, \method matches or exceeds the comparator in 21 of 25 settings, by up to 23\% on Qwen3-8B on Terminal-Bench 2.0; its lower-yield settings are Gemma-3-4B-it on the three $\tau^2$ domains, where it is the static fallback, and MiniMax-M2 on airline (Figure~\ref{fig:yield-atlas}).

\begin{figure*}[tbp]
\centering
\includegraphics{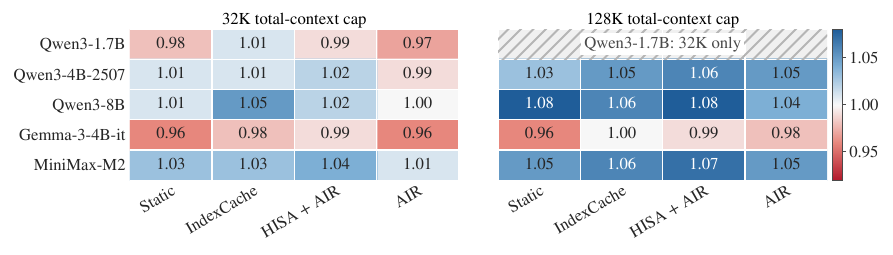}
\caption{\method's whole-agent speedup on SWE-bench Verified is 0.96 to 1.05, smaller than its prefill and decode gains because tools and the environment take about half of each episode.}
\label{fig:agent-speedup}
\end{figure*}

\begin{figure*}[tbp]
\centering
\includegraphics{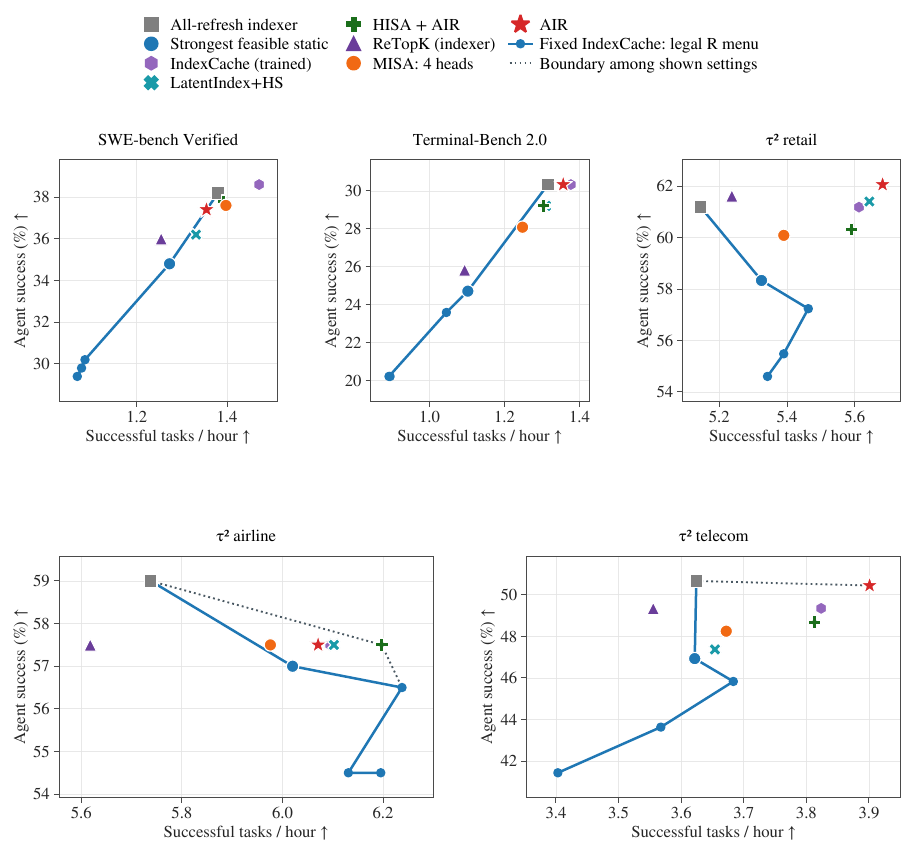}
\caption{On Qwen3-8B at 32K, \method lies on the success-yield boundary (dotted: settings no other displayed setting beats on both axes) in two of the five benchmarks. Blue: static menu.}
\label{fig:detail-qwen8}
\end{figure*}

Table~\ref{tab:family-summary} compares \method with all-refresh on each backbone. Model-side speedups are larger than whole-agent ones: on SWE-bench Verified, \method runs the reference prefill (measured on an isolated reference workload) 1.14 to 1.30 times and model-only decoding 1.05 to 1.17 times as fast as all-refresh on every backbone and cap, while its whole-agent speedup there ranges from 0.96 (Gemma-3-4B-it at 32K) to 1.05 (Figure~\ref{fig:agent-speedup}). On the success-yield plane at 32K, \method lies on the boundary of settings that no other displayed setting beats on both axes in none of the five benchmarks for Qwen3-1.7B and Gemma-3-4B-it, one for Qwen3-4B-Instruct-2507 and MiniMax-M2, and two for Qwen3-8B (Figure~\ref{fig:detail-qwen8} shows Qwen3-8B).

\begin{table}[tbp]
\centering
\small
\setlength{\tabcolsep}{5pt}
\begin{tabular}{llrr}
\toprule
Backbone & Caps & Speedup & Success change (pp) \\
\midrule
Qwen3-4B-Instruct-2507 & 32K, 128K & 0.99--1.11 & $-2.2$ to $+1.1$ \\
Qwen3-8B & 32K, 128K$^\ddagger$ & 1.00--1.13 & $-1.5$ to $+0.9$ \\
Gemma-3-4B-it & 32K$^\dagger$, 128K & 0.96--1.03 & $-5.0$ to $-1.3$ \\
MiniMax-M2 & 32K, 128K & 1.01--1.12 & $-2.6$ to $+1.1$ \\
Qwen3-4B (original) & 32K & 0.99--1.08 & $-2.0$ to $+0.2$ \\
\bottomrule
\end{tabular}
\caption{\method relative to all-refresh on each backbone: ranges of whole-agent speedup and success change over all evaluated caps and benchmarks. $^\dagger$static fallback $R=2$; $^\ddagger$YaRN extension.}
\label{tab:family-summary}
\end{table}

\subsection{Ablations, decision interval, and the paid probe}
\label{app:results-ablations}
\label{app:ext-results-ablations}

\begin{figure*}[tbp]
\centering
\includegraphics{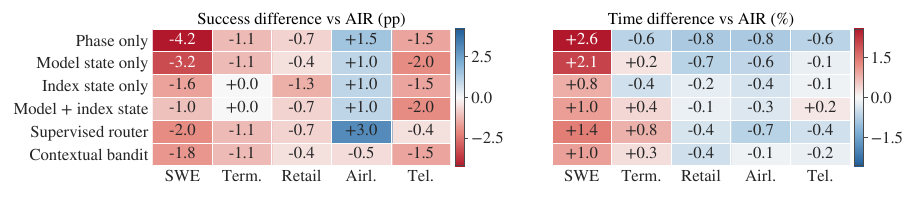}
\caption{Every state ablation and both locally trained alternatives lower SWE-bench Verified success relative to \method, by 1.0 to 4.2 points; on airline five of the six score higher. Left: success difference (pp); right: time change (\%), positive is slower.}
\label{fig:ablation}
\end{figure*}

\begin{figure*}[tbp]
\centering
\includegraphics{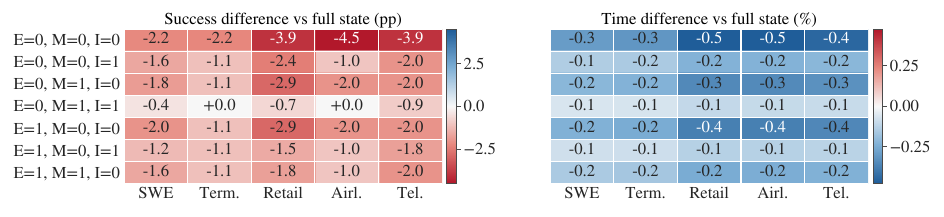}
\caption{The full controller state gives the highest success on all five populations (tied on Terminal-Bench 2.0 and airline with model plus index state); removing all three groups costs 2.2 to 4.5 points. Rows: the other seven combinations of event ($E$), model ($M$), and index ($I$) state.}
\label{fig:feature-factorial}
\end{figure*}

Each ingredient of \method is removed in turn on the primary model (Table~\ref{tab:ablation}, Figure~\ref{fig:ablation}). Periodic masks come within 1.2 points of the greedy mask with the same $R$ on SWE-bench Verified, and seeded staggered masks lose 1.8 to 3.0 points. Across all eight combinations of event, model, and index state, the full state is best on every benchmark, tied on Terminal-Bench 2.0 and airline with model plus index state, and removing all three groups costs 2.2 to 4.5 points (Figure~\ref{fig:feature-factorial}).

\begin{figure*}[tbp]
\centering
\includegraphics{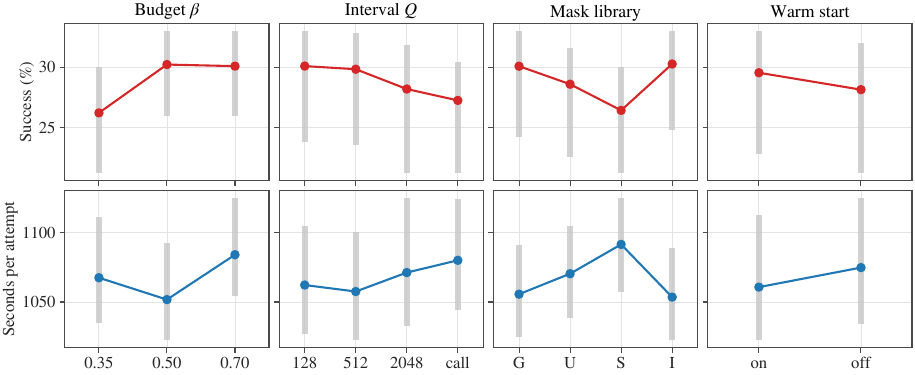}
\caption{Budget, decision interval, and library each move SWE-bench Verified success by up to 4 points, and the warm start adds 1.4 points on average. Dots: mean over the 96 controllers at each level; bars: range; $Q$ in tokens (call: call-only); libraries G greedy, U uniform, S staggered, I independent.}
\label{fig:factorial-main}
\end{figure*}

The factorial study fits 96 controllers on SWE-bench Verified, crossing three budgets, four decision intervals, four mask libraries, and warm start on or off (Figure~\ref{fig:factorial-main}). The budget has the largest effect: configurations at $\beta=0.50$ average 30.2\% success against 26.2\% at $\beta=0.35$, and $\beta=0.70$ adds no success (30.1\%) while taking 32 seconds longer per attempt. Mean success falls from 30.1\% at $Q=128$ to 27.2\% with call-only decisions, and the greedy and independently searched libraries beat uniform and staggered ones. The supervised warm start raises success in all 48 matched pairs, by 0.6 to 2.6 points, and shortens time per attempt by 1.0 to 1.9\%; across matched cells, interactions between factors lie between $-1.5$ and $+1.1$ points of success and between $-7.4$ and $+12.9$ seconds per attempt.

\begin{figure*}[tbp]
\centering
\includegraphics{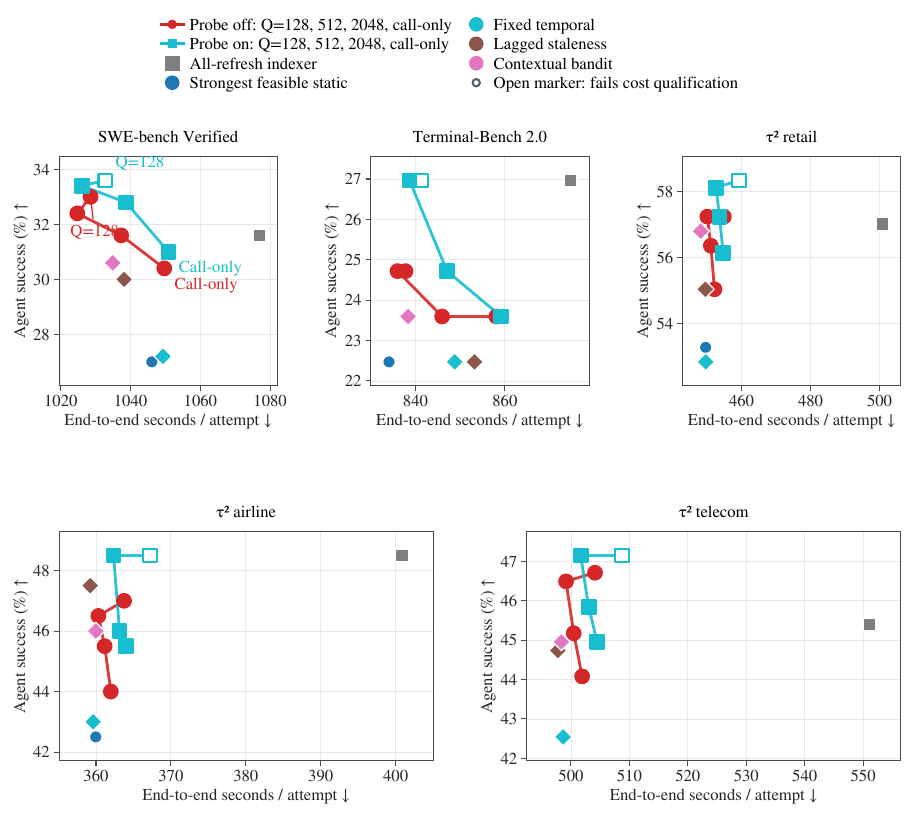}
\caption{The paid probe raises or matches success at every decision interval, by up to 2.3 points, and adds 1 to 5 seconds per attempt. Curves join $Q=128$, 512, 2048, and call-only decisions with the probe off (red) or on (cyan); open markers fail the cost check.}
\label{fig:scheduler-controls}
\end{figure*}

\begin{figure*}[tbp]
\centering
\includegraphics{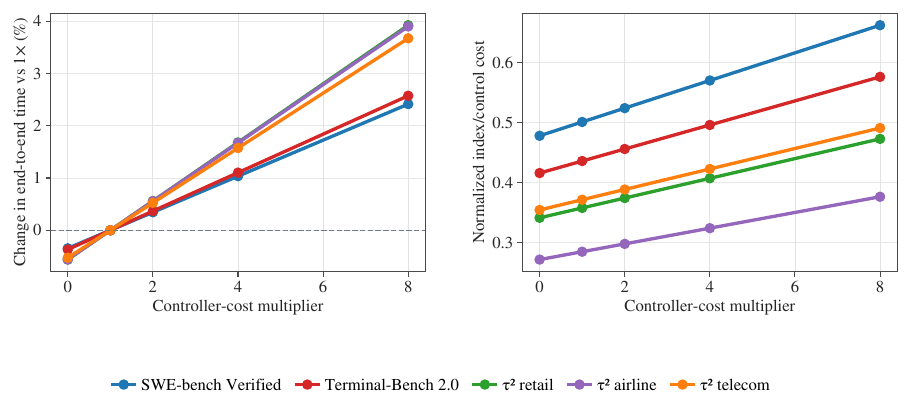}
\caption{Controller cost barely moves end-to-end time: removing it shortens episodes by 0.3 to 0.6\%, and an eight-fold cost lengthens them by 2.4 to 3.9\%. Left: time change; right: normalized index and control cost.}
\label{fig:overhead}
\end{figure*}

The paid probe raises or matches success at every decision interval, by up to 2.3 points, and adds 1 to 5 seconds per attempt; with the probe off, call-only decisions give the lowest or tied-lowest success on every benchmark (Figure~\ref{fig:scheduler-controls}). The probe adds about 0.02 to index and control cost, and 0.03 at $Q=128$, where the probe-on candidates exceed the cost tolerance; its cost includes embeddings, native-prefix processing, and indexer work, and is largest on Gemma-3-4B-it (Table~\ref{tab:probe}). Stochastic deployment scores higher than argmax deployment in 36 of 40 matched settings and ties in the rest, by 1.0 point on average; the deployed controller samples with the probe off at $Q=512$. Removing the controller's cost shortens episodes by 0.3 to 0.6\%, and making it eight times larger lengthens them by 2.4 to 3.9\% (Figure~\ref{fig:overhead}).

\begin{table}[ht]\centering\footnotesize
\caption{Among the $\beta=0.50$ variants, \method has the highest success on SWE-bench Verified, retail, and telecom and ties the best Terminal-Bench 2.0 score; on airline five locally trained or restricted-state variants score higher, the supervised router most. The larger budget $\beta=0.70$ is a separate experiment and scores higher on four populations. Success (\%); last column: SWE-bench Verified speedup over all-refresh.}\label{tab:ablation}
\setlength{\tabcolsep}{4pt}
\begin{tabular}{lrrrrrr}
\toprule
Method & SWE & Terminal & Retail & Airline & Telecom & Speedup \\
\midrule
All-refresh reference & \textcolor{black!60}{31.6} & \textcolor{black!60}{27.0} & \textcolor{black!60}{57.0} & \textcolor{black!60}{48.5} & \textcolor{black!60}{45.4} & \textcolor{black!60}{1.00} \\
\addlinespace[2pt]\multicolumn{7}{l}{\textcolor{black!65}{Placement and state-free variation}} \\
\quad Strongest feasible static ($R=18$) & 27.0 & 22.5 & 53.3 & 42.5 & 42.5 & 1.03 \\
\quad Staggered mask, $R=18$ & 24.0 & 21.3 & 50.4 & 40.5 & 40.1 & 1.00 \\
\quad State-independent mixture & 27.0 & 22.5 & 52.9 & 42.5 & 42.5 & 1.03 \\
\addlinespace[2pt]\multicolumn{7}{l}{\textcolor{black!65}{Rules and local learning}} \\
\quad Event-triggered rule & 29.4 & 23.6 & 53.3 & 48.0 & 41.7 & 1.04 \\
\quad Lagged-staleness rule & 30.0 & 22.5 & 55.0 & 47.5 & 44.7 & 1.04 \\
\quad Supervised router & 30.4 & 23.6 & 56.6 & \textbf{49.5} & 46.1 & 1.04 \\
\quad Contextual bandit & 30.6 & 23.6 & 56.8 & 46.0 & 45.0 & 1.04 \\
\addlinespace[2pt]\multicolumn{7}{l}{\textcolor{black!65}{Restricted controller state}} \\
\quad Event metadata only & 28.2 & 23.6 & 56.6 & 48.0 & 45.0 & 1.02 \\
\quad Model and index statistics & 31.4 & \textbf{24.7} & 56.6 & 47.5 & 44.5 & 1.04 \\
\addlinespace[2pt]\multicolumn{7}{l}{\textcolor{black!65}{\method budgets ($\beta=0.35$ and $0.70$ unranked)}} \\
\rowcolor{AIRRow}\quad $\beta=0.35$ & 29.8 & 23.6 & 56.8 & 45.5 & 44.1 & 1.04 \\
\rowcolor{AIRRow}\quad $\beta=0.50$, deployed & \textbf{32.4} & \textbf{24.7} & \textbf{57.2} & 46.5 & \textbf{46.5} & 1.05 \\
\rowcolor{AIRRow}\quad $\beta=0.70$ & 32.4 & 28.1 & 59.0 & 49.0 & 47.4 & 1.02 \\
\bottomrule
\end{tabular}
\par\smallskip{\scriptsize\textit{Key:} bold marks the best value in each column of the stated comparison set, ties at the reported precision included; \AIRswatch{AIRRow}~\method-based rows; gray: reference, not ranked; other budgets are not ranked.}\par
\end{table}

\begin{table}[ht]\centering\small
\caption{At $Q=512$ the paid probe raises or matches success everywhere; its cost is small except on Gemma-3-4B-it, whose probe runs five native layers. Entries: ranges over populations of probe-on minus probe-off.}\label{tab:probe}
\setlength{\tabcolsep}{4pt}
\begin{tabular}{llrrr}
\toprule
Backbone & Cap & Success change (pp) & Added cost & Time change (\%) \\
\midrule
Qwen3-1.7B & 32K & 0.0 to +2.0 & 0.019 to 0.031 & 0.0 to +0.4 \\
Qwen3-4B-Instruct-2507 & 32K & +0.4 to +2.0 & 0.013 to 0.024 & 0.0 to +0.4 \\
Qwen3-4B-Instruct-2507 & 128K & +0.7 to +2.2 & 0.014 to 0.024 & +0.1 to +0.6 \\
Qwen3-8B & 32K & 0.0 to +1.5 & 0.014 to 0.024 & 0.0 to +0.4 \\
Qwen3-8B & 128K & +1.0 to +1.6 & 0.014 to 0.024 & +0.3 to +0.6 \\
Gemma-3-4B-it & 32K & +2.0 to +3.9 & 0.217 to 0.352 & -1.5 to +3.6 \\
Gemma-3-4B-it & 128K & 0.0 to +2.0 & 0.139 to 0.225 & +1.3 to +3.0 \\
MiniMax-M2 & 32K & +0.7 to +2.2 & 0.009 to 0.015 & 0.0 to +0.4 \\
MiniMax-M2 & 128K & 0.0 to +1.8 & 0.008 to 0.015 & +0.1 to +0.5 \\
\bottomrule
\end{tabular}

\end{table}

\subsection{Where the time goes}
\label{app:results-efficiency}
\label{app:ext-results-efficiency}

\begin{figure*}[tbp]
\centering
\includegraphics{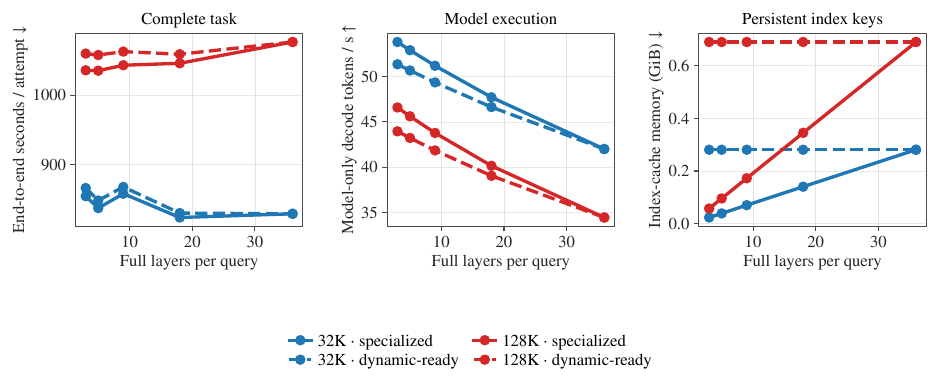}
\caption{Keeping a static mask dynamic-ready leaves success unchanged and costs up to 2.3\% more time, 5.6\% lower decode throughput, and 12 times the index-cache memory at small $R$ (SWE-bench Verified; blue 32K, red 128K).}
\label{fig:static-runtime}
\end{figure*}

\begin{figure*}[tbp]
\centering
\includegraphics{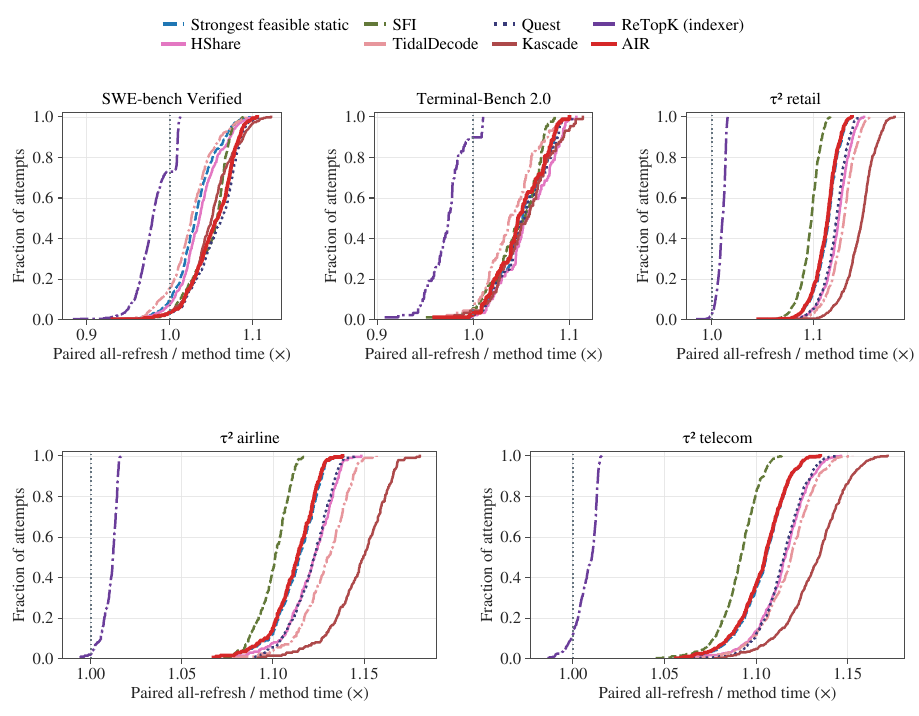}
\caption{\method shortens most attempts relative to all-refresh: median paired speedups are 1.06 on SWE-bench Verified, 1.05 on Terminal-Bench 2.0, and 1.10 to 1.11 on $\tau^2$, and 96 to 100\% of attempts finish faster. Curves: distributions of per-attempt time ratios.}
\label{fig:reuse-methods}
\end{figure*}

\begin{figure*}[tbp]
\centering
\includegraphics{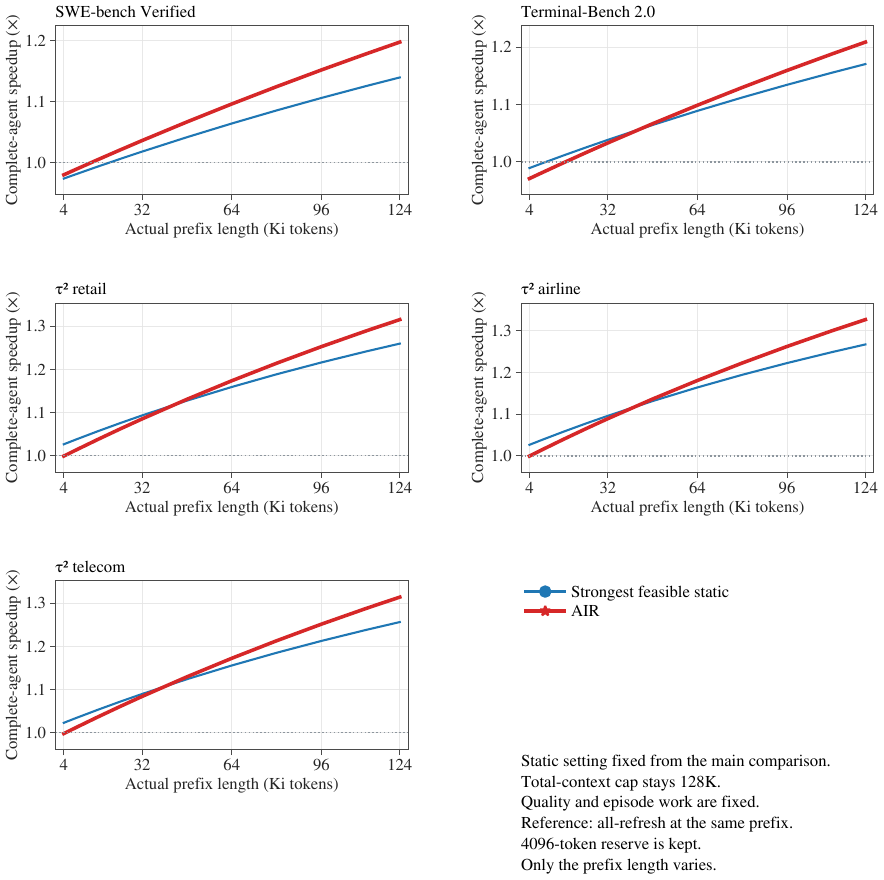}
\caption{The prefix effect holds on all five populations: \method's speedup over all-refresh rises from 0.97--1.00 at 4,096 tokens to 1.20--1.33 at 126,976 tokens, most on $\tau^2$.}
\label{fig:prefix-benchmarks}
\end{figure*}

Keeping a static mask dynamic-ready, with index keys at every eligible layer, leaves success unchanged and costs up to 2.3\% more time, up to 5.6\% lower decode throughput, and up to 12 times the index-cache memory at small refresh counts; at $R=36$ the specialized and dynamic-ready runtimes coincide (Figure~\ref{fig:static-runtime}). \method finishes faster than all-refresh on most attempts, with median paired speedups of 1.06 on SWE-bench Verified, 1.05 on Terminal-Bench 2.0, and 1.10 to 1.11 on $\tau^2$; the indexer adaptation of ReTopK is the one method slower than all-refresh on most SWE-bench Verified and Terminal-Bench 2.0 attempts (Figure~\ref{fig:reuse-methods}). In a prefix study that holds each workload fixed over eleven prefix lengths, with the 4,096-token generation reserve kept, \method's speedup over all-refresh rises from 0.97 to 1.00 at 4,096 tokens to 1.20 to 1.33 at 126,976 tokens, most on $\tau^2$, where model work dominates the episode; on SWE-bench Verified it rises from 0.98 to 1.20 (Figure~\ref{fig:prefix-benchmarks}). Selection work grows with the prefix while key maintenance and controller work stay constant per query.

\subsection{Other selectors, composition, and retrained backbones}
\label{app:results-selectors}
\label{sec:results-selectors}

\begin{figure*}[tbp]
\centering
\includegraphics{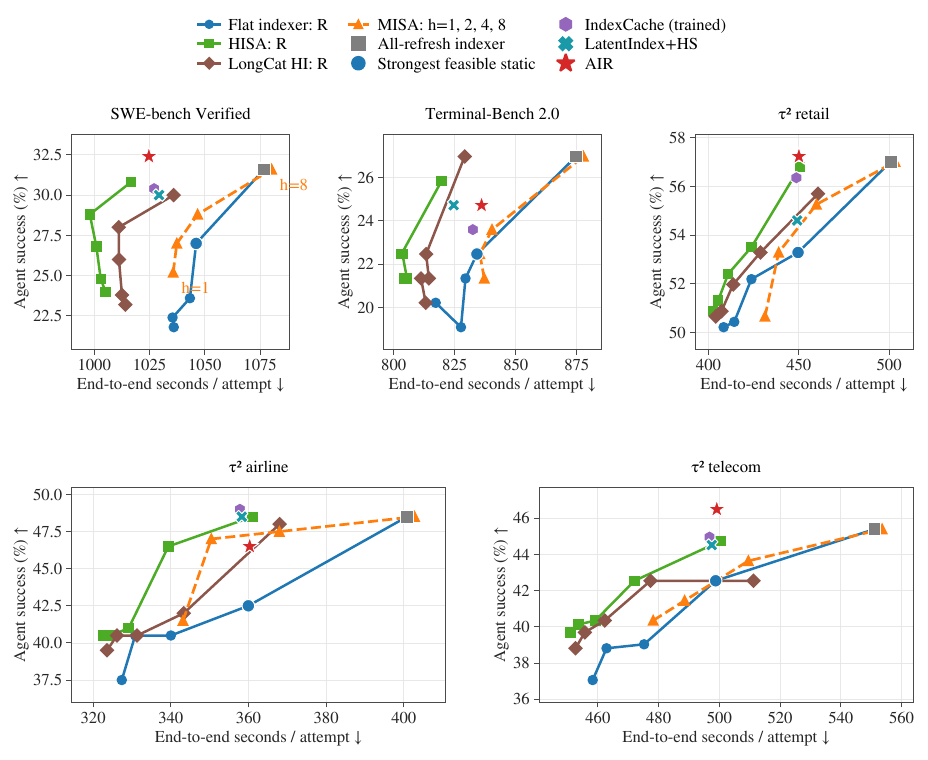}
\caption{\method has the highest success of 24 selector settings on SWE-bench Verified, retail, and telecom; on Terminal-Bench 2.0 and airline other settings score up to 2.2 and 2.5 points higher. Solid: $R$ sweeps per selector; dashed: MISA head sweep.}
\label{fig:related-indexers}
\end{figure*}

\begin{figure*}[tbp]
\centering
\includegraphics{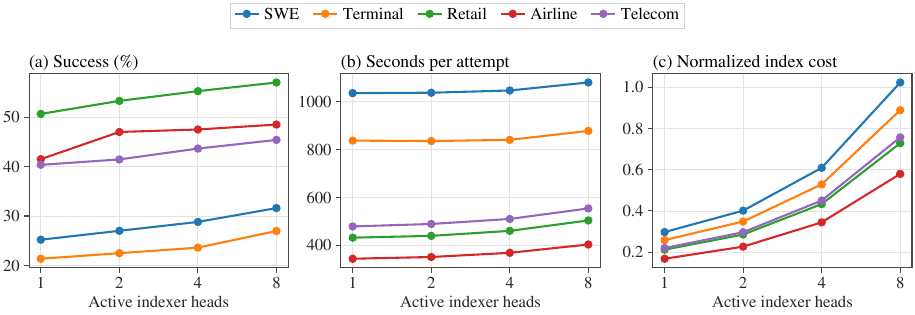}
\caption{Routing fewer of the eight indexer heads (MISA) cuts index cost by about 70\% at one head and costs 5.0 to 7.0 points of success.}
\label{fig:misa-heads}
\end{figure*}

\begin{figure*}[tbp]
\centering
\includegraphics{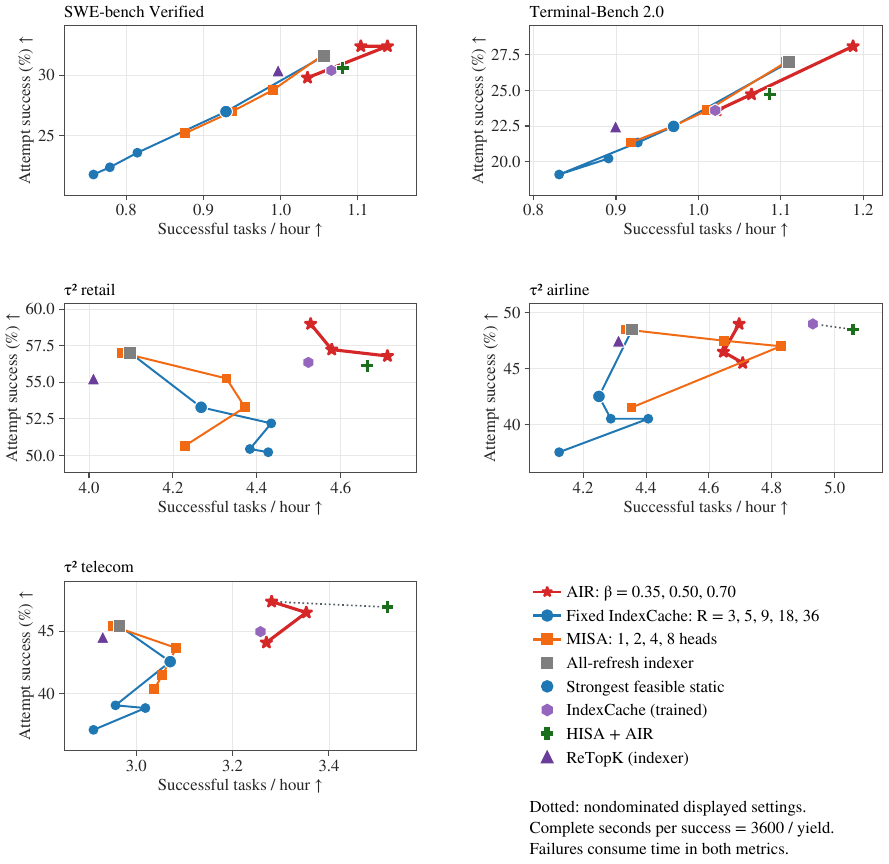}
\caption{\method at $\beta=0.50$ gives the most successful tasks per hour of 17 settings on SWE-bench Verified and ranks second on telecom; other settings yield more on Terminal-Bench 2.0 and airline. Dotted: settings no other setting beats on both axes.}
\label{fig:sweeps}
\end{figure*}

No alternative selector dominates \method on every benchmark (Table~\ref{tab:alternatives}). On SWE-bench Verified, \method has the highest success of 24 selector settings; on Terminal-Bench 2.0 and airline, settings with more Full layers or a different selector score up to 2.2 and 2.5 points higher (Figure~\ref{fig:related-indexers}). Routing fewer of the eight indexer heads, in the style of MISA, cuts index cost by about 70\% at one head and shortens $\tau^2$ episodes by 14 to 15\%, at a cost of 5.0 to 7.0 points of success (Figure~\ref{fig:misa-heads}). In successful tasks per hour, \method at $\beta=0.50$ leads 17 settings on SWE-bench Verified and ranks second on telecom (Figure~\ref{fig:sweeps}).

\begin{figure*}[tbp]
\centering
\includegraphics{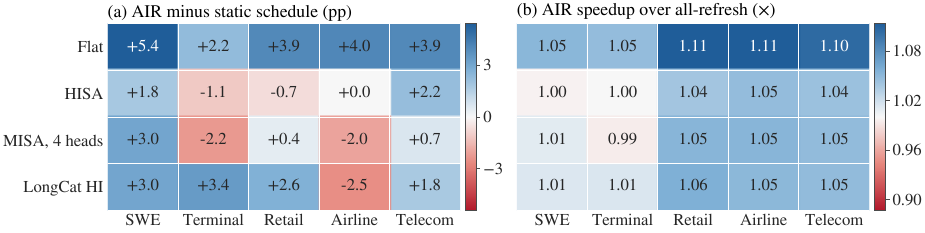}
\caption{Learned allocation helps most with the flat indexer; with cheaper selectors its success gain depends on the population. (a) \method minus the selector's static schedule (pp); (b) speedup over the selector's all-refresh run.}
\label{fig:composition}
\end{figure*}

With the flat indexer, \method scheduling beats the selector's static schedule on every benchmark by 2.2 to 5.4 points; with HISA, four-head MISA, and LongCat hierarchical indexing the change ranges from $-2.5$ to $+3.4$ points (Figure~\ref{fig:composition}). Speedups over each selector's own all-refresh run are 1.05 to 1.11 with the flat selector and up to 1.06 with the others; in 10 of the 15 settings for the three cheaper selectors, the selected static schedule is the selector's own all-refresh ($R=36$). Full LongCat Sparse Attention retrains the backbone in a separate track: it costs 180 GPU-hours of adaptation (192 with \method), changes success by $-1.2$ to $+1.3$ points against the unadapted all-refresh model, and needs 6,403 to 12,858 attempts to repay its preparation through faster serving, counting model-serving GPU time only; adding \method raises its success on four of five benchmarks (Table~\ref{tab:longcat}).

\begin{table}[tbp]\centering\footnotesize
\caption{Among the frozen-backbone methods shown, \method has the highest success on SWE-bench Verified, HISA with \method on telecom, and LongCat CLI on retail, and HISA with \method is the fastest or tied-fastest on every population; Table~\ref{tab:elapsed} lists configurations that are faster on some populations. Gray: native attention, unranked.}\label{tab:alternatives}
\setlength{\tabcolsep}{3pt}
\begin{tabular}{lrrrrrrrrrr}
\toprule
& \multicolumn{5}{c}{Success (\%) $\uparrow$} & \multicolumn{5}{c}{Speedup over all-refresh $\uparrow$} \\
\cmidrule(lr){2-6}\cmidrule(lr){7-11}
Method & SWE & Term. & Retail & Airl. & Tel. & SWE & Term. & Retail & Airl. & Tel. \\
\midrule
\addlinespace[2pt]\multicolumn{11}{l}{\textcolor{black!65}{References}} \\
\quad Native attention & \textcolor{black!60}{32.0} & \textcolor{black!60}{28.1} & \textcolor{black!60}{57.7} & \textcolor{black!60}{50.0} & \textcolor{black!60}{46.5} & \textcolor{black!60}{0.78} & \textcolor{black!60}{0.77} & \textcolor{black!60}{0.70} & \textcolor{black!60}{0.70} & \textcolor{black!60}{0.71} \\
\quad All-refresh sparse & 31.6 & \textbf{27.0} & 57.0 & 48.5 & 45.4 & 1.00 & 1.00 & 1.00 & 1.00 & 1.00 \\
\quad Strongest feasible static & 27.0 & 22.5 & 53.3 & 42.5 & 42.5 & 1.03 & 1.05 & 1.11 & 1.11 & 1.10 \\
\addlinespace[2pt]\multicolumn{11}{l}{\textcolor{black!65}{Alternative injected selectors}} \\
\quad Training-aware IndexCache & 30.4 & 23.6 & 56.4 & 49.0 & 45.0 & 1.05 & 1.05 & 1.12 & 1.12 & 1.11 \\
\quad LatentIndex & 31.6 & \textbf{27.0} & 56.4 & \textbf{49.5} & 45.2 & 1.03 & 1.03 & 1.05 & 1.06 & 1.05 \\
\quad LatentIndex + HS & 30.0 & 24.7 & 54.6 & 48.5 & 44.5 & 1.05 & 1.06 & 1.12 & 1.12 & 1.11 \\
\quad MISA, 4 heads & 28.8 & 23.6 & 55.3 & 47.5 & 43.6 & 1.03 & 1.04 & 1.09 & 1.09 & 1.08 \\
\quad HISA, all refresh & 30.8 & 25.8 & 56.8 & 48.5 & 44.7 & \textbf{1.06} & \textbf{1.07} & 1.11 & 1.11 & 1.10 \\
\quad LongCat CLI & 31.4 & \textbf{27.0} & \textbf{58.6} & 49.0 & 45.0 & \textbf{1.06} & \textbf{1.07} & 1.12 & 1.12 & 1.11 \\
\quad LongCat HI & 30.0 & \textbf{27.0} & 55.7 & 48.0 & 42.5 & 1.04 & 1.06 & 1.09 & 1.09 & 1.08 \\
\quad ReTopK, indexer adapted & 30.4 & 22.5 & 55.3 & 47.5 & 44.5 & 0.98 & 0.97 & 1.01 & 1.01 & 1.01 \\
\addlinespace[2pt]\multicolumn{11}{l}{\textcolor{black!65}{Native-score methods}} \\
\quad ReTopK, per head & 29.8 & \textbf{27.0} & 56.6 & 47.5 & 46.1 & 1.00 & 1.03 & 1.05 & 1.05 & 1.05 \\
\quad HShare & 28.0 & 23.6 & 55.7 & 46.0 & 44.5 & 1.03 & 1.05 & 1.12 & 1.12 & 1.12 \\
\quad Slow-Fast Inference & 30.8 & 23.6 & 56.4 & 49.0 & 45.0 & 1.05 & 1.04 & 1.10 & 1.10 & 1.09 \\
\quad TidalDecode & 26.8 & 22.5 & 54.4 & 47.0 & 43.9 & 1.02 & 1.04 & 1.13 & 1.13 & 1.12 \\
\quad Quest & 30.4 & 23.6 & 56.6 & 48.5 & 45.6 & 1.05 & 1.05 & 1.12 & 1.12 & 1.12 \\
\quad Kascade & 28.0 & 22.5 & 53.3 & 46.0 & 41.9 & 1.05 & 1.05 & 1.15 & 1.15 & 1.13 \\
\addlinespace[2pt]\multicolumn{11}{l}{\textcolor{black!65}{\method-based methods}} \\
\rowcolor{AIRRow}\quad HISA + \method & 30.6 & 24.7 & 56.1 & 48.5 & \textbf{46.9} & \textbf{1.06} & \textbf{1.07} & \textbf{1.16} & \textbf{1.16} & \textbf{1.15} \\
\rowcolor{AIRRow}\quad \method & \textbf{32.4} & 24.7 & 57.2 & 46.5 & 46.5 & 1.05 & 1.05 & 1.11 & 1.11 & 1.10 \\
\bottomrule
\end{tabular}
\par\smallskip{\scriptsize\textit{Key:} bold marks the best value in each column of the stated comparison set, ties at the reported precision included; \AIRswatch{AIRRow}~\method-based rows; gray: unranked reference.}\par
\end{table}

\begin{table}[tbp]\centering\small
\caption{Full LongCat Sparse Attention keeps success within 0.8 points of the unadapted all-refresh model at 1.08 to 1.17 times its speed, and adding \method raises success on four of five populations; repaying 180 GPU-hours (192 with \method) takes 6,403 to 12,858 attempts.}\label{tab:longcat}
\setlength{\tabcolsep}{3.5pt}
\begin{tabular}{lrrrrrrr}
\toprule
& \multicolumn{3}{c}{Success (\%)} & \multicolumn{2}{c}{Speedup} & \multicolumn{2}{c}{Break-even attempts} \\
\cmidrule(lr){2-4}\cmidrule(lr){5-6}\cmidrule(lr){7-8}
Population & Unadapted & LongCat & +\,\method & LongCat & +\,\method & LongCat & +\,\method \\
\midrule
SWE & 31.6 & 30.8 & \textbf{31.8} & 1.09 & 1.07 & 6,403 & 7,290 \\
Terminal & \textbf{27.0} & \textbf{27.0} & 25.8 & 1.10 & 1.08 & 7,369 & 8,390 \\
Retail & 57.0 & 56.6 & \textbf{57.9} & 1.17 & 1.15 & 8,987 & 10,231 \\
Airline & 48.5 & 48.5 & \textbf{49.5} & 1.17 & 1.15 & 11,294 & 12,858 \\
Telecom & 45.4 & 45.0 & \textbf{46.7} & 1.15 & 1.14 & 8,652 & 9,851 \\
\bottomrule
\end{tabular}
\par\smallskip{\scriptsize\textit{Key:} bold marks the best value in each column of the stated comparison set, ties at the reported precision included; \AIRswatch{AIRRow}~\method-based rows.}\par
\end{table}

\subsection{Transfer, in-domain training, reliability, and long histories}
\label{app:results-transfer}
\label{sec:results-transfer}

The repair controller transfers to the other benchmarks with the gains of Table~\ref{tab:primary-contrast}. On held-out subsets of the target benchmarks, controllers trained in-domain raise success over the transferred controller by 5.0 points on Terminal-Bench 2.0 and airline (20 tasks each) and by 1.3 points on retail and telecom (40 each), with time per attempt within 0.5\% (Table~\ref{tab:t08_in_domain}); the two benchmarks with the largest gain are those on which all-refresh beats the transferred controller. These subset scores come from controllers trained and validated on other tasks of the same benchmark and are reported apart from the full-population transfer results.

\begin{figure*}[tbp]
\centering
\includegraphics{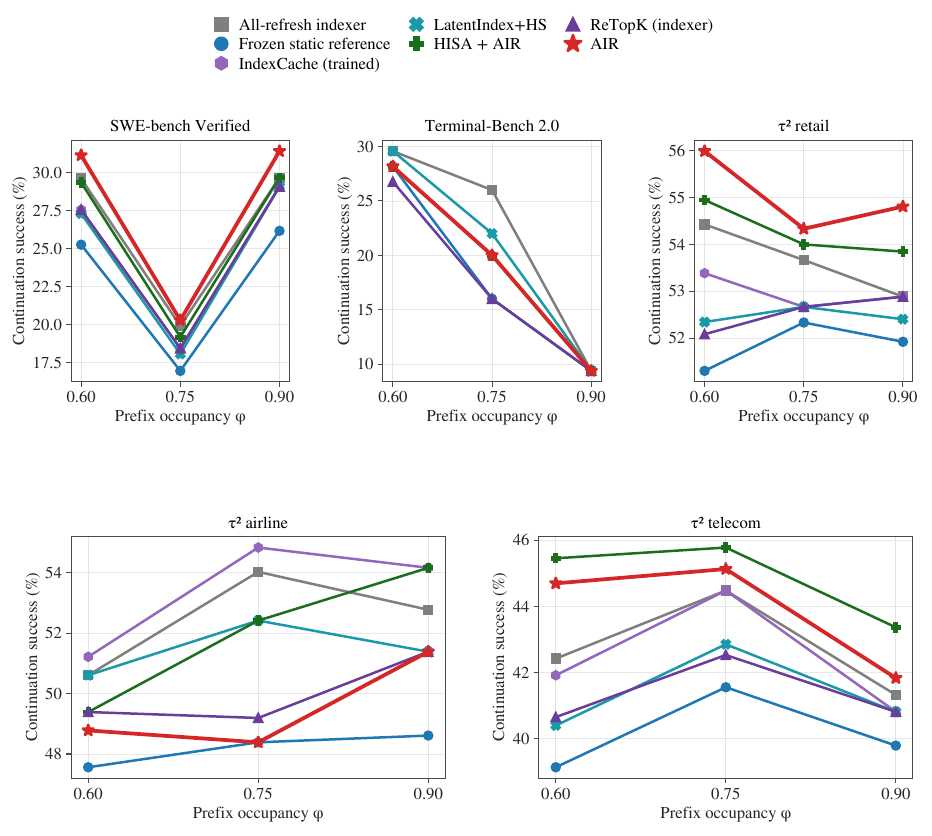}
\caption{In long-history continuations, \method's success exceeds the comparator's in 12 of 15 population and occupancy bands and ties it in 3; all-refresh scores higher in 5 bands. Success covers only the continuation from a shared snapshot.}
\label{fig:continuation-success}
\end{figure*}

\begin{figure*}[tbp]
\centering
\includegraphics{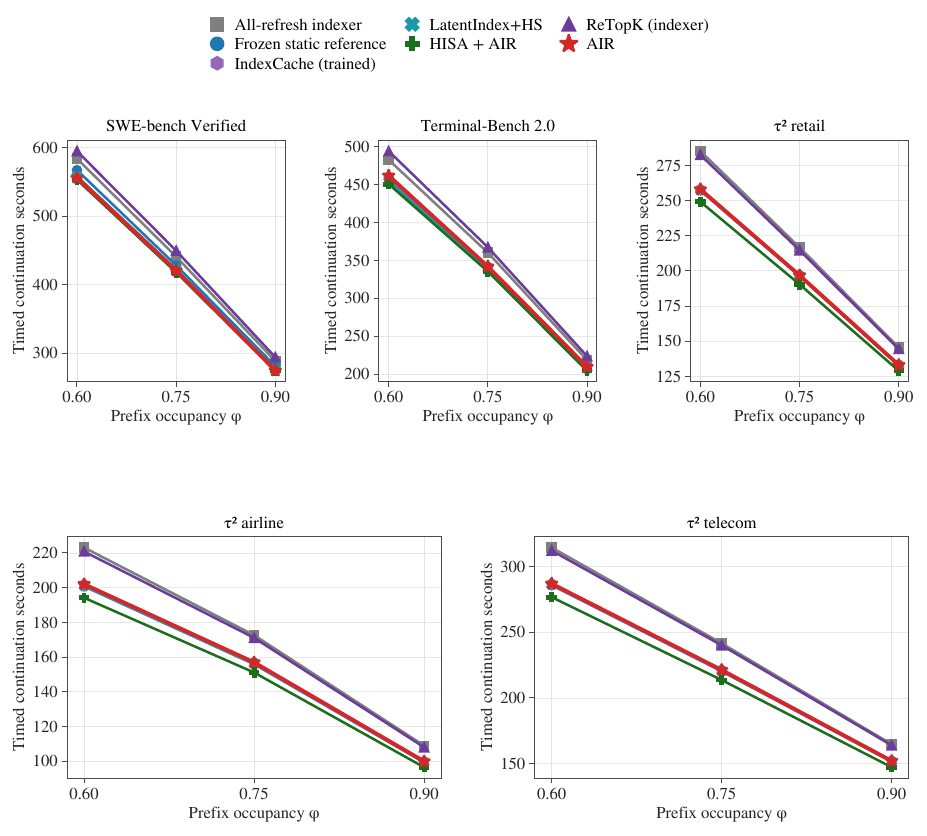}
\caption{In long-history continuations, \method finishes the remaining work 4 to 10\% faster than all-refresh in every band and within 2.2\% of the comparator. Snapshots, bands, and methods as in Figure~\ref{fig:continuation-success}.}
\label{fig:continuation-time}
\end{figure*}

\method keeps four-trial reliability close to native attention and all-refresh: at 128K its pass$^4$ is 21.1\% on retail against 15.8\% for the comparator, and the comparator is 2.0 and 2.6 points lower on airline and telecom; on 32K airline the comparator is higher (16.0 against 12.0\%) (Table~\ref{tab:reliability}). In long-history continuations, every method resumes from the same all-refresh snapshot once the prefix fills a fraction $\varphi\in\{0.60,0.75,0.90\}$ of usable context, with the comparator frozen from the full-episode comparison; a task is eligible at its first boundary whose prefix lies in $[\lceil\varphi L_{\mathrm{use}}\rceil,L_{\mathrm{use}}]$, where $L_{\mathrm{use}}$ is the cap minus the 4,096-token generation reserve (Table~\ref{tab:continuation-pop}). \method's success exceeds the comparator's in 12 of 15 benchmark and occupancy bands and ties it in 3, all-refresh scores higher in 5 bands, including all three airline bands (Figure~\ref{fig:continuation-success}), and \method finishes the remaining work 4 to 10\% faster than all-refresh in every band (Figure~\ref{fig:continuation-time}). On SWE-bench Verified, \method solves 31.1, 20.3, and 31.4\% of the remaining work at $\varphi=0.60$, 0.75, and 0.90, against 25.3, 17.0, and 26.2\% for the static comparator and 29.6, 19.9, and 29.7\% for all-refresh (Figure~\ref{fig:evidence}d).

\begin{table*}[tbp]
\centering
\small
\caption{In-domain training raises held-out success over the transferred repair controller by 1.3 to 5.0 points at equal time per attempt (minutes). Both controllers run on the same held-out tasks; differences are calculated before rounding.}
\label{tab:t08_in_domain}
\setlength{\tabcolsep}{5pt}
\begin{tabular}{lrrrrrr}
\toprule
 & & \multicolumn{3}{c}{Held-out success (\%)} & \multicolumn{2}{c}{Minutes per attempt} \\
\cmidrule(lr){3-5}\cmidrule(lr){6-7}
Subset & Tasks & Transfer & In-domain & $\Delta$ (pp) & Transfer & In-domain \\
\midrule
Terminal-Bench 2.0 & 20 & 15.0 & 20.0 & $+5.0$ & 15.5 & 15.5 \\
$\tau^2$ retail & 40 & 56.9 & 58.1 & $+1.3$ & 7.5 & 7.5 \\
$\tau^2$ airline & 20 & 35.0 & 40.0 & $+5.0$ & 6.5 & 6.5 \\
$\tau^2$ telecom & 40 & 44.4 & 45.6 & $+1.3$ & 9.0 & 9.0 \\
\bottomrule
\end{tabular}
\end{table*}

\begin{table}[tbp]\centering\small
\caption{\method's pass$^4$ (all-four success, \%) exceeds the comparator's in five of six $\tau^2$ cells; on 32K airline the comparator is higher (16.0 against 12.0).}\label{tab:reliability}
\setlength{\tabcolsep}{4pt}
\begin{tabular}{lrrrrrr}
\toprule
& \multicolumn{3}{c}{32K} & \multicolumn{3}{c}{128K} \\
\cmidrule(lr){2-4}\cmidrule(lr){5-7}
Method & Retail & Airline & Telecom & Retail & Airline & Telecom \\
\midrule
Native attention & \textbf{16.7} & \textbf{16.0} & 10.5 & \textbf{21.1} & \textbf{10.0} & \textbf{10.5} \\
All-refresh sparse & \textbf{16.7} & \textbf{16.0} & 10.5 & \textbf{21.1} & \textbf{10.0} & \textbf{10.5} \\
Strongest feasible static & 13.2 & \textbf{16.0} & 8.8 & 15.8 & 8.0 & 7.9 \\
Training-aware IndexCache & \textbf{16.7} & 14.0 & 10.5 & \textbf{21.1} & \textbf{10.0} & \textbf{10.5} \\
LatentIndex + HS & 14.0 & 14.0 & \textbf{11.4} & 18.4 & \textbf{10.0} & \textbf{10.5} \\
\rowcolor{AIRRow}HISA + \method & 15.8 & 14.0 & 9.6 & 20.2 & \textbf{10.0} & \textbf{10.5} \\
\rowcolor{AIRRow}\method & 14.9 & 12.0 & 10.5 & \textbf{21.1} & \textbf{10.0} & \textbf{10.5} \\
\bottomrule
\end{tabular}
\par\smallskip{\scriptsize\textit{Key:} bold marks the best value in each column of the stated comparison set, ties at the reported precision included; \AIRswatch{AIRRow}~\method-based rows.}\par
\end{table}

\begin{table}[tbp]\centering\small
\caption{Eligible long-history snapshots per occupancy band; fewer tasks reach the higher bands (SWE-bench Verified: 392, 271, and 172 of 500).}\label{tab:continuation-pop}
\setlength{\tabcolsep}{4pt}
\begin{tabular}{lrrrr}
\toprule
Population & Tasks & $\varphi=0.60$ & $\varphi=0.75$ & $\varphi=0.90$ \\
\midrule
SWE & 500 & 392 & 271 & 172 \\
Terminal & 89 & 71 & 50 & 32 \\
Retail & 114 & 96 & 75 & 52 \\
Airline & 50 & 41 & 31 & 18 \\
Telecom & 114 & 99 & 77 & 49 \\
\bottomrule
\end{tabular}

\end{table}

\subsection{Training dynamics}
\label{app:results-training}
\label{sec:results-training}

\begin{figure*}[tbp]
\centering
\includegraphics{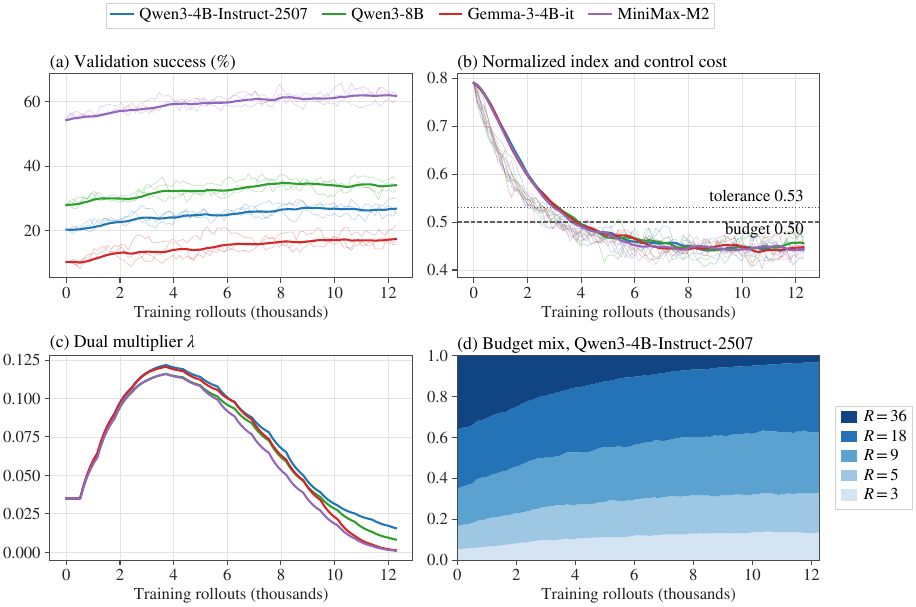}
\caption{Training brings every 128K backbone below the 0.50 budget (dashed; dotted: tolerance) within about 4,000 rollouts while validation success rises by 6.2 to 7.6 points. (d) Budget mix of the primary model.}
\label{fig:training}
\end{figure*}

\begin{figure*}[tbp]
\centering
\includegraphics{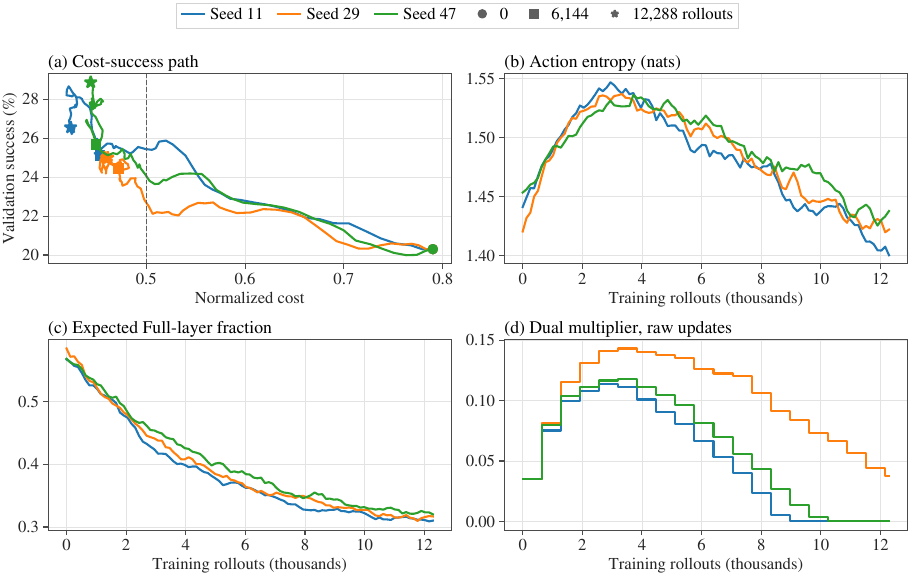}
\caption{Each seed of the primary controller moves toward the budget while raising validation success, explores before it commits, and keeps a mixed allocation. (a) success against cost, with markers at 0, 6,144, and 12,288 rollouts; (b) action entropy; (c) expected Full-layer fraction; (d) dual updates.}
\label{fig:train-joint}
\end{figure*}

Constrained training behaves the same way on every backbone at 128K (Figure~\ref{fig:training}). Normalized cost starts at 0.79 and crosses the 0.50 budget after about 3,700 to 4,000 rollouts, as the dual multiplier peaks near 0.12; once the constraint is slack, the multiplier decays toward zero and cost settles at 0.44 to 0.46, while validation success rises by 6.2 to 7.6 points. The primary controller moves most of its probability from $R=36$ (36\% at the start, 3\% at the end) to $R=9$ and $R=18$, which together receive about two thirds of its choices, and it still uses $R=3$ and $R=5$ for a third of its decisions. In every seed of the primary controller, action entropy rises to about 1.55 nats in the first 3,000 to 3,700 rollouts and then falls to 1.40 to 1.44, the expected Full-layer fraction falls from 0.57 to 0.31 or 0.32, and seed 29 keeps a positive dual price longest, at 0.04 after the last update (Figure~\ref{fig:train-joint}).

\begin{figure*}[tbp]
\centering
\includegraphics{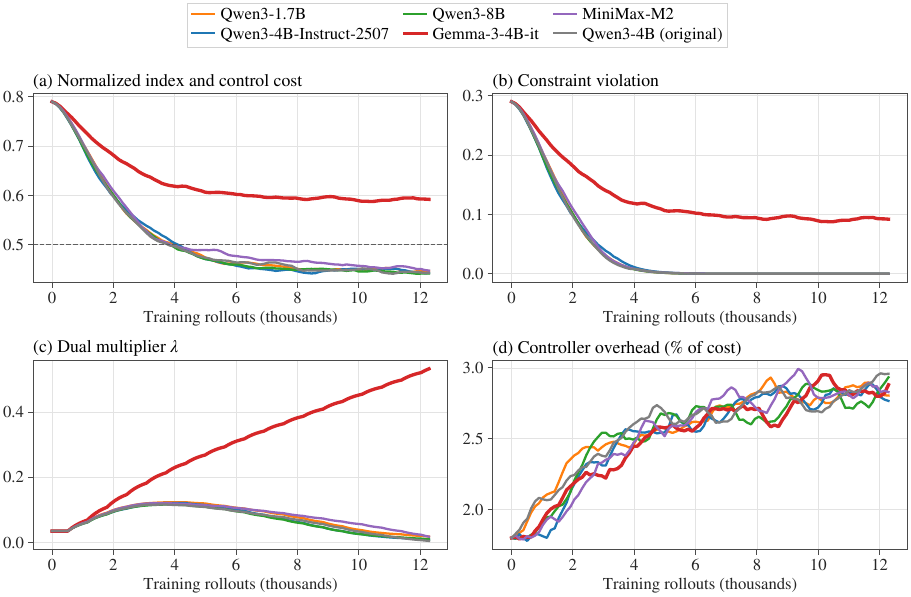}
\caption{At 32K, cost falls from 0.79 to 0.44--0.45 on five of six backbones and their violations reach zero; Gemma-3-4B-it ends at 0.59 with its dual multiplier still rising. Dashed: the 0.50 budget.}
\label{fig:train-resources}
\end{figure*}

\begin{figure*}[tbp]
\centering
\includegraphics{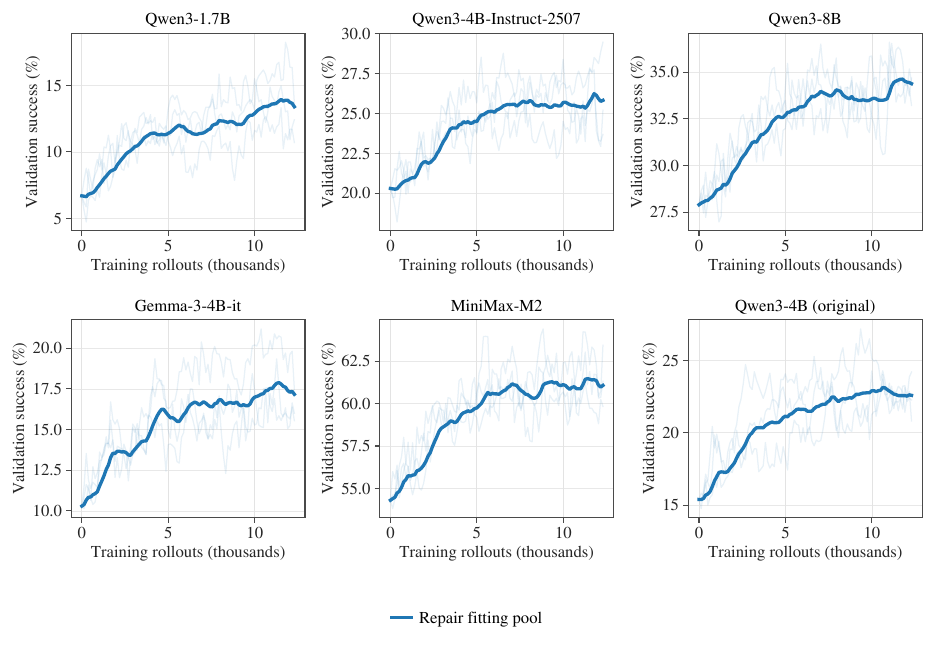}
\caption{At 32K, seed-mean validation success rises by 5.6 to 7.2 points for every backbone over 12,288 rollouts.}
\label{fig:train-validation}
\end{figure*}

\begin{figure*}[tbp]
\centering
\includegraphics{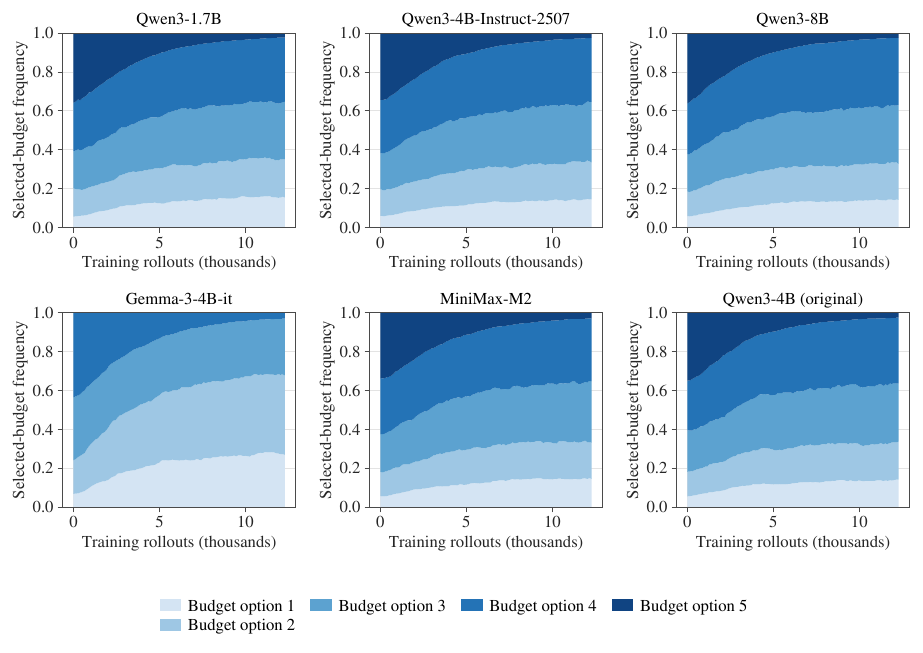}
\caption{Every 32K controller still uses at least three refresh options more than 10\% of the time at the end of training. Options run from fewest to most Full layers within each backbone.}
\label{fig:train-mix}
\end{figure*}

\begin{figure*}[tbp]
\centering
\includegraphics{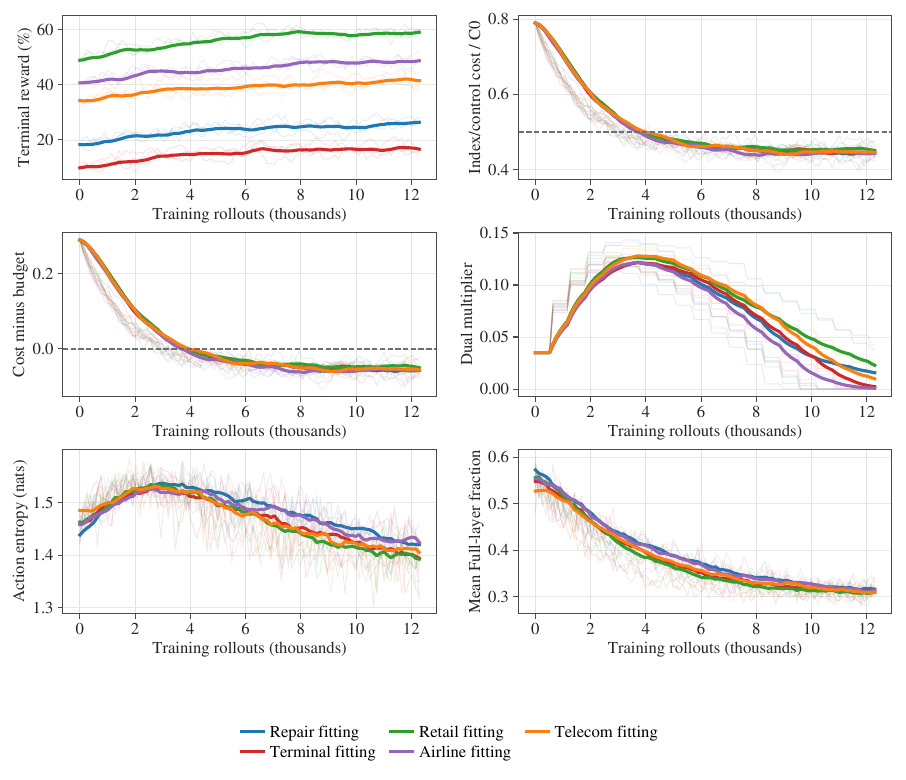}
\caption{Controllers trained in each of the five fitting domains converge to a cost of 0.44 to 0.46, below the 0.50 budget, and raise validation success by 5.7 to 9.7 points.}
\label{fig:train-domains}
\end{figure*}

At 32K, five of six backbones reach the budget and their constraint violations fall to zero; Gemma-3-4B-it, with five eligible layers, ends at a cost of 0.59 with a violation of 0.09 and its dual multiplier still rising, at 0.53, which is why validation deploys its static fallback (Figure~\ref{fig:train-resources}). Seed-mean validation success at 32K rises by 5.6 to 7.2 points and terminal reward by 5.4 to 9.0 points for every backbone over 12,288 rollouts (Figure~\ref{fig:train-validation}). Every 32K controller still uses at least three refresh options more than 10\% of the time at the end of training; the Qwen and MiniMax-M2 controllers put about two thirds of their choices on their third and fourth options, and Gemma-3-4B-it almost all on its three smallest (Figure~\ref{fig:train-mix}). Controllers trained separately in each of the five fitting domains converge to a cost of 0.44 to 0.46, place about two thirds of their choices on $R=9$ and $R=18$, and raise validation success by 5.7 to 9.7 points (Figure~\ref{fig:train-domains}); the terminal and tool-use controllers among them are the in-domain controllers above.

\section{Prompts, Model Adapters, and Case Studies}
\label{app:cases}

These pages show what each agent sees and what the controller decides at events that change the allocation question. Every population uses the same system instruction, given once below, followed by its own domain instruction, task input, and compact tool interface; every method on a backbone receives the same instruction text and tool schemas. Benchmark task messages and policy documents come verbatim from the pinned harnesses, the $\tau^2$ user simulator is \texttt{gpt-4.1-2025-04-14} at temperature 0, and telecom keeps separate agent and user actors. A further page defines how each backbone serializes and validates tool calls.

The case studies follow \method and comparators from the same starting task: an SWE-bench Verified event card (the example of Section~\ref{sec:results-case}), an airline task in which \method ends worse than its comparator, and a telecom task that separates a requested user action from its completion. Columns align at informative observed events, and branches may contain different numbers of model calls, so prefixes and counters can differ after the branches separate. At every call the teal panel reports the refresh choice, the prefix length, and two cumulative counters: Full-layer selections ($qR$ for an invocation with $q$ prefill and decode queries at budget $R$) and index-key updates (36$q$ for \method, which maintains keys at every eligible layer, and $9q$ for the specialized fixed $R=9$ runtime). The counters cover index work only, so their difference is not a speedup. The panels use one color per role: purple for shared instructions, blue for agent operations, slate for tool, environment, or user observations, teal for controller telemetry, and amber for end states. A single episode cannot rank methods; the success rates are those of Section~\ref{sec:results}.

\label{app:prompts}
\begin{figure*}[tbp]\begin{minipage}{\textwidth}\caseboxes\input{results/cases/prompt_shared.tex}\end{minipage}\end{figure*}
\begin{figure*}[tbp]\begin{minipage}{\textwidth}\caseboxes\input{results/cases/prompt_swe.tex}\end{minipage}\end{figure*}
\begin{figure*}[tbp]\begin{minipage}{\textwidth}\caseboxes\input{results/cases/prompt_terminal.tex}\end{minipage}\end{figure*}
\begin{figure*}[tbp]\begin{minipage}{\textwidth}\caseboxes\input{results/cases/prompt_retail.tex}\end{minipage}\end{figure*}
\begin{figure*}[tbp]\begin{minipage}{\textwidth}\caseboxes\input{results/cases/prompt_airline.tex}\end{minipage}\end{figure*}
\begin{figure*}[tbp]\begin{minipage}{\textwidth}\caseboxes\input{results/cases/prompt_telecom.tex}\end{minipage}\end{figure*}

\label{app:contracts}
\begin{figure*}[tbp]\begin{minipage}{\textwidth}\caseboxes\input{results/cases/model_adapter_contract.tex}\end{minipage}\end{figure*}

\label{app:case-studies}
\begin{figure*}[t]
\centering
\begin{minipage}{\textwidth}
\caseboxes
\input{results/cases/case_maintext.tex}
\end{minipage}
\caption{SWE-bench Verified event card for Section~\ref{sec:results-case}: after a new test diagnostic, \method raises its budget from $R=3$ to $R=18$ at the next model call, while fixed IndexCache keeps $R=9$. Counts are cumulative Full-layer selections and index-key updates.}
\label{fig:case}
\end{figure*}
\newcommand{\casetext}[1]{\par\begingroup\caseboxes\input{#1}\par\endgroup}
\newcommand{\caseevent}[3]{\begin{figure*}[tbp]\begin{minipage}{\textwidth}\caseboxes\input{results/cases/case_#1_event#2.tex}\end{minipage}\caption{#3}\label{fig:case-#1-#2}\end{figure*}}

\setcounter{dbltopnumber}{3}\renewcommand{\dbltopfraction}{.8}
\clearpage\raggedbottom
\par\begingroup\caseboxes\caseheading{A late constraint can expose a scheduling miss}{What does an unfavorable allocation response look like in an event-aligned comparison?}
\begin{rolebox}[title={Shared initial task}]{taskpurple}
Please move my B-booking to a later departure on the same date. I need to keep my checked bag. The same booking, user preference and initial policy are available. The booking interface exposes later-option as a candidate whose conditions still need a quote.
\end{rolebox}
\par\endgroup
\vspace{7pt}{\noindent\figfont Figures~\ref{fig:case-airline-1}--\ref{fig:case-airline-3} align the two branches at three events.}\par
\par\begingroup\caseboxes\begin{rolebox}[title={Visible final messages}]{agentblue}
\textbf{AIR.} ``I still need to reconcile the checked-bag condition before changing the booking.'' \textbf{Fixed IndexCache R9.} ``The later fare adds a \$35 checked-bag charge. Do you approve that charge before I change the flight?''
\end{rolebox}
\begin{rolebox}[title={End state and interpretation}]{authoramber}
\textbf{AIR.} The change is still unresolved at the end of the episode.

\textbf{Fixed IndexCache R9.} The necessary customer clarification has been requested; no unconfirmed change is committed.\par

AIR's allocation response is unfavorable here: it needs an additional recovery call and leaves the change unresolved, while fixed IndexCache requests the customer clarification first. The case identifies a failure mode to test for; its cause and frequency need the aggregate results.
\end{rolebox}
{\figfont AIR maintains all 36 eligible-layer keys.}
\par\endgroup
\begin{figure}[H]\begin{minipage}{\textwidth}\caseboxes\input{results/cases/case_airline_event1.tex}\end{minipage}\caption{Airline case, Event 1: both methods read the current booking, \method at $R=5$ and fixed IndexCache at $R=9$.}\label{fig:case-airline-1}\end{figure}
\begin{figure}[H]\begin{minipage}{\textwidth}\caseboxes\input{results/cases/case_airline_event2.tex}\end{minipage}\caption{Airline case, Event 2: both methods request the same quote, \method at $R=3$; the quote adds a \$35 bag charge that needs the customer's confirmation.}\label{fig:case-airline-2}\end{figure}
\begin{figure}[H]\begin{minipage}{\textwidth}\caseboxes\input{results/cases/case_airline_event3.tex}\end{minipage}\caption{Airline case, Event 3: after the bag condition appears, \method answers at $R=3$ and needs a second call at $R=18$, while fixed IndexCache stays at $R=9$.}\label{fig:case-airline-3}\end{figure}

\newpage
\par\begingroup\caseboxes\caseheading{Dual control separates a request from its completion}{Does the controller react to observed shared-state changes without treating a requested user action as completed?}
\begin{rolebox}[title={Shared initial task}]{taskpurple}
Mobile data is not working on line T-line. I can help with checks on my phone. Agent-side tools and user-device tools remain assigned to their respective actors.
\end{rolebox}
\par\endgroup
\vspace{7pt}{\noindent\figfont Figures~\ref{fig:case-telecom-1}--\ref{fig:case-telecom-3} align the three methods at three events.}\par
\par\begingroup\caseboxes\begin{rolebox}[title={Visible final messages}]{agentblue}
\textbf{AIR.} ``Your device reports mobile data connected, and the line status now shows active provisioning.''

\textbf{Fixed IndexCache R9.} ``Your device reports mobile data connected, and the line status now shows active provisioning.''

\textbf{ReTopK (indexer-adapted).} ``Your device reports mobile data connected, and the line status now shows active provisioning.''
\end{rolebox}
\begin{rolebox}[title={End state and interpretation}]{authoramber}
\textbf{AIR.} The end state records matching backend and user observations.

\textbf{Fixed IndexCache R9.} The same observations are obtained.

\textbf{ReTopK (indexer-adapted).} The same observations are obtained with exact tool-output prefill.\par The controller updates its features only after the backend or the user reports the shared state, so a requested user action counts as done only once it is observed. All three methods reach the same observed end state; the case shows actor separation and feature timing.
\end{rolebox}
{\figfont AIR maintains all 36 eligible-layer keys.}
\par\endgroup
\begin{figure}[H]\begin{minipage}{\textwidth}\caseboxes\input{results/cases/case_telecom_event1.tex}\end{minipage}\caption{Telecom case, Event 1: all three methods read the line state, which reports stale provisioning; \method runs at $R=5$.}\label{fig:case-telecom-1}\end{figure}
\begin{figure}[H]\begin{minipage}{\textwidth}\caseboxes\input{results/cases/case_telecom_event2.tex}\end{minipage}\caption{Telecom case, Event 2: after the user confirms the backend refresh, \method raises its budget to $R=18$; the API acknowledges the request, which does not restore connectivity.}\label{fig:case-telecom-2}\end{figure}
\begin{figure}[H]\begin{minipage}{\textwidth}\caseboxes\input{results/cases/case_telecom_event3.tex}\end{minipage}\caption{Telecom case, Event 3: once the user-side actions are observed, all three methods read the line status again, which now shows active provisioning; \method runs at $R=9$.}\label{fig:case-telecom-3}\end{figure}

\end{document}